\documentclass{article}

\PassOptionsToPackage{numbers, compress}{natbib}

\usepackage[preprint]{neurips_2025}

\usepackage[utf8]{inputenc} 
\usepackage[T1]{fontenc}    
\usepackage{hyperref}       
\usepackage{url}            
\usepackage{booktabs}       
\usepackage{amsfonts}       
\usepackage{nicefrac}       
\usepackage{microtype}      
\usepackage{xcolor}         

\usepackage{enumitem}
\usepackage{amsmath}
\usepackage{graphicx}
\usepackage{multirow}
\usepackage{subcaption}
\usepackage{listings}
\usepackage{xcolor}
\title{GridPE: A Grid Cell-Inspired Unified Position Embedding for Arbitrary-Dimensional Spaces}

\author{%
  Boyang Li \\
  New York University\\
  \texttt{boyang.li@nyu.edu} \\
  \And
  Yulin Wu \\
  Peking University \\
  \texttt{cover.wu@stu.pku.edu.cn} \\
  \And
  Nuoxian Huang\\
  Imperial College London\\
  \texttt{n.huang25@imperial.ac.uk}
  \And
  Wenjia Zhang \\
  Tongji University \\
  \texttt{wenjiazhang@tongji.edu.cn}
}

\begin{document}

\maketitle

\begin{abstract}
    Understanding spatial relationships across all dimensions is fundamental for intelligent systems. However, existing positional embeddings, such as Rotary Positional Embedding (RoPE), lack theoretical guarantees for high-dimensional spatiotemporal tasks like video understanding and robotic navigation. Inspired by the hexagonal periodic coding of grid cells in mammalian spatial cognition, we propose \textbf{GridPE}---a novel positional embedding framework that integrates computational neuroscience principles with harmonic analysis. Our approach builds upon Random Fourier Features and leverages principles from neuroscience to construct efficient embeddings. Theoretically, we prove that any translation-invariant spatial function can be approximated by a finite sum of Fourier bases, which naturally reduces to RoPE in the one-dimensional case. We then derive the directions and quantities of frequency vectors at each scale in any Euclidean dimension, along with the optimal ratio between different scales, from a bioavailability perspective. These derivations are equivalent to the relationship between the centroid and the vertices of a regular simplex in that dimension. We validate GridPE across a range of spatial modeling tasks, including 2D image classification (ImageNet100) and 3D point cloud recognition (ModelNet40). Our theoretical analysis establishes GridPE as a unified framework for positional embedding in arbitrary-dimensional spaces, while empirical results demonstrate its superiority over existing methods.
\end{abstract}

\section{Introduction}
Understanding spatial relation plays a crucial role in constructing an internal model of the world, enabling systematic organization of knowledge across various domains of behavior \cite{tolman_cognitive_1948}. This capability is therefore essential for intelligent systems to structure knowledge within complex, high-dimensional cognitive maps \cite{behrens2018cognitive}. Meanwhile, the Transformer architecture—thanks to its self-attention mechanism and distinctive network organization—excels at processing and representing such high-dimensional, complex topological relationships, which likely underlies its success across a wide range of tasks \cite{reinauer2022persformertransformerarchitecturetopological}.

However, the Transformer itself is not inherently equipped to understand the positional information of input units, which is critical for many tasks. As a result, substantial research has focused on enhancing position embedding to enable Transformers to better capture this information \cite{li2021learnable,su2024roformer, raffel2020exploring, press2021train}. Among these advancements, Rotary Position Embedding (RoPE) \cite{su2024roformer} has gained significant attention. By applying a rotational matrix to the query and key vectors, RoPE helps Transformers capture the relative positional relationships between tokens. Due to its excellent performance in long-context comprehension, RoPE has been widely adopted in major large language models, such as LLaMA and Qwen \cite{wang2024qwen2, touvron2023llamaopenefficientfoundation}. However, RoPE was originally designed for language tasks in 1D space. While some efforts have been made to extend it to high-dimensional position embeddings for complex spatiotemporal tasks---such as image recognition, video processing, 3D object detection, and multimodal applications \cite{wang2024qwen2,heo2024rotary,wei2025videorope,ji2025ropetr,lu2024unified}---a unified theoretical approach remains absent.

\begin{figure}[t]
  \centering
  \includegraphics[width=0.9\linewidth]{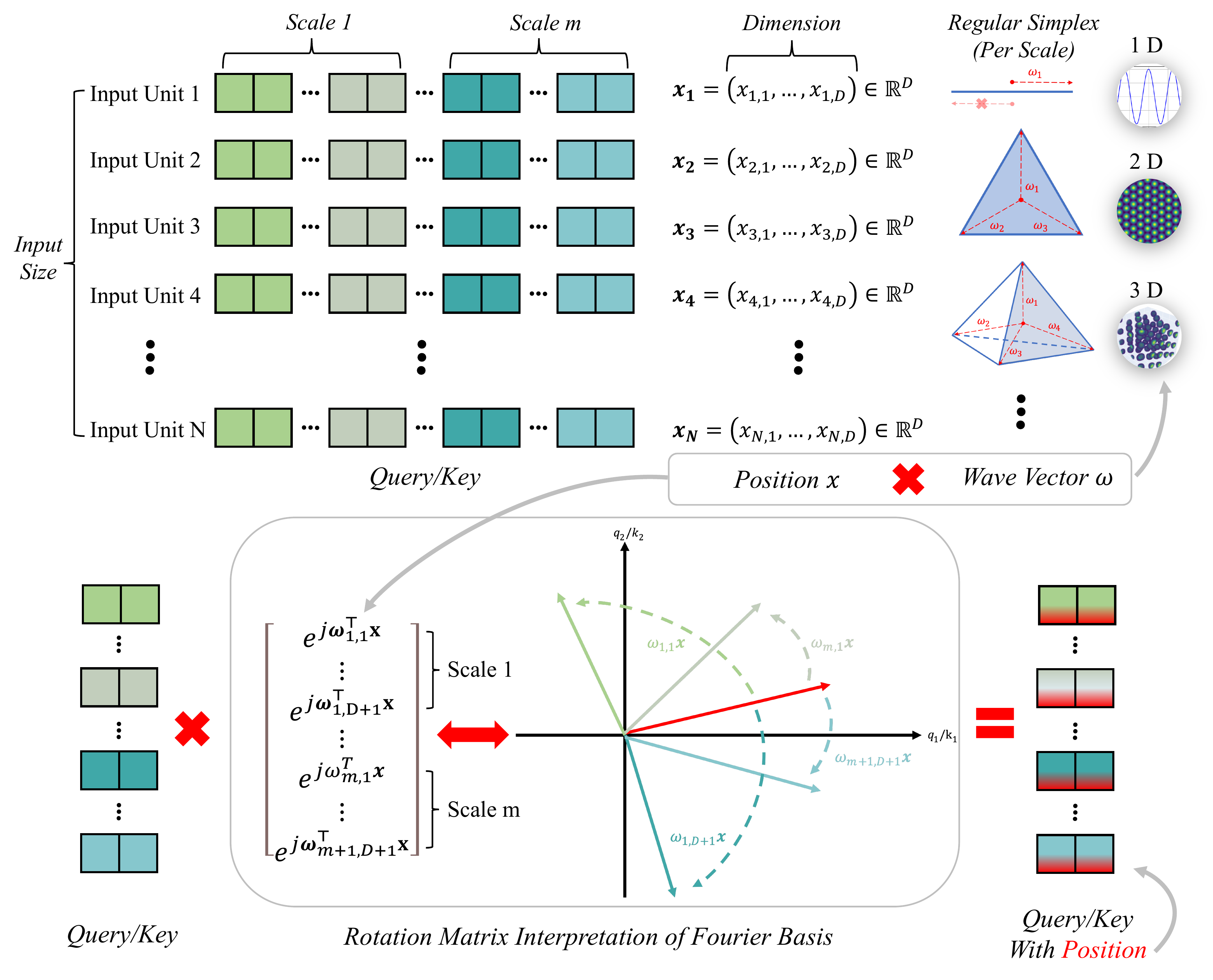}
  \caption{Overview of GridPE: a Fourier-based positional embedding framework for arbitrary dimensions, inspired by grid cells and constructed using simplex geometry.}
  \label{fig:overall}
\end{figure}
This challenge is closely tied to a fundamental question in neuroscience---how does the mammalian brain represent spatial features and relationships in multi-dimensional euclidean space? The idea of a “cognitive map” was first proposed by Tolman to describe an internal representation that enables flexible navigation and behavior \cite{tolman_cognitive_1948}. Decades later, this concept found its neural substrate in the discovery of grid cells—neurons in the medial entorhinal cortex (MEC) that fire in spatially periodic patterns. These cells form a regular hexagonal lattice across the environment, activating whenever the animal is near a vertex of this internal grid and effectively tiling space with repeating receptive fields. This spatial code is thought to support multi-scale self-localization, encode metric distances, and facilitate vector-based navigation and path integration \cite{gorchetchnikov2007space, moser2008place, banino2018vector, gao2019learning,bush2015using,rowland2016ten}. Due to their close association with spatial representation, the principles of grid cell firing have been widely adopted by AI researchers, extending beyond neuroscience to fields such as robotics navigation system design \cite{yuan2015entorhinal}, reinforcement learning agent trajectory planning \cite{yu2020prediction}, and positional encoding for geospatial data \cite{mai2020multi, mai2023sphere2vec,mai2022review}.

While many studies heuristically apply grid-cell-like patterns in 2D without a principled spectral or kernel-based foundation, our method grounds positional embedding design in Random Fourier Features~\cite{rahimi2007random} and velocity-controlled oscillation theory (VCO)~\cite{burgess2007oscillatory,hasselmo2008grid}, enabling biologically inspired representations in arbitrary dimensions. To support this intuition, we first prove that, from the viewpoint of translation invariance, the activation intensity of grid populations within a high-dimensional Euclidean space (e.g., 1D, 2D, or 3D spaces) can be expressed as a sum of Fourier basis functions. In the one-dimensional scenario, this expression naturally reduces to RoPE. Building on this, we extend the periodic coding of space by grid cells as described in the VCO theory, and derive the directions and quantities of frequency vectors at each scale in any Euclidean dimension. These wave vectors can be understood as analogous to the directions and quantities from the centroid to the vertices of a regular simplex in that dimension, where each scale corresponds to a different configuration of vectors. Figure~\ref{fig:overall} illustrates how the geometry of a regular simplex gives rise to the wave vectors used in constructing the positional embedding. From a bioavailability perspective, we also determine the optimal ratio between different scales, which ensures the most efficient representation of spatial information. This yields a scalable and generalizable positional embedding framework that improves upon existing methods and enhances performance across diverse tasks and dimensions.

To evaluate the proposed approach, we conducted experiments on two benchmark tasks: image classification on the ImageNet-100 dataset 
and point cloud classification on the ModelNet-40 dataset. 
We compared our method against existing positional embedding techniques. The experimental results show that our approach consistently improves the performance of the state-of-the-art self-attention mechanisms by preserving strong extrapolation capabilities across dimensions, underscoring the promise of incorporating insights from neuroscience into the design of artificial intelligence systems.

\section{Related Work}
\paragraph{Positional Embedding for Length Extrapolation.} 
In the context of 1D sequence length extrapolation, sinusoidal positional encodings (PEs), first introduced in the Transformer model, face challenges when extrapolating to longer sequences beyond those seen positions during training. Several methods have been proposed to address this limitation. For instance, integrating shift invariance into absolute positional encodings (APEs) \cite{kiyono2021shape} or improving smoothness through continuous functions \cite{wang2019encoding} have been suggested to enhance extrapolation for longer sequence lengths. While absolute positional encodings (APEs) map each position to a unique representation, relative positional encodings (RPEs) capture the relative distances between tokens, which allows for better generalization to unseen sequence lengths. Further simplifications of RPEs, such as T5-Bias \cite{raffel2020exploring} and ALiBi \cite{press2021train}, reduce computational complexity while maintaining extrapolation capabilities.  However, despite their strengths in handling sequence length variations, RPEs generally decouple positional information from semantic content, limiting their ability to effectively model complex relationships between positions in a sequence. Methods like RoPE offer an advantage by utilizing Fourier-based distance-attention functions to model relative positions in a more nuanced way \cite{li2021learnable,zhao2024lengthextrapolationtransformerssurvey,zheng2021rethinking}. This approach has been widely adopted in large language models (LLMs) due to its ability to represent relative positional information effectively, while also maintaining strong extrapolation performance.

\paragraph{Extending RoPE to High-Dimensional Data.} 
Several works have explored extending RoPE to high-dimensional spaces, with a primary focus on image and video large models. One straightforward approach involves flattening the high-dimensional inputs into a 1D sequence, enabling the direct application of 1D RoPE. However, despite improvements in indexing and attention mechanisms proposed by variants such as TAD-RoPE \cite{gao2024tcllavarethinkingtransferimage}, these extensions still struggle to capture the rich spatiotemporal structures inherent in modalities in video \cite{wang2024qwen2}. Building on this, some approaches address the high-dimensional nature of the data by treating each dimension independently and applying separate rotations to each. For instance, the axial RoPE proposed by Su et al \cite{su2024roformer} provides a theoretical formulation for 2D RoPE, which has been implemented in several models to enhance their spatial representation. Examples include EVA-02 \cite{fang2024eva}, which incorporates 2D axial RoPE into a language-aligned vision model, and Unified-IO \cite{lu2024unified}, which employs 2D RoPE for novel multi-modal modeling to capture both spatial and temporal information. Additionally, models like Qwen2-VL's M-RoPE \cite{wang2024qwen2} and VideoRoPE \cite{wei2025videorope} extend RoPE to 3D, aiming to improve the understanding of spatiotemporal relationships, with a particular focus on resolution scaling. However, despite their success in spatial representation learning, these methods face limitations when it comes to modeling distances between positions in arbitrary directions in the high-dimensional space, thereby constraining their ability to fully capture the complex spatial relationships inherent in high-dimensional data. Another approach considers treating multiple axes as a unified entity for rotation, as demonstrated by RoPE-Mixed \cite{heo2024rotary}. This method allows for modeling diagonal distances but lacks a rigorous theoretical justification for its effectiveness, particularly in determining the required number and directions of wave vectors for higher dimensions.

\section{Preliminaries}
\subsection{Grid Cell Modeling Based on VCO Theory}
\begin{figure}[ht]
    \centering
    \begin{minipage}{0.43\linewidth}
        \centering
        \includegraphics[width=\linewidth]{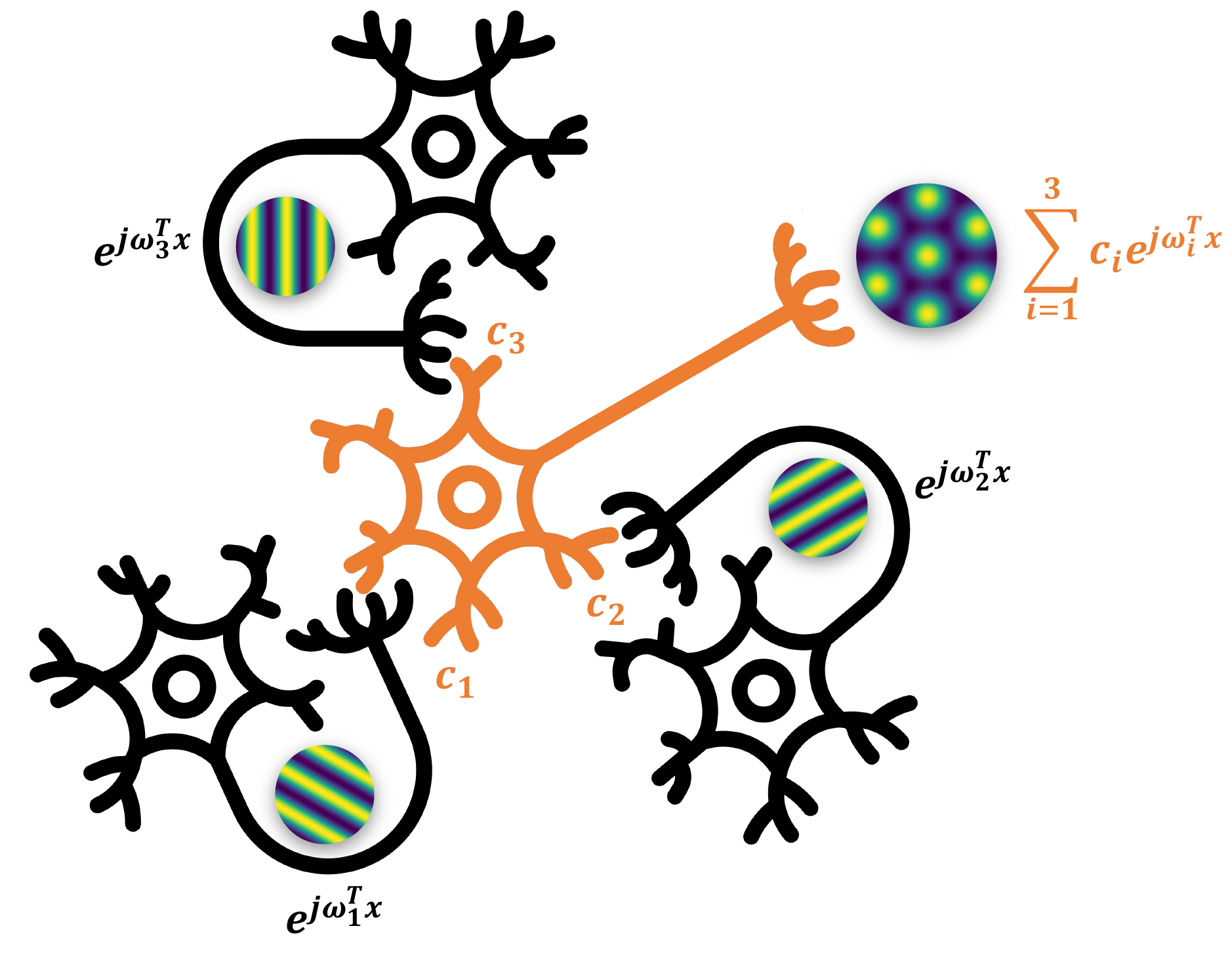}
        \caption{Illustration of VCO theory. Black neurons represent presynaptic cells with simple dendritic oscillations, while the orange neuron is a grid cell whose grid-like firing pattern emerges from the summation of three oscillations with different preferred directions.}
        \label{fig:vco_illustration}
    \end{minipage}
    \hspace{0.05\linewidth}
    \begin{minipage}{0.5\linewidth}
        \centering
        \includegraphics[width=\linewidth]{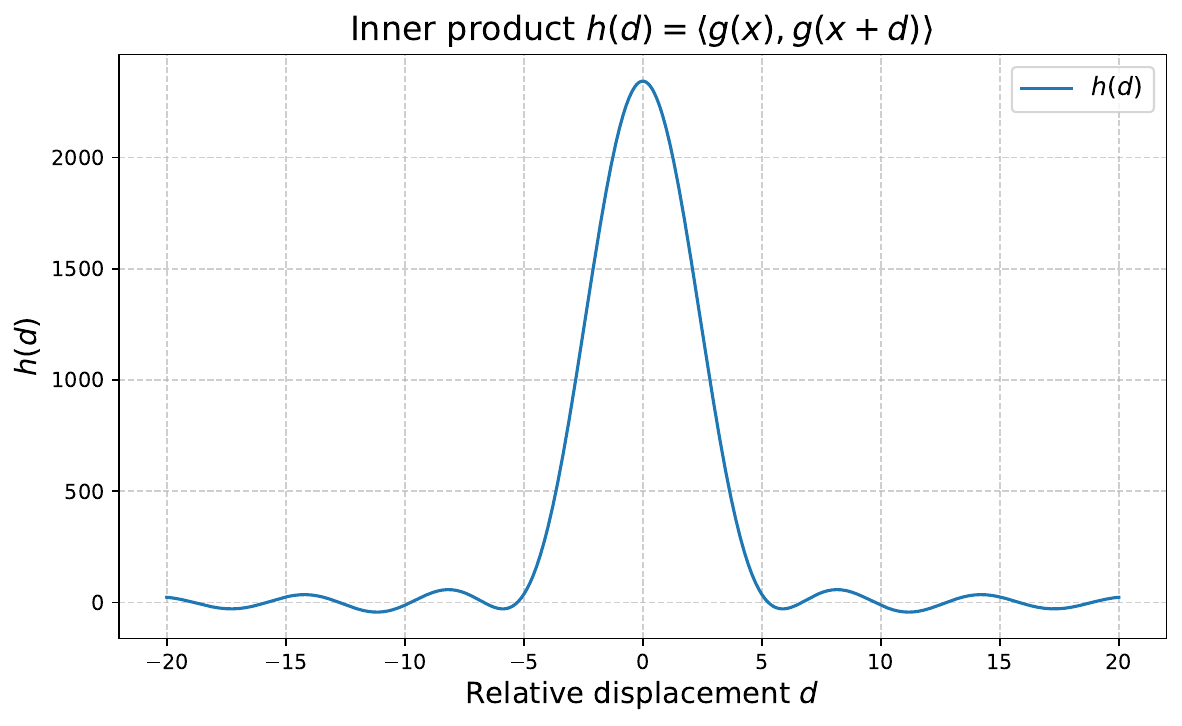} 
        \caption{Decay of $h(\mathbf{d})$ with displacement $\mathbf{d}$, assuming uniformly distributed frequency components \( \boldsymbol{\omega} \). The function shows periodic behavior with decaying amplitude as the distance increases.}
        \label{fig:gd_decay}
    \end{minipage}
\end{figure}

In this work, we model grid cells based on the velocity-controlled oscillation (VCO) theory \cite{burgess2007oscillatory, hasselmo2008grid}. Under VCO, each presynaptic input to a grid cell generates an oscillation whose instantaneous frequency shifts with the agent’s velocity projected onto that input’s preferred direction, superimposed on a baseline frequency \cite{anselmi2020computational,yu2020prediction}:
\begin{equation}
    \frac{d\Phi}{dt} = \omega_d = \omega_s + \beta \, \mathbf{v}^\top \boldsymbol{\omega}
    \label{eq:vco}
\end{equation}
Here, \(\omega_d = d\Phi/dt\) is the instantaneous oscillation frequency, and \(\omega_s\) the baseline frequency.  The vectors \(\mathbf{v}\) and \(\boldsymbol{\omega}\in\mathbb R^2\) represent the agent’s velocity and a preferred direction (with magnitude as spatial frequency), respectively, while \(\beta\) scales the effect of velocity. When we integrate the modulated frequency \( \omega_d \) over the time \( t \), we obtain the oscillation phase:
\begin{equation}
    \Phi(t) = \int_0^t \left( \omega_s + \beta \, \mathbf{v}^\top \boldsymbol{\omega} \right) dt 
    = \omega_s t + \beta \, \boldsymbol{\omega}^\top \mathbf{x}(t)
    \label{eq:phi_integral}
\end{equation}
where \( \mathbf{x}(t) \) is the displacement vector from time 0 to time \( t \). Evaluating Equation~\ref{eq:phi_integral} at a fixed time \(t_0\) lets us reinterpret the phase as a function of space alone, i.e., \(\Phi\) as a function of \(\mathbf{x}\).  We therefore write:
\begin{equation}
\Phi(\mathbf{x})
= \Phi(t)\bigl|_{t=t_0}
= \omega_s t_0 + \beta\,\boldsymbol{\omega}^\top\mathbf{x},
\label{eq:phi_integral_t0}
\end{equation}
Here, \(\mathbf{x} = \mathbf{x}(t_0)\) indicates the displacement vector from time 0 to time \( t_0 \).  Equation \ref{eq:phi_integral_t0} emphasizes \(\Phi\)'s dependence on the displacement \(\mathbf{x}\). As a result, all points with the same projection \(\boldsymbol{\omega}^\top \mathbf{x}\) share the same phase, creating a parallel “stripe” of activation across space. By summing several such stripe patterns—each generated by a different presynaptic cell’s preferred direction \(\boldsymbol{\omega}_i\)—we obtain the familiar hexagonal grid of grid cell firing as shown in Figure \ref{fig:vco_illustration}. Equivalently, this process can be written as a sum of Fourier basis functions~\cite{dang2021grid}:
\begin{equation}
    g(\mathbf{x}) = \sum_{i=1}^{n} c_i \cos\left(\Phi_i(\mathbf{x})\right) 
    = \sum_{i=1}^{n} \frac{c_i}{2} \left( e^{j\Phi_i(\mathbf{x})} + e^{-j\Phi_i(\mathbf{x})} \right)
    \label{eq:grid_fourier_expansion}
\end{equation}
Here, \(n\) is the number of presynaptic inputs ($n=3$ in Figure \ref{fig:vco_illustration}), \(\Phi_i(\mathbf{x})\) denotes the spatial phase of the \(i\)-th oscillatory input at position \(\mathbf{x}\), and \( c_i\) is its corresponding weight. Because negative frequencies add no new information about spatial periodicity or phase relationships, we retain only positive‑frequency components. Moreover, since each presynaptic input has its own baseline theta frequency \(\omega_{s}\), we absorb that dependence into \(\tilde{c}(\boldsymbol{\omega})\) and fix \(t = t_0\), treating these coefficients as constants. Here, \(\tilde{c}(\boldsymbol{\omega})\) captures the time‑independent complex amplitude for each Fourier basis, and we omit the parameter \( \beta \), which can be integrated into the wave vector \( \boldsymbol{\omega} \). For notational convenience, we then replace the sum over input indices \(i\) with a sum over their corresponding wave vectors \(\boldsymbol{\omega}\). Combining these choices, the grid cell’s activation at location \(\mathbf{x}\) can be written as:
\begin{equation}
    g(\mathbf{x})
    = \sum_{\boldsymbol{\omega}\in\Omega} \tilde{c}(\boldsymbol{\omega})\,e^{\,j\boldsymbol{\omega}^\top\mathbf{x}},
    \quad
    \tilde{c}(\boldsymbol{\omega})
    = \frac{c_i}{2}\,e^{\,j\omega_s t_0},
    \label{eq:fourier_grid_dynamic}
\end{equation}
Previous research typically uses three Fourier bases with directional preferences forming \(120^\circ\) angles~\cite{dang2021grid, blair2007scale}. This configuration results in a hexagonal lattice structure, which efficiently and uniformly covers two-dimensional space, minimizing overlap and maximizing the distinctiveness of each cell's spatial representation. An illustration of this hexagonal firing pattern can be found in Figure~\ref{fig:vco_illustration}. This geometric efficiency is crucial for spatial navigation and representation, providing a regular, repeating pattern that enhances the precise encoding of an animal’s environment and location.
\subsection{Translational Invariance of the Grid Cell}  
\label{sec:grid_translation}
Grid cells represent not only absolute locations in the Euclidean space but also relative spatial shifts between locations, enabling the encoding of free vectors and facilitating vector-based navigation. The inner product of grid cell representations sheds light on this capability. At a fixed time point, the representation of a grid cell at location \( \mathbf{x} \) is given by Equation~\ref{eq:fourier_grid_dynamic}.  If two locations differ by \(\mathbf{d}\), i.e.\ \(\mathbf{x}-\mathbf{y}=\mathbf{d}\), then the functional inner product between \(g(\mathbf{x})\) and its shifted version \(g(\mathbf{y})\) over the entire space is:
\begin{equation}
\langle g(\mathbf{x}), g(\mathbf{y}) \rangle = \sum_{\boldsymbol{\omega}_1} \sum_{\boldsymbol{\omega}_2} c(\boldsymbol{\omega}_1) \, c(\boldsymbol{\omega}_2)^* \, 
e^{j \boldsymbol{\omega}_2^\top \mathbf{d}} \int_{-\infty}^{+\infty} 
e^{j (\boldsymbol{\omega}_1^\top - \boldsymbol{\omega}_2^\top) \mathbf{x}} \, d\mathbf{x}
\label{eq:inner_product}
\end{equation}
In \(n\)-dimensional space, a standard result from Fourier analysis gives:
\begin{equation}
\int_{-\infty}^{+\infty} e^{j (\boldsymbol{\omega}_1^\top - \boldsymbol{\omega}_2^\top) \mathbf{x}} \, d\mathbf{x} = (2\pi)^n \delta(\boldsymbol{\omega}_1 - \boldsymbol{\omega}_2),
\end{equation}
where \( \delta(\cdot) \) is the Dirac delta function, which is nonzero only when \(\boldsymbol{\omega}_1 = \boldsymbol{\omega}_2\). Substituting this into Equation~\ref{eq:inner_product}, we obtain:
\begin{equation}
    \langle g(\mathbf{x}), g(\mathbf{x} + \mathbf{d}) \rangle 
    = (2\pi)^n \sum_{\boldsymbol{\omega}} \left| c(\boldsymbol{\omega}) \right|^2 e^{j \boldsymbol{\omega}^\top \mathbf{d}} 
    = h(\mathbf{d})
    \label{eq:grid_shift_kernel}
\end{equation}
Therefore, for a given grid cell, the inner product of its activation map and its shifted version represents the shift vector, independent of specific locations. Grid cells in the mammalian brain span multiple scales and orientations~\cite{mathis2012optimal}. Consequently, their joint response to an offset vector \( \mathbf{d} \) can be modeled by integrating the frequency-tuned activations of individual cells. This generalizes the discrete sum in Equation~\ref{eq:grid_shift_kernel} into a continuous integral over the population, aggregating responses across the frequency domain. Summing these over the entire population yields a translationally invariant kernel:
\begin{equation}
    k(\mathbf{x}, \mathbf{y}) = (2\pi)^n \int_{\boldsymbol{\omega}} \left| c(\boldsymbol{\omega}) \right|^2 
    e^{j \boldsymbol{\omega}^\top (\mathbf{x} - \mathbf{y})} \, d\boldsymbol{\omega}
    \label{eq:translation_invariant_kernel}
\end{equation}
Grid cells suggest that the brain represents Euclidean space via multi‐dimensional Fourier analysis, much like the auditory system processes sound waves and the visual system processes light waves. By superimposing oscillations across multiple spatial frequencies, the grid‐cell network achieves an efficient, high‐resolution encoding of space~\cite{tancik2020fourier,rowland2016ten}. This mechanism underlies the brain’s remarkable precision in building the “cognitive map” Tolman first postulated~\cite{tolman_cognitive_1948}. Moreover, as illustrated in Figure~\ref{fig:gd_decay}, the translational‐invariant kernel \(h(\mathbf{d})\) decays with increasing \(\|\mathbf{d}\|\), showing how the influence of a shift vector naturally wanes with spatial separation.

\section{Method}
\label{sec:method}
We propose \textbf{GridPE}, a grid cell-inspired positional embedding that approximates shift-invariant attention kernels using structured Fourier embeddings. We first describe the theoretical construction of the embedding and then explain its spectral design based on coverage and symmetry.
\vspace{-0.3cm}
\paragraph{GridPE Construction.}
We construct shift-invariant positional embeddings \( f(q, \mathbf{x}_1) \) and \( f(k, \mathbf{x}_2) \) whose inner product approximates any radial attention kernel of the form \( h(q, k, \mathbf{d}) \), where \( \mathbf{d} = \mathbf{x}_1 - \mathbf{x}_2 \in \mathbb{R}^n \) denotes the relative displacement. For any tolerance \(\varepsilon > 0\), we construct \( S \) frequency scales, each scale defined by a set of equal-norm vectors \(\{\boldsymbol{\omega}^{(s)}_{i}\}_{i=1}^{n+1} \subset \mathbb{R}^n\), whose directions form a regular \((n+1)\)-simplex. With these vectors, we can build a positional embedding \( f \) such that:
\begin{equation}
h(q, k, \mathbf{x}_1 - \mathbf{x}_2)
= \left\langle f(q, \mathbf{x}_1), f(k, \mathbf{x}_2) \right\rangle + \mathcal{O}(\varepsilon)
\label{eq:approx}
\end{equation}
 We fix the query \( q \) for exposition, as the treatment for the key \( k \) is symmetric. The embedding is defined by separating content and position:
\begin{equation}
f(q, \mathbf{x}) = c(q) \odot z(\mathbf{x}), \qquad
z(\mathbf{x}) =
\left[ z^{(1)}(\mathbf{x}) \;\Vert\; \cdots \;\Vert\; z^{(S)}(\mathbf{x}) \right]
\in \mathbb{C}^{S(n+1)}
\label{eq:gridpe_embedding}
\end{equation}
where \( c(q) \) and \( z(\mathbf{x}) \) are finite-dimensional content and position vectors, respectively. Each positional block is given by:
\begin{equation}
z^{(s)}(\mathbf{x}) =
\left[
e^{j (\boldsymbol{\omega}_1^{(s)})^\top \mathbf{x}},
\;\dots,\;
e^{j (\boldsymbol{\omega}_{n+1}^{(s)})^\top \mathbf{x}}
\right], \qquad s = 1, \dots, S.
\label{eq:block}
\end{equation}
Each scale contributes the \emph{minimal} set of \(n+1\) Fourier bases required to span \(\mathbb{R}^n\) with maximal rotational and reflectional symmetry. Concatenating \(S\) such scales yields a shift-invariant approximation of \(h\) with accuracy \(\mathcal{O}(\varepsilon)\). The complex exponentials in Equation \ref{eq:block} can equivalently be rewritten using real \(2 \times 2\) rotation matrices (as in RoPE), allowing the embedding to remain entirely in the real domain.
\vspace{-0.3cm}
\paragraph{Implications and Roadmap.}
GridPE provides a Fourier-based approximation to shift-invariant attention kernels using a biologically inspired set of wave vectors. The next two subsections detail: (1) how these kernels are approximated via structured Fourier bases; and (2) how simplex-based wave vector arrangements achieve efficient, symmetric spectral coverage in \(\mathbb{R}^n\), echoing geometric principles observed in grid cell representations.

\subsection{Grid Cell-inspired Fourier Embedding}
\label{sec:gridpe}
Inspired by the structured spatial coding of grid cells and the translation invariance exhibited by grid cell populations, we view their wave vectors as a natural foundation for position embedding. To bridge this biological insight with neural attention mechanisms, we modulate query and key vectors using position-dependent functions, analogous to RoPE~\cite{su2024roformer}:
\begin{equation}
    \mathbf{q}_{\mathbf{x}_1} = f(q, \mathbf{x}_1), \quad \mathbf{k}_{\mathbf{x}_2} = f(k, \mathbf{x}_2)
    \label{eq:qk_encoding}
\end{equation}
where \( f \) combines the input vectors with positional information. To ensure that attention depends only on the relative displacement \(\mathbf{d} = \mathbf{x}_1 - \mathbf{x}_2\), we impose the following constraint:
\begin{equation}
    \left\langle f(q, \mathbf{x}_1), f(k, \mathbf{x}_2) \right\rangle = h(q, k, \mathbf{d})
    \label{eq:relative_attention_kernel}
\end{equation}
We assume that \(f(q, \mathbf{x}_1)\) and \(f(k, \mathbf{x}_2)\) reside in an infinite-dimensional space, with no approximation incurred by the expansion.  The inner product then takes the form:
\begin{equation}
    \left\langle f(q, \mathbf{x}_1), f(k, \mathbf{x}_2) \right\rangle 
    = \sum_{i=0}^{\infty} f_i(q, \mathbf{x}_1) \, f_i(k, \mathbf{x}_2)
    \label{eq:infinite_inner_product}
\end{equation}
The infinite-dimensional embedding \( f(q, \mathbf{x}_1) \in \mathbb{R}^\infty \) is the projection of a square-integrable function \( \gamma(q, \cdot) \in L^2(\mathbb{R}^n) \) onto a set of orthonormal basis functions \( \{ \phi_i(\cdot - \mathbf{x}_1) \} \), centered at spatial location \( \mathbf{x}_1 \), where the placeholder \( \cdot \) denotes the spatial integration variable. Specifically, we define:
\begin{equation}
f_i(q, \mathbf{x}_1) = \langle \gamma(q, \cdot), \phi_i(\cdot - \mathbf{x}_1) \rangle.
\label{eq:hillbert_projection}
\end{equation}
This representation enables us to relate inner products between embeddings to those between the underlying functions. By Parseval’s identity, the positional shift can be equivalently transferred from the basis functions to the function \( \gamma(q, \cdot) \), resulting in an inner product between the shifted functions \( \gamma(q, \cdot + \mathbf{x}_1) \) and \( \gamma(k, \cdot + \mathbf{x}_2) \). We thus obtain:
\begin{equation}
    \left\langle f(q, \mathbf{x}_1), f(k, \mathbf{x}_2) \right\rangle = \int_{\mathbb{R}^n} \gamma(q, \mathbf{x} + \mathbf{x}_1) \, \gamma(k, \mathbf{x} + \mathbf{x}_2)^* \, d\mathbf{x} = \int_{\mathbb{R}^n} \gamma(q, \mathbf{x} + \mathbf{d}) \, \gamma(k, \mathbf{x})^* \, d\mathbf{x}
    \label{eq:translation_equiv_inner}
\end{equation}
where the second equality follows from a change of variables \( \mathbf{x} \mapsto \mathbf{x} - \mathbf{x}_2 \), which makes the dependence on the relative displacement \( \mathbf{d} \) explicit. Here, \(^*\) denotes complex conjugation, ensuring conjugate symmetry of the inner product, and for real-valued \( \gamma \), we have \( \gamma^* = \gamma \). This integral realizes the same relative attention kernel \( h(q, k, \mathbf{d}) \) as defined in Equation~\ref{eq:relative_attention_kernel}. Since \( h(q, k, \mathbf{d}) \) is translation-invariant by construction, we further assume it is radial. By Bochner’s theorem, such kernels correspond to the Fourier transform of a probability distribution \( p(\boldsymbol{\omega}) \), and can be approximated using random Fourier features (RFF)~\cite{bochner2005harmonic}. Specifically, the kernel admits the representation:
\begin{equation}
    h(q, k, \mathbf{d}) = \int_{\mathbb{R}^n} p_{(q,k)}(\boldsymbol{\omega}) \, e^{j \boldsymbol{\omega}^\top \mathbf{d}} \, d\boldsymbol{\omega}
    \label{eq:bochner_generalized}
\end{equation}
To focus on the spatial component, we treat the query and key as fixed.  
Let $\Gamma(q,\boldsymbol{\omega})$ denote the Fourier transform of $\gamma(q,\mathbf{x})$. By Parseval’s theorem, the inner product in Equation \ref{eq:translation_equiv_inner} can be equivalently written in the frequency domain as:
\begin{equation}
    \int_{\mathbb{R}^n} \gamma(q, \mathbf{x} + \mathbf{d}) \, \gamma(k, \mathbf{x})^* \, d\mathbf{x} 
    = \int_{\mathbb{R}^n} e^{j \boldsymbol{\omega}^\top \mathbf{d}} \, \Gamma(q, \boldsymbol{\omega}) \, \Gamma(k, \boldsymbol{\omega})^* \, d\boldsymbol{\omega}
    \label{eq:parseval_shift}
\end{equation}
Combining Equations~\ref{eq:translation_equiv_inner}, \ref{eq:bochner_generalized}, and \ref{eq:parseval_shift}, we obtain:
\begin{equation}
p_{(q,k)}(\boldsymbol{\omega}) = \Gamma(q, \boldsymbol{\omega}) \Gamma(k, \boldsymbol{\omega})^*
\label{eq:p_qk_def}
\end{equation}
Viewed in the Fourier basis, \( \gamma(q, \mathbf{x}) \) is characterized by its spectrum \( \Gamma(q, \boldsymbol{\omega}) \), which serves as the coordinate of \( \gamma \) under the orthonormal Fourier basis. The Equation~\ref{eq:p_qk_def} can thus be seen as a rank-1 outer product for each frequency \( \boldsymbol{\omega} \). This structure admits the decomposition:
\begin{equation}
p_{(q,k)}(\boldsymbol{\omega}) = \lambda(\boldsymbol{\omega}) \, \psi(q, \boldsymbol{\omega}) \, \psi(k, \boldsymbol{\omega})^*
\label{eq:eig_decomp}
\end{equation}
where \( \lambda(\boldsymbol{\omega}) \geq 0 \) is the spectral density and \( \psi(q, \boldsymbol{\omega}) \in \mathbb{C} \) is a normalized eigenfunction. Combining Equations~\ref{eq:eig_decomp} and~\ref{eq:p_qk_def}, we obtain \( \Gamma(q, \boldsymbol{\omega}) = \sqrt{\lambda(\boldsymbol{\omega})} \, \psi(q, \boldsymbol{\omega}) \). Accordingly, the positional embedding \( \gamma(q, \mathbf{x}) \) in Equation~\ref{eq:parseval_shift} can be expressed as:
\begin{equation}
\gamma(q, \mathbf{x}) = \int_{\mathbb{R}^n} \Gamma(q, \boldsymbol{\omega}) \, e^{j \boldsymbol{\omega}^\top \mathbf{x}} \, d\boldsymbol{\omega} 
= \int_{\mathbb{R}^n} \sqrt{\lambda(\boldsymbol{\omega})} \, \psi(q, \boldsymbol{\omega}) \, e^{j \boldsymbol{\omega}^\top \mathbf{x}} \, d\boldsymbol{\omega}
\label{eq:ift_fq}
\end{equation}
Assuming a separable form \( \psi(q, \boldsymbol{\omega}) = C(q) \, p(\boldsymbol{\omega}) \), where \( C(q) \) and \( p(\boldsymbol{\omega}) \) represent the content-dependent and spectral components, respectively, we substitute this expression into Equation~\ref{eq:ift_fq} to obtain the factorized representation \( \gamma(q, \mathbf{x}) = C(q)\, Z(\mathbf{x}) \), with:
\begin{equation}
Z(\mathbf{x}) = \int_{\mathbb{R}^n} \tilde{\rho}(\boldsymbol{\omega})\, e^{j \boldsymbol{\omega}^\top \mathbf{x}}\, d\boldsymbol{\omega}, 
\quad \text{where } \tilde{\rho}(\boldsymbol{\omega}) = p(\boldsymbol{\omega}) \sqrt{\lambda(\boldsymbol{\omega})}.
\label{eq:Z_fourier}
\end{equation}
Under this factorization, Equation~\ref{eq:translation_equiv_inner} becomes:
\begin{equation}
\left\langle f(q, \mathbf{x}_1), f(k, \mathbf{x}_2) \right\rangle 
= C(q)\, C(k)^* \underbrace{(2\pi)^n \int_{\mathbb{R}^n} \, \left| \tilde{\rho}(\boldsymbol{\omega}) \right|^2 e^{j \boldsymbol{\omega}^\top \mathbf{d}} \, d\boldsymbol{\omega}}_{\text{position-dependent term}}.
\label{eq:position_inner}
\end{equation}
Rather than sampling \(\tilde{\rho}(\boldsymbol{\omega}) \), we approximate it using a structured set of wave vectors inspired by the multi-scale tiling patterns of grid cell populations~\cite{sorscher2019unified,rodriguez2019hexagonal}, which correspond to the translation-invariant kernel in Equation~\ref{eq:translation_invariant_kernel}, introduced in Subection~\ref{sec:grid_translation}. We select a finite set of \( m \) wave vectors, each defined by a specific scale and orientation, and construct a feature map \( z(\mathbf{x}) \in \mathbb{C}^m \) that serves as a finite-dimensional approximation of the original infinite-dimensional embedding \( f(\mathbf{x}) \):
\begin{equation}\label{eq:z_multiscale_row}
z(\mathbf{x}) =
\bigl[
  \underbrace{e^{j (\boldsymbol{\omega}^{(1)}_1)^\top \mathbf{x}},\; \dots,\; e^{j (\boldsymbol{\omega}^{(1)}_{m_1})^\top \mathbf{x}}}_{z^{(1)}(\mathbf{x})} \;\Vert\;
  \cdots \;\Vert\;
  \underbrace{e^{j (\boldsymbol{\omega}^{(S)}_1)^\top \mathbf{x}},\; \dots,\; e^{j (\boldsymbol{\omega}^{(S)}_{m_S})^\top \mathbf{x}}}_{z^{(S)}(\mathbf{x})}
\bigr]
\in \mathbb{C}^{\sum_{s=1}^S m_s}.
\end{equation}
In the finite-dimensional setting, we approximate the embedding \( f(q, \mathbf{x}) \) as the elementwise product \( c(q) \odot z(\mathbf{x}) \), where \( c(q) \) encodes content and \( z(\mathbf{x}) \) captures positional structure through complex exponentials. Each complex exponential is expanded into its sine and cosine components, yielding a real-valued feature vector \( z(\mathbf{x}) \in \mathbb{R}^n \) compatible with rotary position encoding (RoPE)~\cite{su2024roformer}, where relative phase shifts are implemented as 2D blockwise rotations. The number and directions of wave vectors \( \boldsymbol{\omega} \) per scale are adapted to the ambient spatial dimensionality, following principles derived from grid cell coding, as detailed in the next subsection.

\subsection{\(\boldsymbol{\omega}\)-Components of GridPE}
\label{sec:omega_choice}
Periodic grids in \(n\)-dimensional space are constructed by superposing sinusoidal components defined by multiple wave vectors that satisfy two key criteria: \emph{coverage} and \emph{maximum symmetry}, while minimizing cell count. Coverage ensures that the grid tiles space completely, supporting full positional resolution. Maximum symmetry promotes invariance under the largest group of geometric transformations (e.g., rotations and reflections), resulting in isotropic encoding. Together, these criteria guide wave vector selection in arbitrary \(n\)-dimensional spaces, favoring lattice structures with dense tiling and high geometric symmetry.
\vspace{-0.3cm}
\paragraph{Coverage.} To ensure spatial coverage, we analyze the relationship between wave vectors and grid point locations. Grid cell firing patterns can be modeled as the superposition of spatially synchronized oscillations, where each wave vector \( \boldsymbol{\omega}_i \) imposes a planar constraint \( \boldsymbol{\omega}_i^\top \mathbf{x} = 2\pi k_i \).

To formalize this, let \( \mathbf{x} \in \mathbb{R}^n \) denote a spatial location where the phases of all oscillations align. Collecting the constraints from each wave vector yields a linear system of the form \( \boldsymbol{\Omega} \mathbf{x} = 2\pi \mathbf{k} \), where \( \boldsymbol{\Omega} \in \mathbb{R}^{N \times n} \) stacks the wave vectors row-wise, and \( \mathbf{k} = [k_1, \dots, k_N]^\top \in \mathbb{Z}^N \) contains the corresponding integer phase indices. This system admits a unique solution only if \( \text{rank}(\boldsymbol{\Omega}) = n \), in which case the grid spans the full space without degeneracy. If the rank is less than \(n\), the solution becomes underdetermined, leading to ambiguous or incomplete spatial coverage. 
\vspace{-0.3cm}
\paragraph{Maximum Symmetry.} Full coverage alone does not guarantee high symmetry, which depends critically on the arrangement of wave vectors. Let the number of wave vectors be \( M \). When \( M = n \), the grid can cover the space but typically exhibits limited symmetry. For instance, in 2D, two orthogonal vectors (90\(^\circ\)) generate a square grid with fourfold rotational symmetry, while three wave vectors arranged at 120\(^\circ\) produce a hexagonal grid with sixfold symmetry. To maximize symmetry, wave vectors should have equal magnitudes and be uniformly distributed in direction. This corresponds to a regular simplex in \( n \)-dimensional reciprocal space. In this setting, each vector points from the centroid to a vertex of the simplex, producing the highest possible symmetry group in real space. Thus, to achieve both coverage and maximum symmetry, the number of wave vectors should be \( M = n + 1 \), arranged as a regular simplex. While in 1D, symmetry is more restricted, with the symmetry group limited to translation and reflection, making a single wave vector sufficient.

For GridPE (\(n \ge 2\)), each scale employs \(M = n + 1\) complex exponentials \(e^{j\boldsymbol{\omega}_i^\top \mathbf{x}}\), where the wave vectors \(\boldsymbol{\omega}_i\) point from the centroid to the vertices of a regular simplex in reciprocal space. In the 1D case, this reduces to the RoPE form with \(M = 1\). The construction of these wave vectors is detailed in Appendix~\ref{appendix:gridpe_cal}. This simplex-based frequency organization induces an isotropic and periodic spatial representation, aligning with the high symmetry observed in biological grid cells. To further avoid ambiguity across scales and ensure efficient spatial representation, we derive the optimal ratio between wave vector magnitudes across consecutive scales in \(n\)-dimensional space based on the economy principle. The resulting optimal ratio is \(\sqrt[n]{e}\), which provides theoretical guidance for selecting the base frequency, as detailed in Appendix~\ref{appendix:grid_scale}.

\section{Experiment and Result}
\label{section:experiment}
\subsection{Experiment Setup}
\label{sec:exp_setup}
In this section, we validate the effectiveness of our GridPE on two distinct tasks: 2D image classification and 3D point-cloud classification. The ImageNet100 and ModelNet40 datasets are used as representative benchmarks for these two tasks. We push GridPE's limits by scaling the input dimensionality through higher image resolutions and larger point counts. 
\vspace{-0.3cm}
\paragraph{Baseline.} For comparison, we choose several baseline position embedding methods: 1) \textbf{Learned Positional Embedding} \cite{devlin2019bert}, which directly maps the positional information of each input token into an embedding space; 2) \textbf{RoPE-axial} \cite{su2024roformer}, where RoPE is extended to higher-dimensional spaces in a per-dimension fashion; and 3) \textbf{Rope-Mixed} \cite{heo2024rotary}, which incorporates diagonal distances suitable for vision models. These baselines are selected to provide a comprehensive evaluation of GridPE’s performance against widely used position embedding strategies in both 2D and 3D vision tasks.
\vspace{-0.3cm}
\paragraph{Implementation Details.} We use the ViT-S model~\cite{dosovitskiy2020image} for 2D image classification and the PCT model~\cite{guo2021pct} for 3D point cloud classification. For the 2D task, the DeiT training recipe~\cite{touvron2021training} is employed using the ImageNet-100 dataset~\cite{deng2009imagenet}. The model is trained for 150 epochs with a batch size of 32, and the learning rate is set to \( 5 \times 10^{-4} \), with a cosine annealing scheduler. Both Top-1 and Top-5 accuracies are used as evaluation metrics. For the 3D task, we use the ModelNet-40 dataset~\cite{wu20153d}. The PCT model is trained for 100 epochs with a learning rate of \(5 \times 10^{-5}\), and Top-1 accuracy is used for evaluation. The Adam optimizer is used for both tasks, and standard train/test splits from each dataset are followed. All models are trained under identical configurations, with the only difference being the positional embedding technique. All experiments used RTX 4090 GPUs (24GB), with 2D tasks requiring approximately $128$ GPU-hours and 3D tasks around $8$ GPU-hours.
\subsection{Experiments on the 2D Image Classification Task}
\vspace{-0.5cm}
\begin{table}[h]
    \centering
    \footnotesize
    \caption{Top-1 and Top-5 accuracy (\%) of ViT-S models with different positional embeddings on ImageNet-100 for 2D image classification under varying image resolutions.}
    \begin{tabular}{c|cc|cc|cc|cc}
    \toprule
    \multirow{2}{*}{\shortstack{\textbf{Evaluation}\\\textbf{Resolution}}} 
      & \multicolumn{2}{c|}{\textbf{Learnable-PE}~\cite{dosovitskiy2020image}} 
      & \multicolumn{2}{c|}{\textbf{RoPE-Axial}~\cite{su2024roformer}} 
      & \multicolumn{2}{c|}{\textbf{RoPE-Mixed}~\cite{heo2024rotary}} 
      & \multicolumn{2}{c}{\textbf{GridPE}} \\
    \cmidrule{2-9}
     & Top-1 & Top-5 & Top-1 & Top-5 & Top-1 & Top-5 & Top-1 & Top-5 \\
    \midrule
    160 & 49.70 & 75.94 & 55.68 & 81.38 & 55.16 & 80.06 & \textbf{57.24} & \textbf{81.64} \\
    192 & 56.06 & 81.24 & 63.32 & 86.40 & 61.78 & 85.12 & \textbf{63.42} & \textbf{87.08} \\
    224 & 59.82 & 83.92 & 66.90 & 88.64 & 65.02 & 87.90 & \textbf{67.34} & \textbf{89.34} \\
    256 & 60.54 & 85.18 & 67.96 & 89.24 & 65.76 & 88.60 & \textbf{68.54} & \textbf{89.96} \\
    320 & 61.88 & 86.10 & 66.04 & 88.80 & 64.82 & 88.20 & \textbf{67.20} & \textbf{90.12} \\
    384 & 61.12 & 85.74 & 63.06 & 87.14 & 61.44 & 86.94 & \textbf{64.82} & \textbf{88.42} \\
    448 & 59.76 & 84.54 & 60.06 & 85.52 & 58.96 & 84.74 & \textbf{61.46} & \textbf{86.52} \\
    512 & 57.24 & 82.70 & 57.32 & 83.50 & 54.44 & 81.78 & \textbf{58.72} & \textbf{84.06} \\
    \bottomrule
    \end{tabular}
    \label{tab:input_size_accuracy}
\end{table}
All ViT‑S models were trained at a input size of 224×224 and then evaluated at various resolutions to assess the robustness of different positional embedding methods for spatial interpolation and extrapolation. As shown in Table~\ref{tab:input_size_accuracy}, GridPE outperforms ohter positional embeddings across a wide range of resolutions. While all methods benefit from modest resolution increases, most exhibit performance drops as resolution continues to grow. Learnable-PE degrades more gradually under extrapolation, yet its accuracy remains consistently lower. In contrast, GridPE consistently achieves the highest accuracy across resolutions and maintains robust performance even under substantial resolution shifts. This advantage likely stems from GridPE’s structured construction of frequency vectors, which encode spatial positions using directionally balanced and geometrically symmetric bases. Such design enables more accurate modeling of relative spatial relationships, especially under unseen resolutions. In contrast, RoPE relies on axis-aligned or narrowly distributed frequency directions, limiting its ability to capture diverse spatial patterns. Learnable-PE, lacking any inductive bias toward spatial structure, correspondingly yields lower accuracy. To further evaluate robustness under different training configurations, we additionally report results using alternative training input sizes. The performance trends are visualized in subfigures~\ref{fig:2d_224}–\ref{fig:2d_320} of Figure~\ref{fig:2d3d_inputsize_comparison} in Appendix~\ref{appendix:accuracy_trends}, offering a more intuitive understanding of these comparisons.

\subsection{Experiments on the 3D Point Net Classification Task}
We evaluate several positional embedding strategies on the PCT model for 3D point cloud classification, including learnable positional embedding and RoPE-Axial. RoPE-Mixed, originally designed for 2D tasks, has not been thoroughly explored in 3D—particularly in terms of wave vector design. To address this, we use the original PCT model (without positional encoding) as a fair baseline.

When loading the ModelNet-40 dataset, we first uniformly sample each point cloud to 2048 points. During training, 512 points are selected as input to balance computational efficiency and enable evaluation of extrapolation capability. To amplify fine-grained geometric variations, we apply a scale factor to compensate for the small numerical differences inherent in point cloud coordinates. As shown in Table~\ref{tab:pct_accuracy}, the Learnable-P shows the most pronounced accuracy drop under both interpolation and extrapolation settings. Original PCT model offers minor gains but remains sensitive to input size, while RoPE-Axial provides slightly improved stability. In contrast, GridPE consistently achieves the highest accuracy across all input points and demonstrates better resilience to input density variations. 

Compared to 2D images with regular pixel grids, 3D point clouds lack a fixed spatial structure and exhibit greater variation in local density and scale. In this context, methods like RoPE-Axial and GridPE benefit from explicitly encoding geometric relationships through frequency-based constraints. GridPE, in particular, constructs directionally symmetric bases that better reflect spatial periodicity and orientation in unstructured 3D space, enabling more robust generalization under varying point distributions. These results highlight GridPE’s superior ability to capture relative spatial structures in 3D settings. Subfigures~\ref{fig:3d_512}–\ref{fig:3d_1024} of Figure~\ref{fig:2d3d_inputsize_comparison} in Appendix~\ref{appendix:accuracy_trends} further illustrate performance trends under different training input sizes, confirming the robustness of GridPE in 3D settings.
\vspace{-0.3cm}
\begin{table}[ht]
    \centering
    \footnotesize
    \caption{The test accuracy of PCT model variants with different position embedding methods across different input point numbers on the ModelNet-40 dataset for 3D point cloud classification.}
    \begin{tabular}{r|c|c|c|c}
        \toprule
        \textbf{Points} & \textbf{No-PE}\cite{guo2021pct} & \textbf{Learnable-PE}\cite{guo2021pct} & \textbf{RoPE-Axial}\cite{su2024roformer} & \textbf{GridPE} \\
        \midrule
        256  & 0.8059 & 0.8225 & 0.8464 & \textbf{0.8501} \\
        384  & 0.8821 & 0.8902 & 0.8942 & \textbf{0.8951} \\
        512  & 0.8857 & 0.9007 & \textbf{0.9052} & \textbf{0.9052} \\
        640  & 0.8825 & 0.8918 & \textbf{0.9020} & \textbf{0.9020} \\
        768  & 0.8610 & 0.8748 & 0.8857 & \textbf{0.8890} \\
        896  & 0.8460 & 0.8456 & 0.8732 & \textbf{0.8776} \\
        1024 & 0.8177 & 0.8270 & 0.8513 & \textbf{0.8558} \\
        1152 & 0.7824 & 0.7885 & 0.8124 & \textbf{0.8282} \\
        1280 & 0.7484 & 0.7464 & 0.7735 & \textbf{0.7934} \\
        1408 & 0.7010 & 0.6994 & 0.7196 & \textbf{0.7561} \\
        1536 & 0.6584 & 0.6442 & 0.6787 & \textbf{0.7204} \\
        1664 & 0.6155 & 0.5904 & 0.6337 & \textbf{0.6864} \\
        1792 & 0.5737 & 0.5450 & 0.5895 & \textbf{0.6487} \\
        1920 & 0.5458 & 0.5061 & 0.5571 & \textbf{0.6058} \\
        2048 & 0.5012 & 0.4656 & 0.5263 & \textbf{0.5721} \\
        \bottomrule
    \end{tabular}
    \label{tab:pct_accuracy}
\end{table}
\subsection{Ablation Studies and Analysis}
In this subsection, we investigate two aspects: (1) the reasons behind GridPE’s superior performance in high-dimensional tasks and (2) the key hyperparameters that most affect its behavior. 
\vspace{-0.3cm}
\paragraph{Attention Map Analysis.} To better understand GridPE’s effectiveness, we analyze attention maps using two metrics—attention distance and entropy—following a previously established method~\cite{heo2024rotary}. As detailed in Appendix~\ref{appendix:attn_map}, GridPE exhibits stable attention behavior across block layers and resolutions, consistently integrating local and global context. This pattern emerges across both 2D and 3D settings, highlighting GridPE’s ability to preserve structured and spatially aware attention even under significant variations in scale or input modality. Among the baselines, RoPE-Axial shows a similar trend but with less smooth transitions across layers, while other positional embeddings often exhibit erratic or overly concentrated attention profiles, limiting their generalization under resolution or density shifts.
\vspace{-0.3cm}
\paragraph{Ablation on Hyperparameters.} To further understand GridPE’s robustness, we conduct ablation studies on two key factors: the number of frequency scales, which are tied to attention heads under a fixed embedding size, and the orientation of wave vectors. Results are provided in Appendix~\ref{appendix:gridpe_aliation}. We find that performance peaks when the allocation of scales to attention heads is balanced, as opposed to extreme setups where too few heads handle many scales or where too many heads receive overly coarse scale information—both of which can dilute scale-specific representations and reduce spatial sensitivity. Moreover, randomly oriented wave vectors consistently outperform fixed-direction variants, underscoring the importance of directional diversity in capturing complex spatial patterns. This effect reflects GridPE’s multi-scale design, where randomly rotated simplex-based vectors provide symmetric and diverse spatial coverage across frequency bands.

\section{Conclusion}
In this paper, we present \textbf{GridPE}, a novel positional embedding method inspired by grid cells, tailored for spaces with Euclidean translation invariance. GridPE encodes relative relationships in a metric space via absolute position embeddings in arbitrary dimensions, making it well-suited for tasks involving spatial and temporal structure. Theoretical analysis shows the method's broad applicability, and experiments on high-dimensional classification tasks demonstrate its competitive accuracy and strong extrapolation. While GridPE is effective in Euclidean domains, it is not directly applicable to non-Euclidean or topologically irregular structures. Nevertheless, it provides a scalable framework connecting neuroscience-inspired representations with practical machine learning applications.

\clearpage
{
\bibliographystyle{unsrtnat}
\bibliography{references}
}

\appendix
\section{Limitations and Future Work}
\label{sec:limit_work}
\paragraph{Limitations.} While GridPE shows strong performance across structured Euclidean tasks, several limitations remain:
\begin{enumerate}[leftmargin=1.5em]
    \item \textbf{Geometric generality.} The current formulation assumes a fixed, simplex-based basis configuration in Euclidean space, which may not generalize effectively to topologically irregular domains such as graphs or curved manifolds.
    
    \item \textbf{Empirical coverage.} All experiments are conducted on classification tasks within standard Euclidean domains (e.g., images and point clouds). Although GridPE theoretically supports spatial encoding in arbitrary dimensions via Fourier basis superposition, we have not yet empirically validated this potential in more complex scenarios such as video, spatio-temporal data, or multi-modal tasks, primarily due to computational constraints.
\end{enumerate}
\paragraph{Future Work.} Several promising directions can be explored to extend the applicability and impact of GridPE:
\begin{enumerate}[leftmargin=1.5em]
    \item \textbf{Non-Euclidean extension.} Extending GridPE to irregular or non-Euclidean domains may require manifold-aware frequency design or spectral learning techniques, in order to preserve directional symmetry and resolution consistency when modeling curved or graph-based spatial structures.
    \item \textbf{Cross-domain composition.} Real-world tasks often involve combinations of multiple Euclidean subspaces—for example, 2D visual frames with 1D temporal sequences in video, or two distinct 2D spaces such as geographic context and trajectory patterns in human mobility analysis. These heterogeneous subspaces differ in semantics and structure, posing challenges for unified frequency-based modeling that can capture both intra-space patterns and inter-space interactions.
\end{enumerate}
Overall, we believe that GridPE provides a principled foundation for positional embedding in continuous and structured domains. With further development, it may serve as a stepping stone toward more generalizable and interpretable spatial representations—ultimately contributing to cognitive map models of how agents perceive and internalize the structure of the external world.

\section{Derivation of Wave Vectors Based on Regular Simplex}
\label{appendix:gridpe_cal}
This section aims to construct a regular simplex in the $n$-dimensional hyperplane, starting from the standard orthogonal basis in $\mathbb{R}^{n+1}$. We then derive the dimensionality reduction method for its wave vectors while maintaining the geometric properties of the regular simplex.

In $\mathbb{R}^{n+1}$, the standard orthogonal basis consists of vectors $\mathbf{e}_1, \mathbf{e}_2, \dots, \mathbf{e}_{n+1}$, where each basis vector $\mathbf{e}_i$ is defined as:
\begin{equation}
\mathbf{e}_i = (0,0,\dots,1,\dots,0)
\end{equation}
with the 1 appearing in the $i$-th coordinate and 0 elsewhere. These vectors form $n{+}1$ linearly independent points in $\mathbb{R}^{n+1}$, defining the vertices of a simplex. However, they do not yet satisfy the geometric conditions of a regular simplex: their centroid does not lie at the origin, and they span the full $(n{+}1)$-dimensional space rather than an $n$-dimensional hyperplane.

To construct a regular simplex, we first translate the standard basis vectors so that their centroid is positioned at the origin. This is achieved by introducing an offset vector:
\begin{equation}
\mathbf{L} = (1, 1, \dots, 1) \in\mathbb{R}^{ n+1}
\end{equation}
The translated points are then defined as:
\begin{equation}
\mathbf{v}_i = \mathbf{e}_i - \frac{1}{n+1} \mathbf{L}
\end{equation}
This translation ensures the centroid lies at the origin and the vectors lie on the $n$-dimensional hyperplane defined by:
\begin{equation}
\sum_{i=1}^{n+1} x_i = 0
\end{equation}
which corresponds to the hyperplane orthogonal to the normal vector \(\boldsymbol{\nu} = (1, 1, \dots, 1)\). We now reinterpret the translated points \(\mathbf{v}_i\) as frequency vectors and denote them by \(\boldsymbol{\omega}_i\), i.e., \(\boldsymbol{\omega}_i := \mathbf{v}_i\). These vectors emanate from the origin, lie on the same hyperplane, and together form a regular simplex centered at the origin.
\begin{equation}
\boldsymbol{\omega}_i = \mathbf{e}_i - \frac{1}{n+1} \mathbf{L}
\end{equation}

Thus, the translated vectors \(\boldsymbol{\omega}_i\) are linearly independent, centered at the origin, and lie in the same hyperplane, forming a regular simplex. These vectors live in \(\mathbb{R}^{n+1}\) and are used to construct the wave vector matrix \(\boldsymbol{\Omega}^{(n+1)}\):
\begin{equation}
\boldsymbol{\Omega}^{(n+1)} = 
\begin{bmatrix}
\omega_{11} & \cdots & \omega_{1,n+1} \\
\vdots & \ddots & \vdots \\
\omega_{n+1,1} & \cdots & \omega_{n+1,n+1}
\end{bmatrix}
\end{equation}
To reduce these $(n{+}1)$-dimensional wave vectors to $n$ dimensions while preserving their geometric structure, we perform singular value decomposition (SVD) on the wave vector matrix:
\begin{equation}
\boldsymbol{\Omega}^{(n+1)} = \mathbf{U} \boldsymbol{\Sigma} \mathbf{W}^\top
\end{equation}
Here, $\mathbf{U} \in \mathbb{R}^{(n+1) \times (n+1)}$, $\boldsymbol{\Sigma}$ is diagonal, and $\mathbf{W}$ is the right singular matrix. We take the first $n$ columns of $\mathbf{U}$:
\begin{equation}
\mathbf{U}^{(n)} = [\mathbf{u}_1, \dots, \mathbf{u}_n]
\end{equation}
which form an orthonormal basis for the hyperplane. We then project the original $(n{+}1)$-dimensional wave vectors onto this subspace to obtain the reduced matrix:
\begin{equation}
\boldsymbol{\Omega}^{(n)} = \boldsymbol{\Omega}^{(n+1)} \mathbf{U}^{(n)}
\end{equation}
This guarantees that the resulting $n$-dimensional wave vectors preserve both magnitude and angular relationships, enabling isotropic and stable spatial representation. Here, \(\boldsymbol{\Omega}^{(n)} \in \mathbb{R}^{(n+1) \times n}\) represents the set of projected wave vectors in \(n\)-dimensional space, reduced from the original \((n{+}1)\)-dimensional simplex construction.

\section{Optimal Ratio of Scale under Economy Principle}
\label{appendix:grid_scale}
Neuroscience evidence indicates that the medial entorhinal cortex consists of several modules. 
Within each module, the direction and period of grid cells remain consistent, 
but the scales gradually decrease along the ventral-dorsal axis~\cite{stensola2012entorhinal}, 
transitioning from coarse, low-frequency to granular, high-frequency spatial information. 
This hierarchical spatial organization reduces the number of required grid cells 
while maintaining fine spatial resolution, embodying the economy principle~\cite{wei2015principle}.

The challenge is to determine the optimal scale ratio that achieves fine resolution with fewer grid cells, similar to minimizing parameters in positional embeddings for neural networks. The relationship between scale scales and wave vectors mirrors the frequency in Transformers, and insights from grid cell theory can help optimize positional embeddings for efficient spatial resolution. Recall that in the original Transformer positional encoding~\cite{vaswani2017attention}, spatial frequency is defined as \(\omega_k = 1 / 10000^{2k/d}\), where \( d \) represents the dimensionality of the embedding space, forming a geometric series of wave lengths. The rationale behind this choice remains unclear, and we anticipate that computational theories of grid cells can provide insights into frequency determination.

We provide a generalized derivation based on the proof in~\cite{wei2015principle}, aiming to represent \( n \)-dimensional Euclidean space with the least number of grid cells, in line with the principle of economy. Suppose there are \( m \) discrete grid scales. The period of the \( i \)-th module layer (i.e., the distance between adjacent grid vertices) is denoted by \( \lambda_i \), and the period increases with \( i \). Each grid field is modeled as a hyperspherical region of diameter \( l_i \), centered at a grid vertex.

\begin{figure}[htbp]
    \centering
    \includegraphics[width=0.7\linewidth]{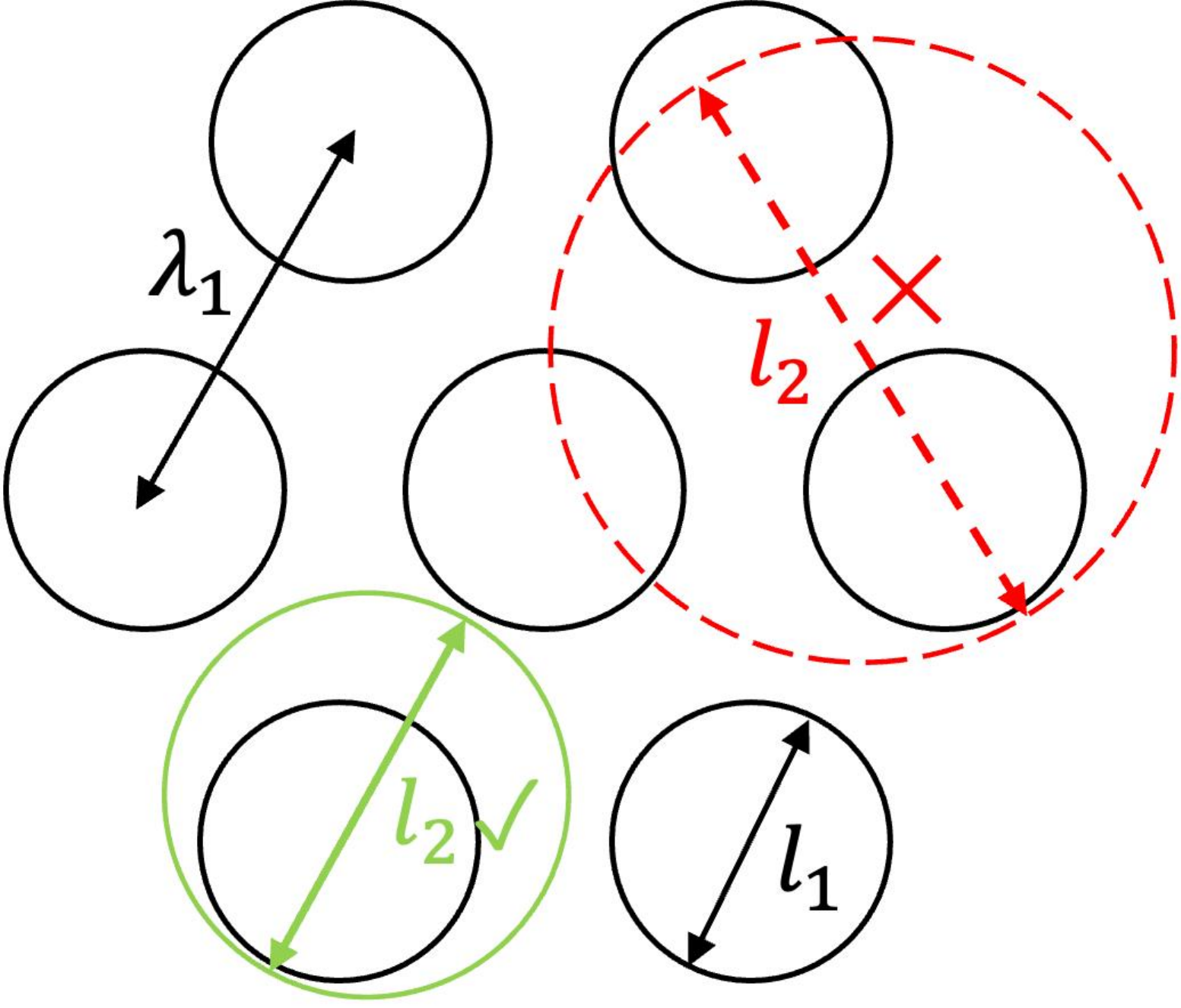}
    \caption{Illustration of the relationship between firing field diameter and periodicity. For two grid cells with adjacent discrete scales, the periodicity of the smaller scale must be less than the diameter of the larger one; otherwise, spatial disambiguation is not guaranteed.}
    \label{fig:grid_scale_overlap}
\end{figure}

In 2D space, this corresponds to a circular region activating a grid cell. To ensure unambiguous spatial representation, the periodicity of the finer scale must not exceed the spatial span of the coarser ones. This geometric relationship is illustrated in Figure~\ref{fig:grid_scale_overlap}, where the firing field diameter \( l_i \) and period \( \lambda_i \) of adjacent modules are shown in two dimensions.
The scale ratio of adjacent layers is defined as \( r_i = \lambda_{i+1} / \lambda_i > 1 \), and for the first layer we let \( r_1 = \lambda_1 / l_1 \). To cover the entire space without gaps, multiple grid cells with different phases are required. The larger the ratio \( \lambda_i / l_i \), the more grid cells are needed. If at least \( d \) cells must correspond to a spatial location, the total number of grid cells across all layers is denoted as \( N \):
\begin{equation}
    N = \sum_{i=1}^{m} d \left( \frac{\lambda_i}{l_i} \right)^n
    \label{eq:total_grid_cells}
\end{equation}

Additionally, the condition \( l_i \leq \lambda_{i-1} \) must be satisfied. Otherwise, the grid cell system cannot accurately determine the current location when cells from both layer \( i \) and \( i+1 \) fire simultaneously. This is because the larger firing field contains more than one smaller field, as shown in Figure~\ref{fig:grid_scale_overlap}. The condition leads to the inequality \( l_i \leq \lambda_i / r \). If the grid cell system is required to represent the space at a given constant spatial resolution \( R \), that is, we predefined the measurement relationship between the largest scale and the smallest firing field unit:
\begin{equation}
R = \left( \frac{\lambda_m}{l_1} \right)^n 
= \left( \prod_{i=1}^{m-1} \frac{\lambda_{i+1}}{\lambda_i} \frac{\lambda_1}{l_1} \right)^n 
= (r^n)^m
\label{eq:resolution_def}
\end{equation}

Here we set \( \lambda_0 = l_1 \). By denoting \( r^n = \rho \), we obtain:
\begin{equation}
N = d \sum_{i=1}^{m} \left( \frac{\lambda_i}{l_i} \right)^n 
\geq d m r^n = d \rho \log_{\rho} R
\label{eq:n_lower_bound}
\end{equation}

Differentiating this expression, we find that the minimal number of grid cells is achieved when \( \rho = e \), which implies \( r = \sqrt[n]{e} \). This value defines the maximum allowable ratio between adjacent scales for efficient spatial tiling for \( n \)-dimensional space. Based on this, we derive a constraint relating the frequency base and the encoding dimension \( d \) for GridPE. Let \( M \) denote the number of Fourier bases per scale. The scale ratio induced by the hierarchical representation is:
\begin{equation}
\mathcal{R} = \frac{\text{base}^{\frac{2M(k+1)}{d}}}{\text{base}^{\frac{2Mk}{d}}} = \text{base}^{\frac{2M}{d}}.
\end{equation}

To ensure this ratio does not exceed \(\sqrt[n]{e}\) and thus avoid overlap or gaps between scales, we require:
\begin{equation}
\text{base}^{\frac{2M}{d}} \leq \sqrt[n]{e} \quad \Rightarrow \quad \text{base} \leq e^{\frac{d}{2Mn}}.
\end{equation}

This establishes a theoretical upper bound on the scale base as a function of the embedding dimension \( d \), the number of Fourier bases \( M \), and the spatial dimension \( n \).

\section{Additional Experimental Results}
\subsection{Extrapolation Performance under Varying Input Size}
\label{appendix:accuracy_trends}
\begin{figure}[htbp]
    \centering
    \begin{subfigure}{0.48\linewidth}
        \centering
        \includegraphics[width=\linewidth]{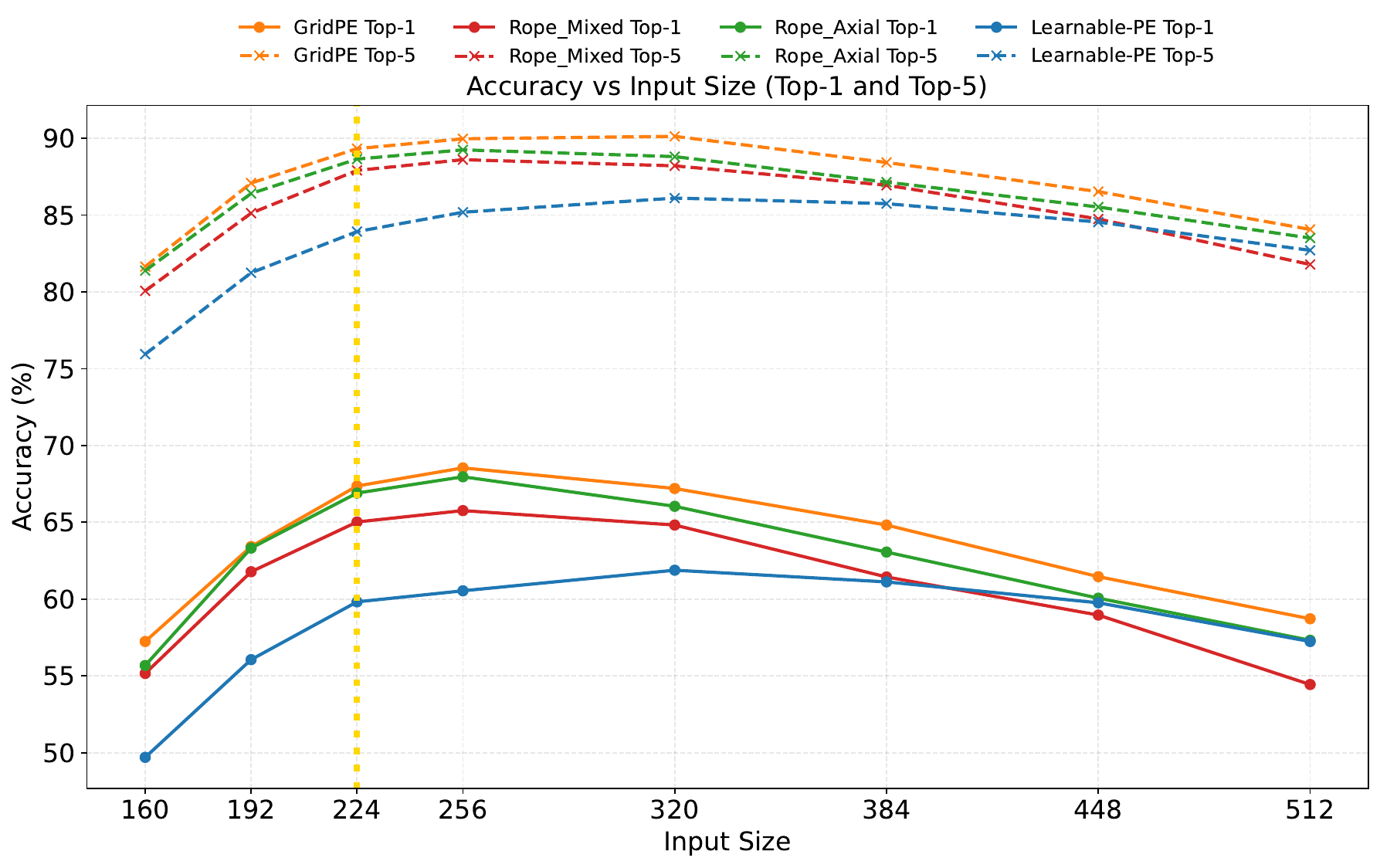}
        \caption{2D – Accuracy at 224 input size}
        \label{fig:2d_224}
    \end{subfigure}
    \hfill
    \begin{subfigure}{0.48\linewidth}
        \centering
        \includegraphics[width=\linewidth]{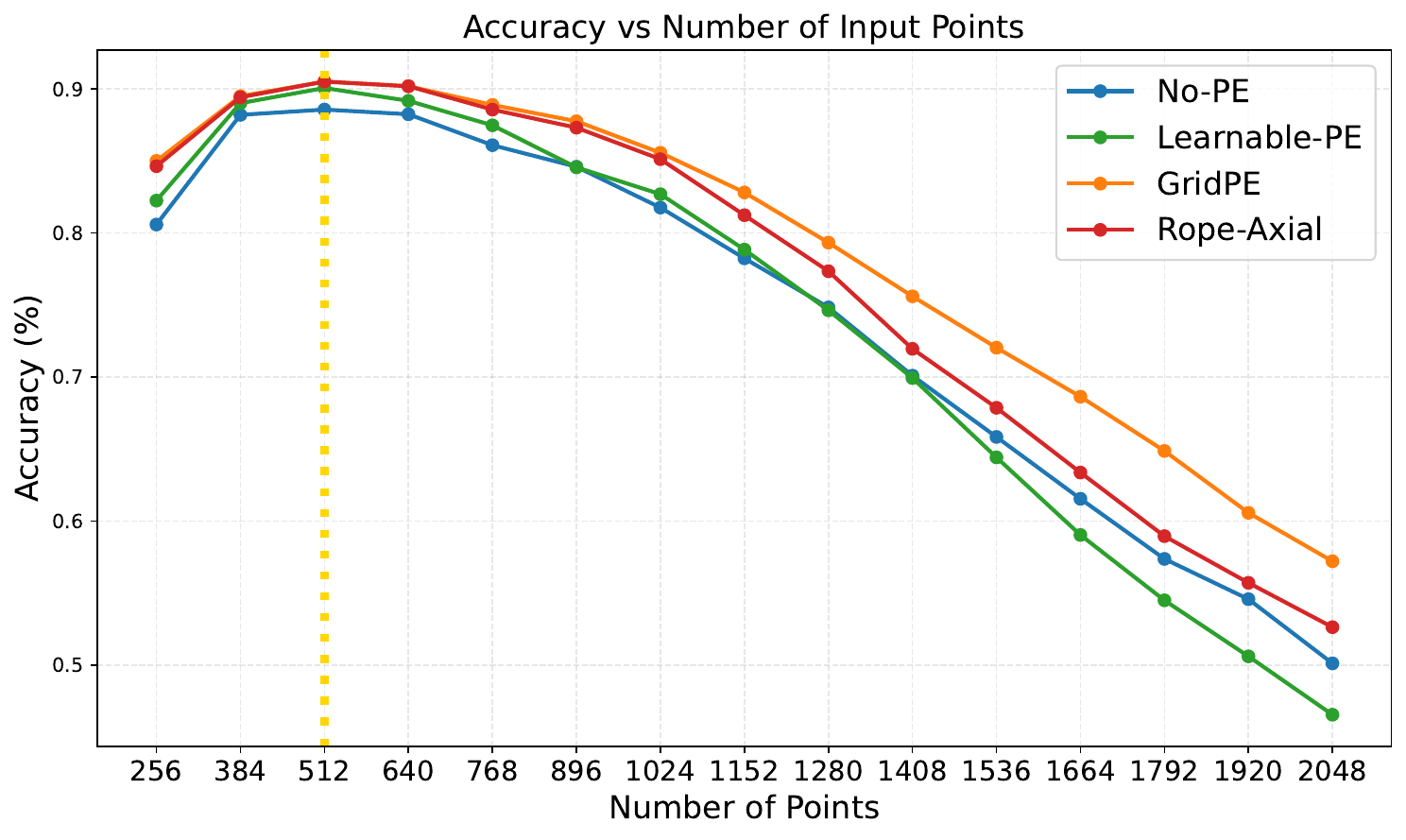}
        \caption{3D – Accuracy at 512 input points}
        \label{fig:3d_512}
    \end{subfigure}
    \begin{subfigure}{0.48\linewidth}
        \centering
        \includegraphics[width=\linewidth]{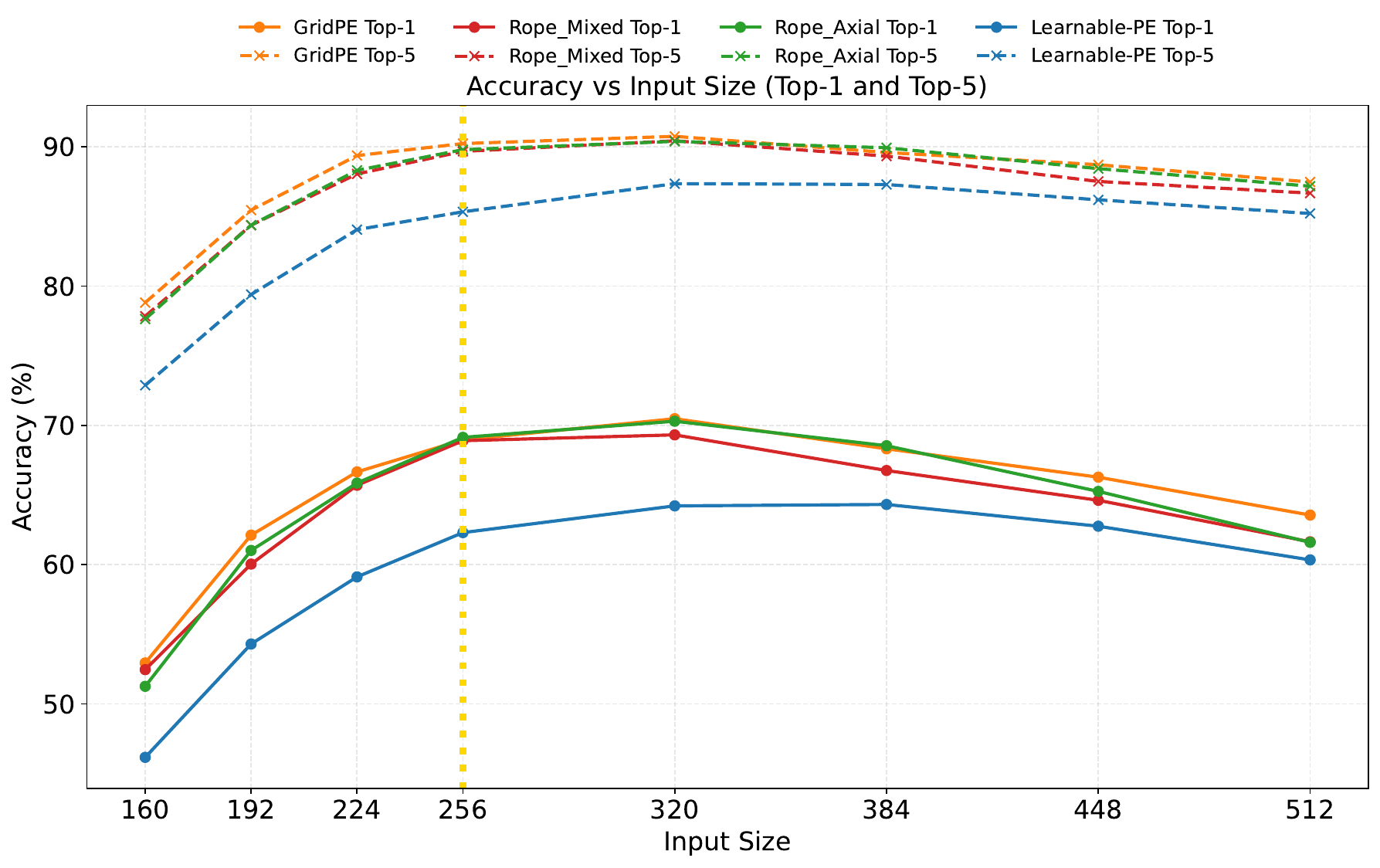}
        \caption{2D – Accuracy at 256 input size}
        \label{fig:2d_256}
    \end{subfigure}
    \hfill
    \begin{subfigure}{0.48\linewidth}
        \centering
        \includegraphics[width=\linewidth]{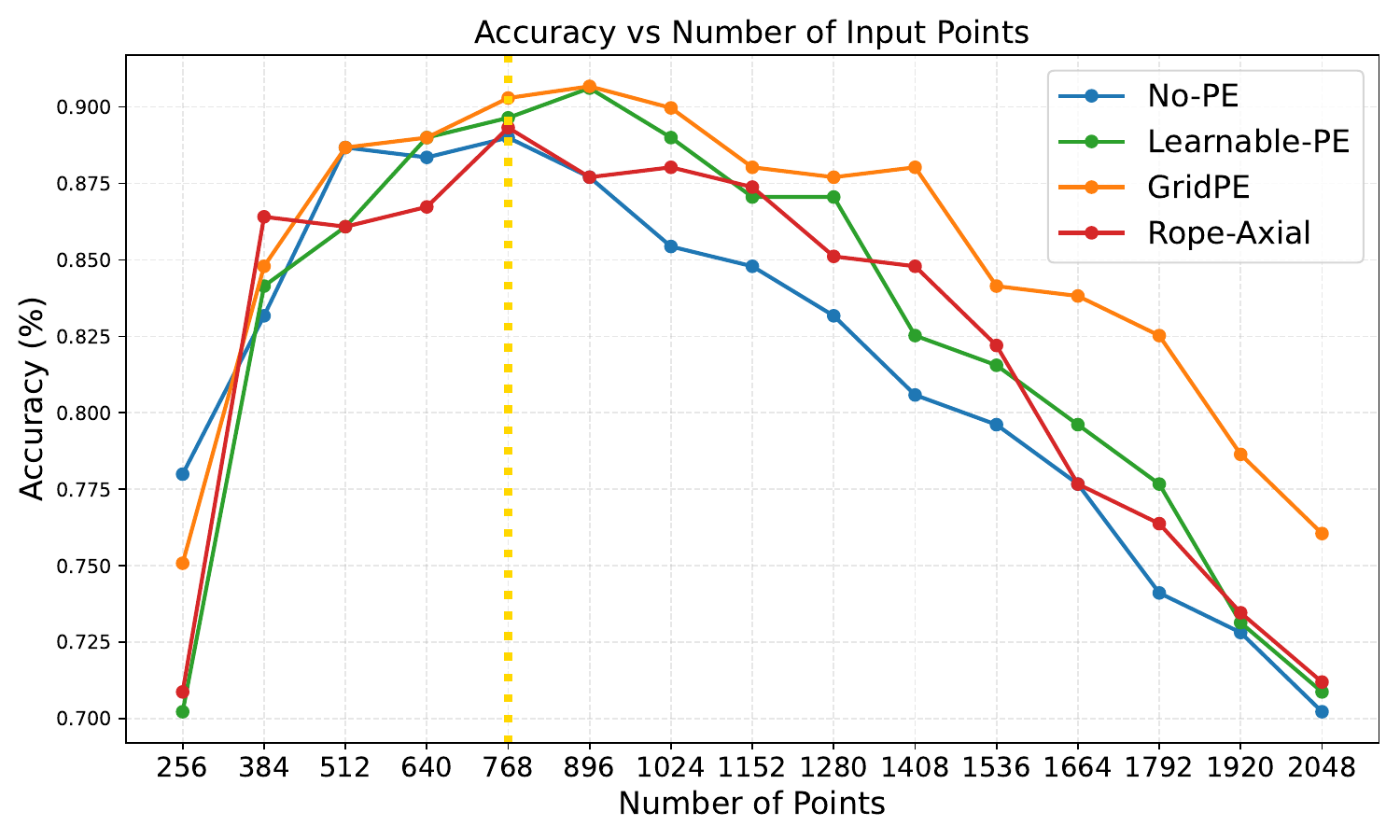}
        \caption{3D – Accuracy at 768 input points}
        \label{fig:3d_768}
    \end{subfigure}
    \begin{subfigure}{0.48\linewidth}
        \centering
        \includegraphics[width=\linewidth]{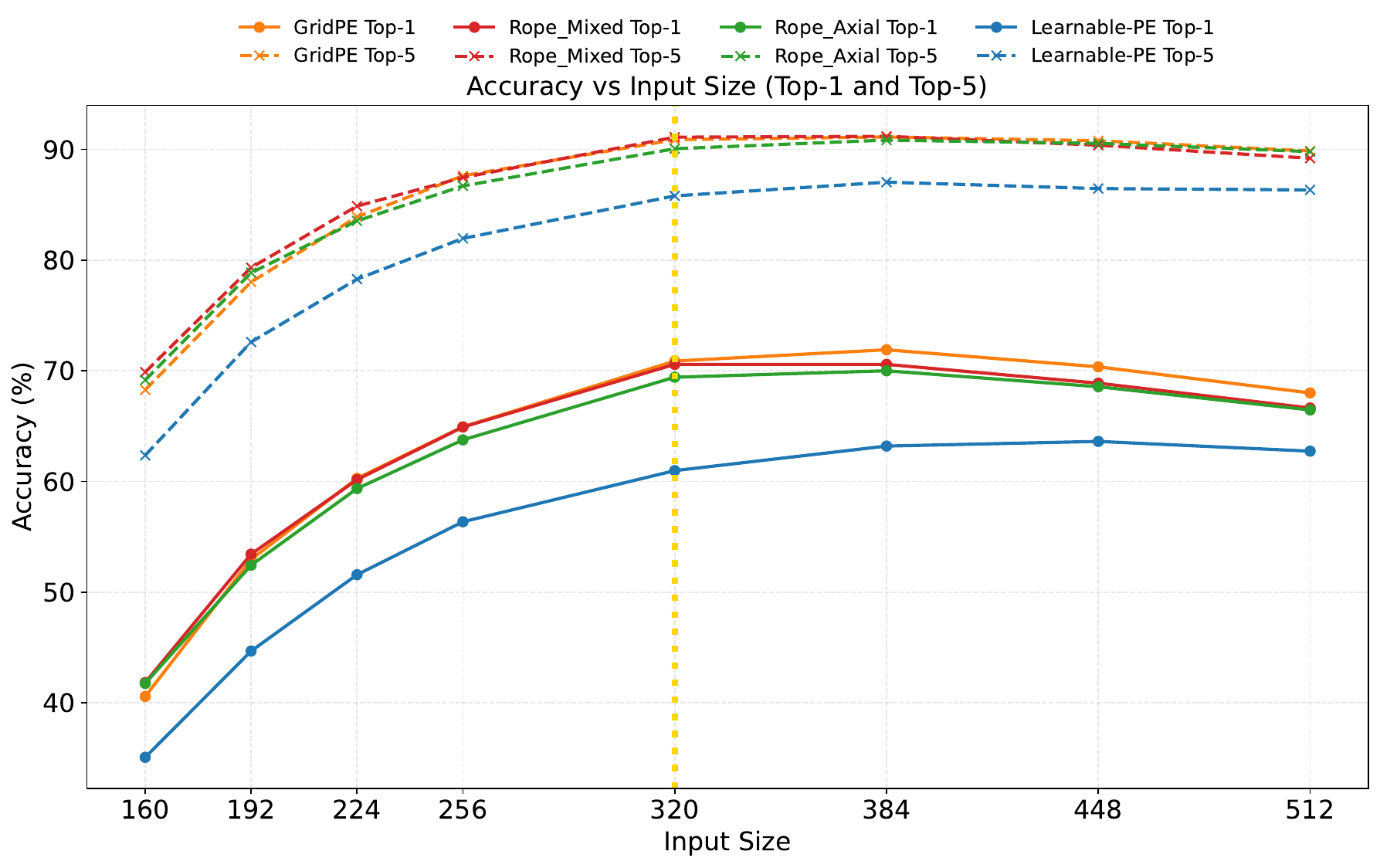}
        \caption{2D – Accuracy at 320 input size}
        \label{fig:2d_320}
    \end{subfigure}
    \hfill
    \begin{subfigure}{0.48\linewidth}
        \centering
        \includegraphics[width=\linewidth]{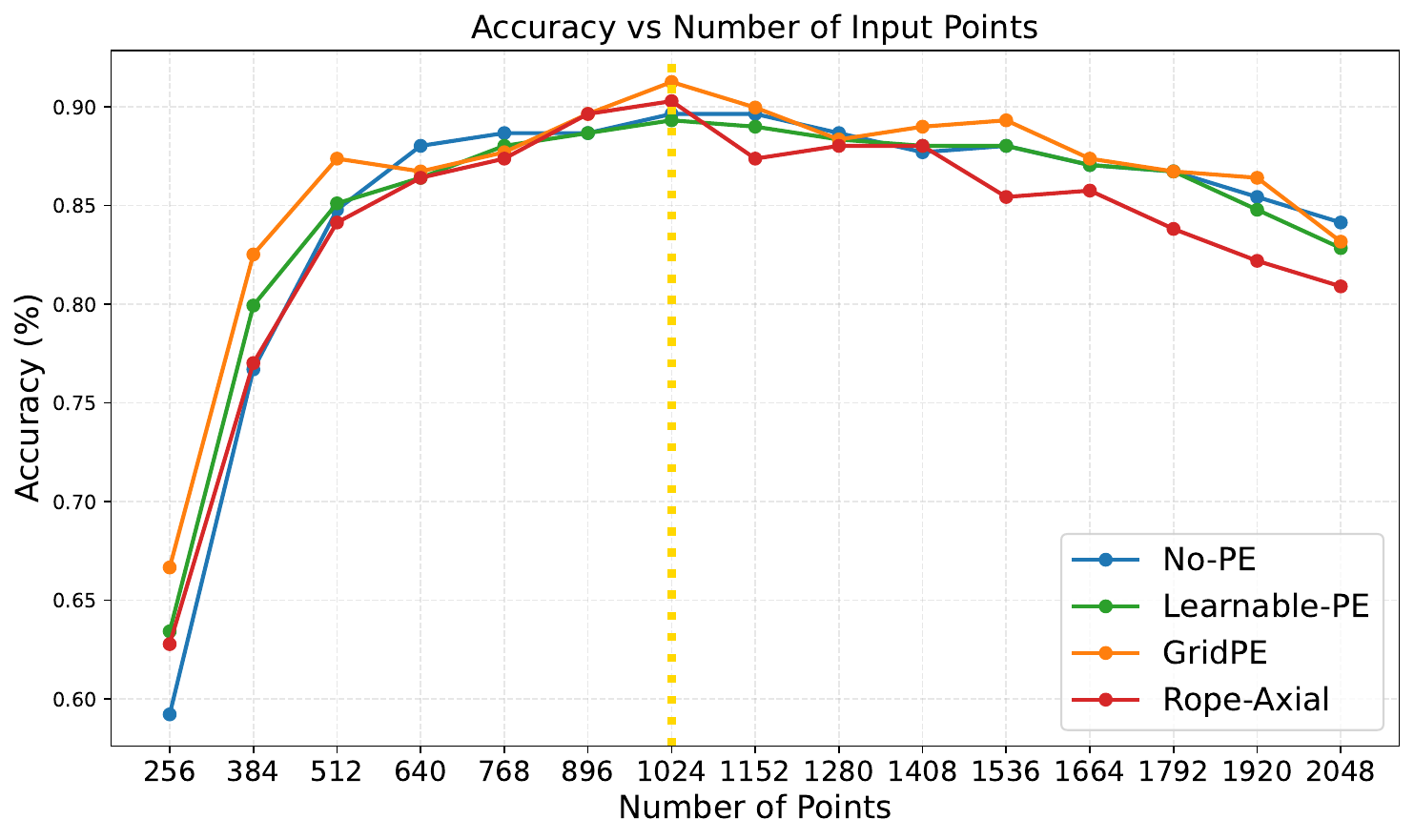}
        \caption{3D – Accuracy at 1024 input points}
        \label{fig:3d_1024}
    \end{subfigure}
    \caption{Comparison of accuracy (\%) using different positional embedding methods under various input sizes for 2D (ViT-S) and 3D (PCT) tasks.}
    \label{fig:2d3d_inputsize_comparison}
\end{figure}

Figure~\ref{fig:2d_224}–\ref{fig:2d_320} report the Top‑1/Top‑5 accuracy of different positional embedding methods on ViT‑S across a range of image resolutions, while Figure~\ref{fig:3d_512}–\ref{fig:3d_1024} present the Top-1 accuracy of PCT models with varying numbers of 3D input points. Together, these subfigures provide a comparative view of how positional embedding methods perform across input scales in both 2D and 3D tasks. Covering diverse training input sizes, the subfigures enable analysis of how the initial resolution influences model robustness under interpolation and extrapolation scenarios. Notably, the relative performance gains of GridPE tend to decrease as the training input size increases—possibly because larger initial resolutions reduce the extent of input-scale discrepancy specifically under extrapolation settings, thereby attenuating the benefits of GridPE’s extrapolation capability.

\subsection{Comparative Study of Positional Embedding Schemes in Attention Maps}
\label{appendix:attn_map}
\begin{figure}[t]
    \centering
    \begin{subfigure}[b]{0.24\linewidth}
    \includegraphics[width=\linewidth]{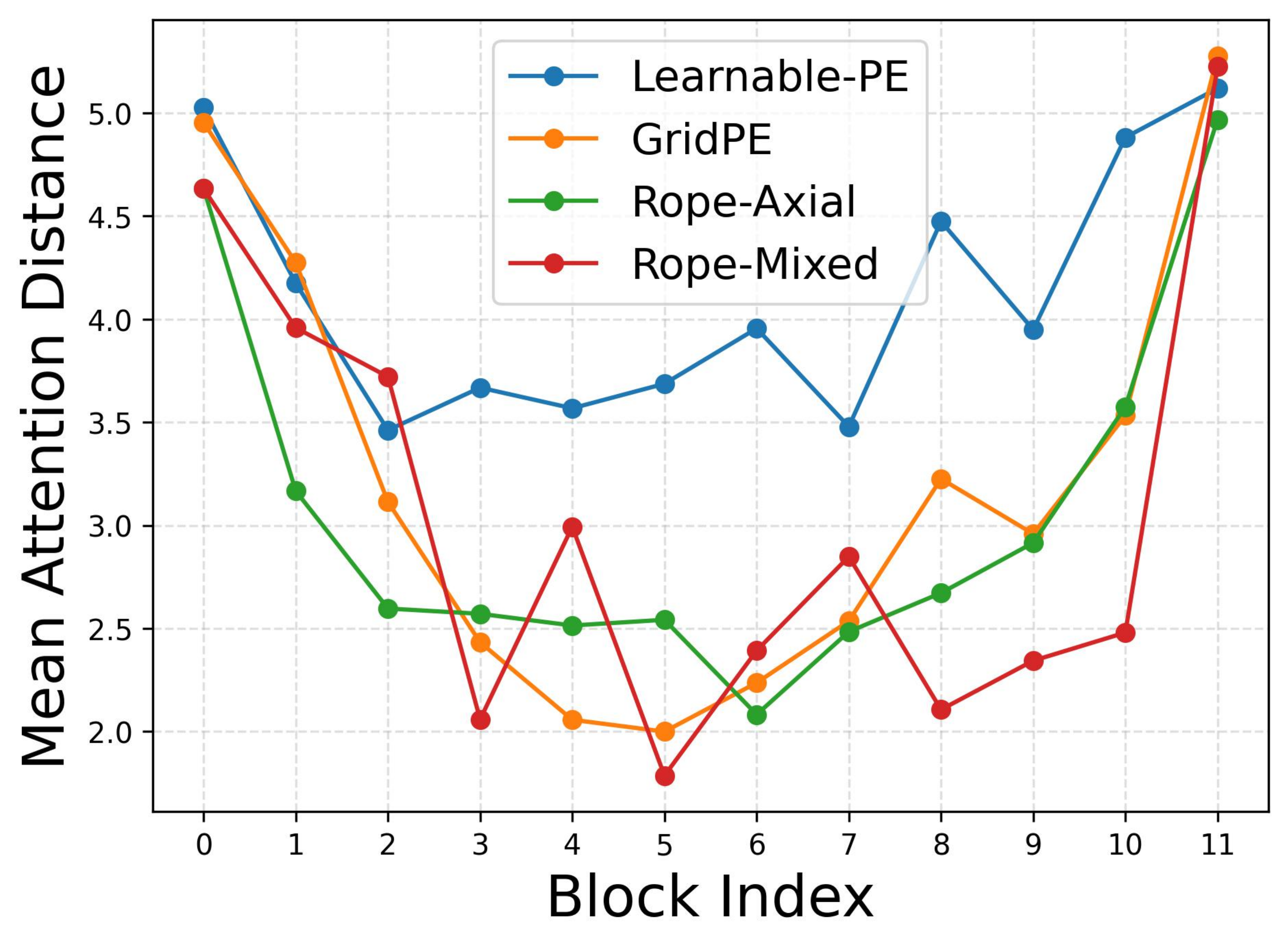}
    \caption{Input size = 160}
    \label{fig:attn_160}
    \end{subfigure}
    \begin{subfigure}[b]{0.24\linewidth}
      \includegraphics[width=\linewidth]{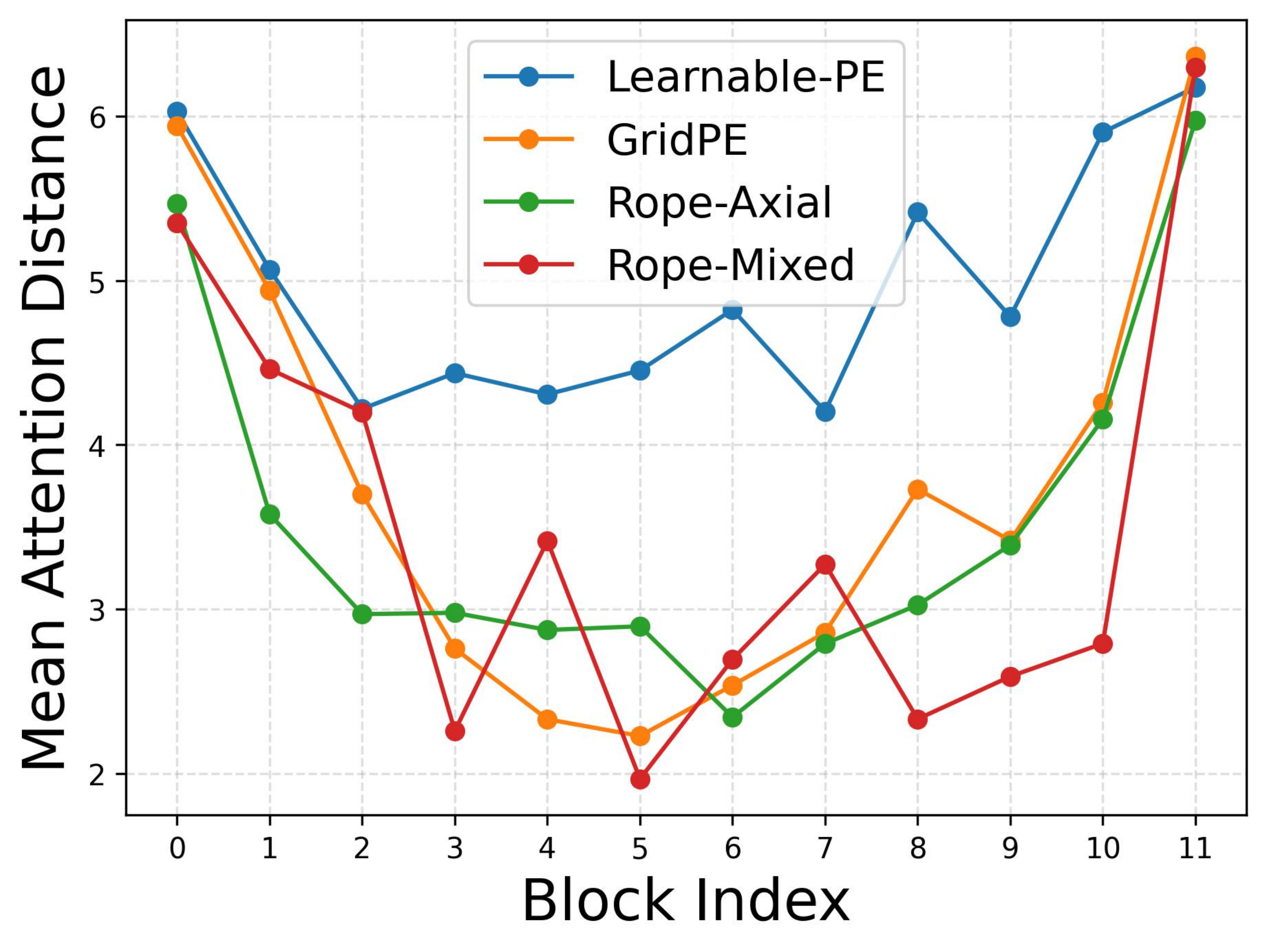}
      \caption{Input size = 192}
      \label{fig:attn_192}
    \end{subfigure}
    \begin{subfigure}[b]{0.24\linewidth}
      \includegraphics[width=\linewidth]{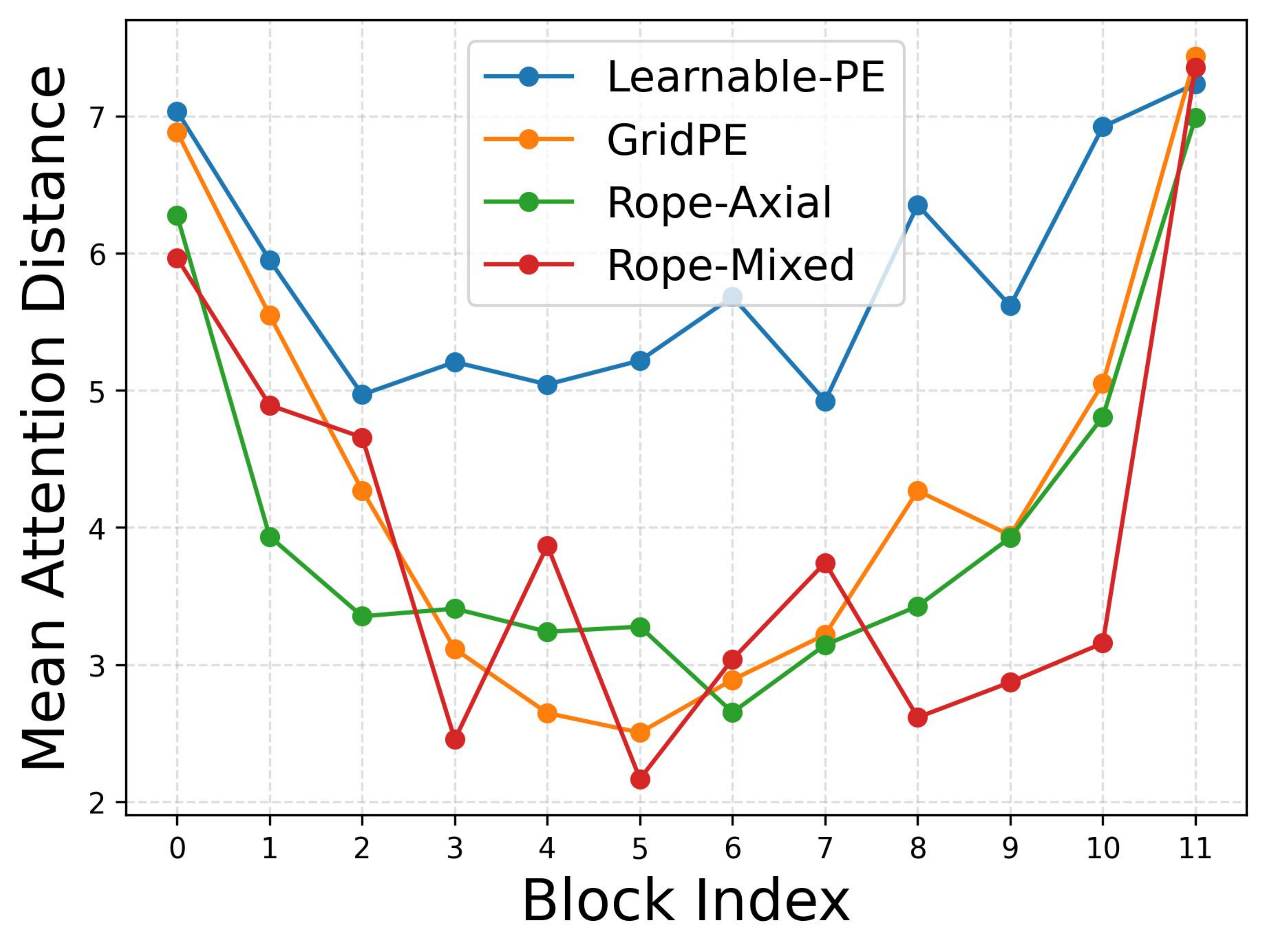}
      \caption{Input size = 224}
      \label{fig:attn_224}
    \end{subfigure}
    \begin{subfigure}[b]{0.24\linewidth}
      \includegraphics[width=\linewidth]{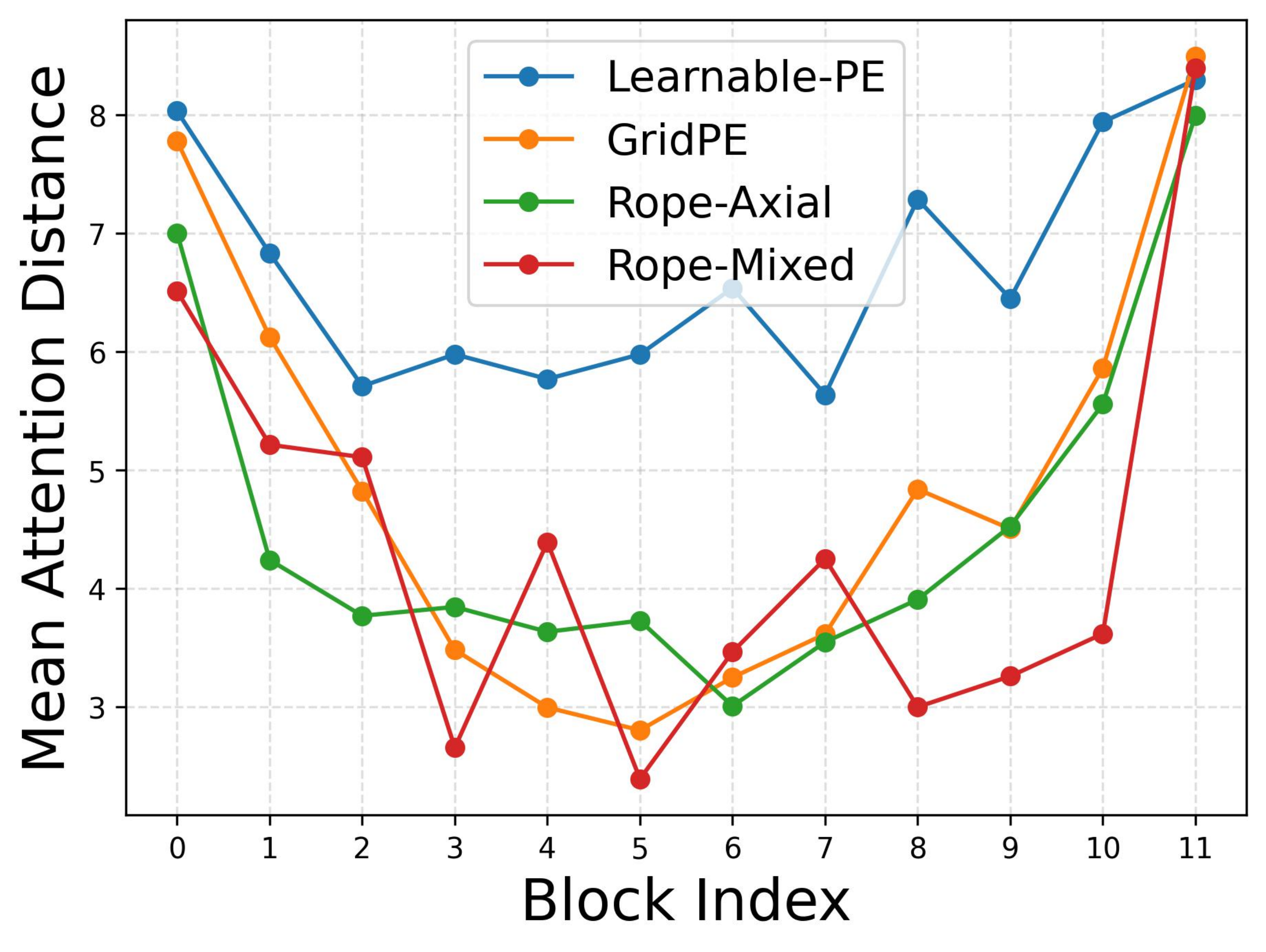}
      \caption{Input size = 256}
      \label{fig:attn_256}
    \end{subfigure}
    \vspace{1ex}
    \begin{subfigure}[b]{0.24\linewidth}
      \includegraphics[width=\linewidth]{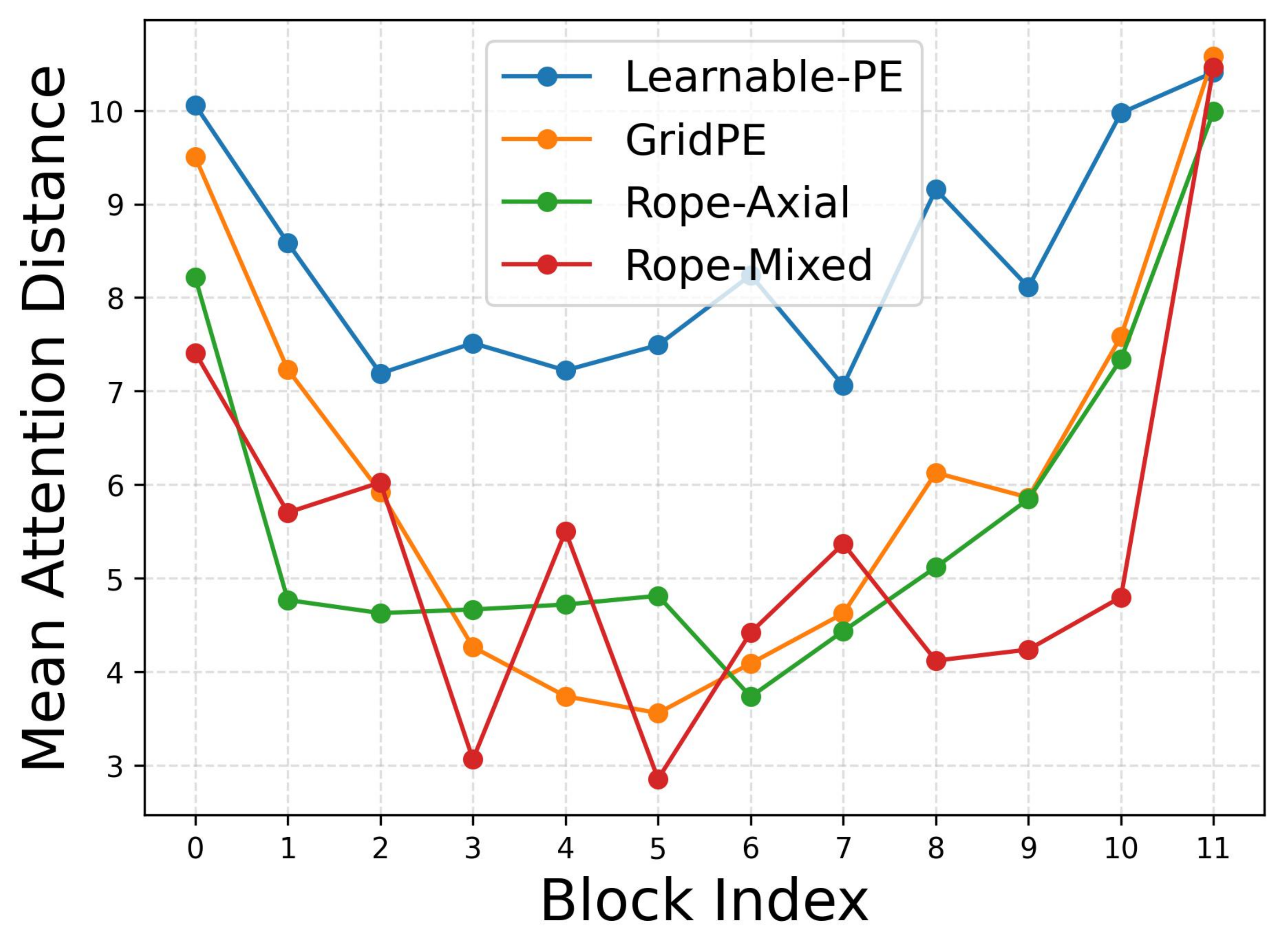}
      \caption{Input size = 320}
      \label{fig:attn_320}
    \end{subfigure}
    \begin{subfigure}[b]{0.24\linewidth}
      \includegraphics[width=\linewidth]{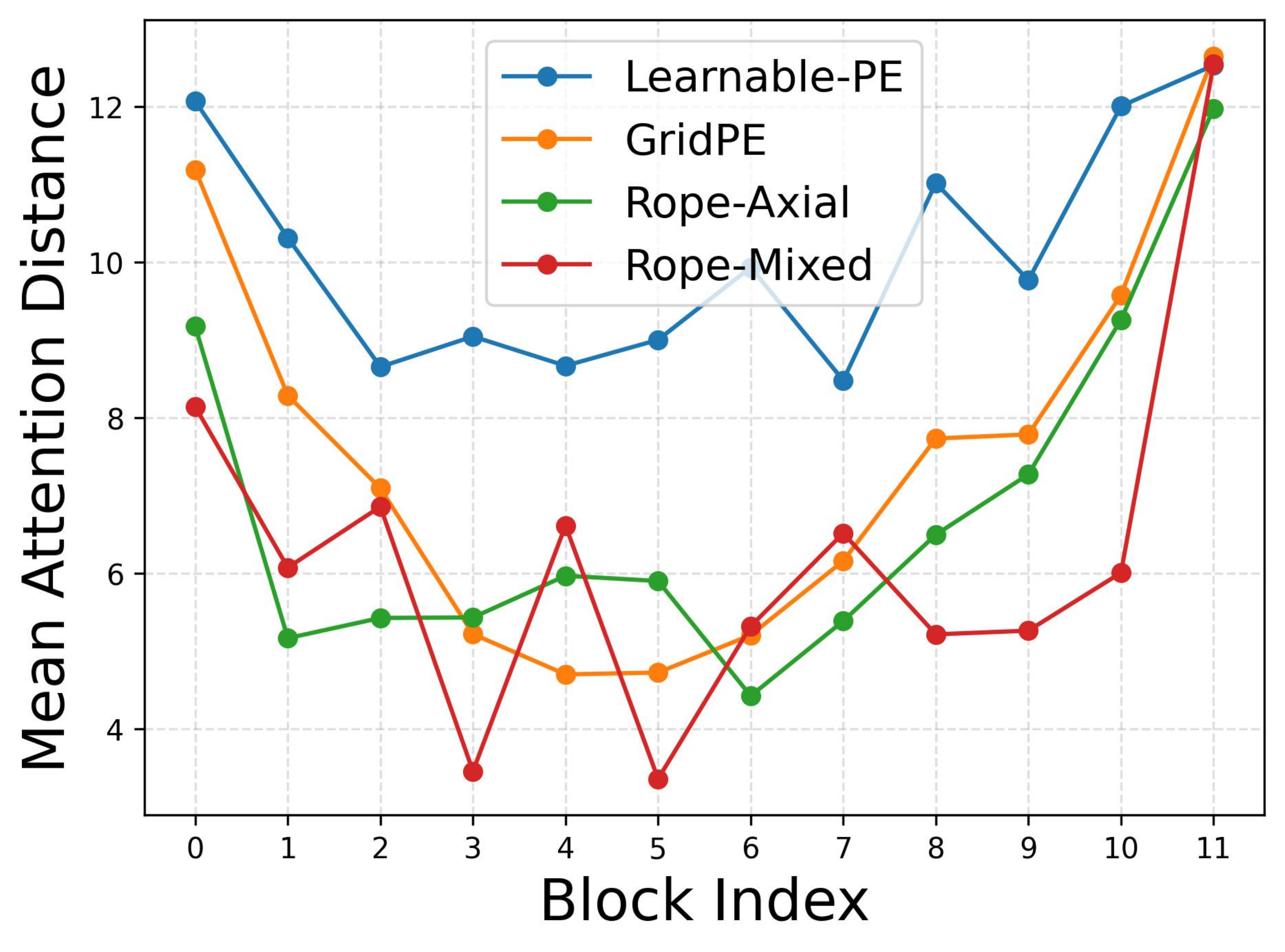}
      \caption{Input size = 384}
      \label{fig:attn_384}
    \end{subfigure}
    \begin{subfigure}[b]{0.24\linewidth}
      \includegraphics[width=\linewidth]{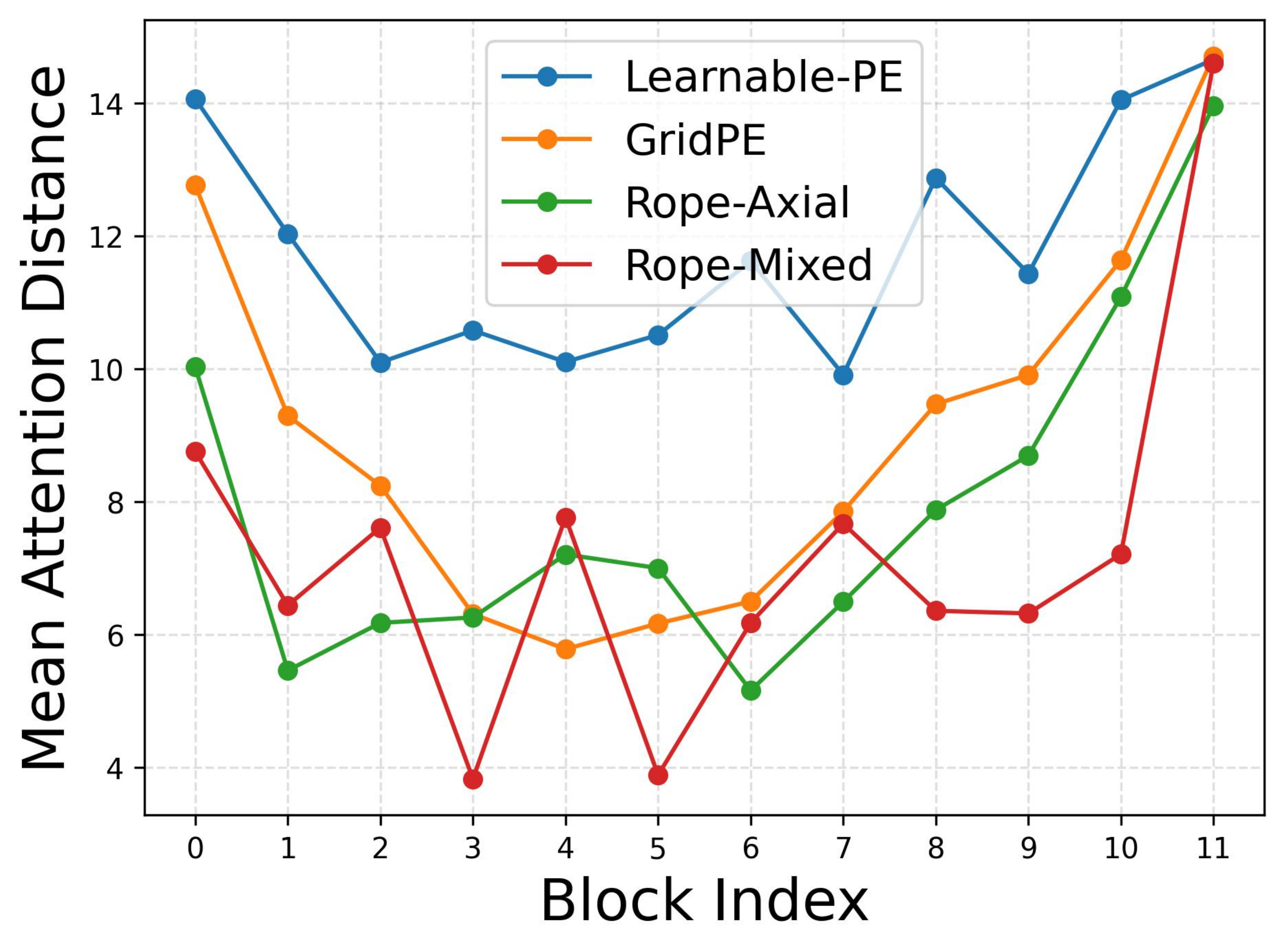}
      \caption{Input size = 448}
      \label{fig:attn_448}
    \end{subfigure}
    \begin{subfigure}[b]{0.24\linewidth}
      \includegraphics[width=\linewidth]{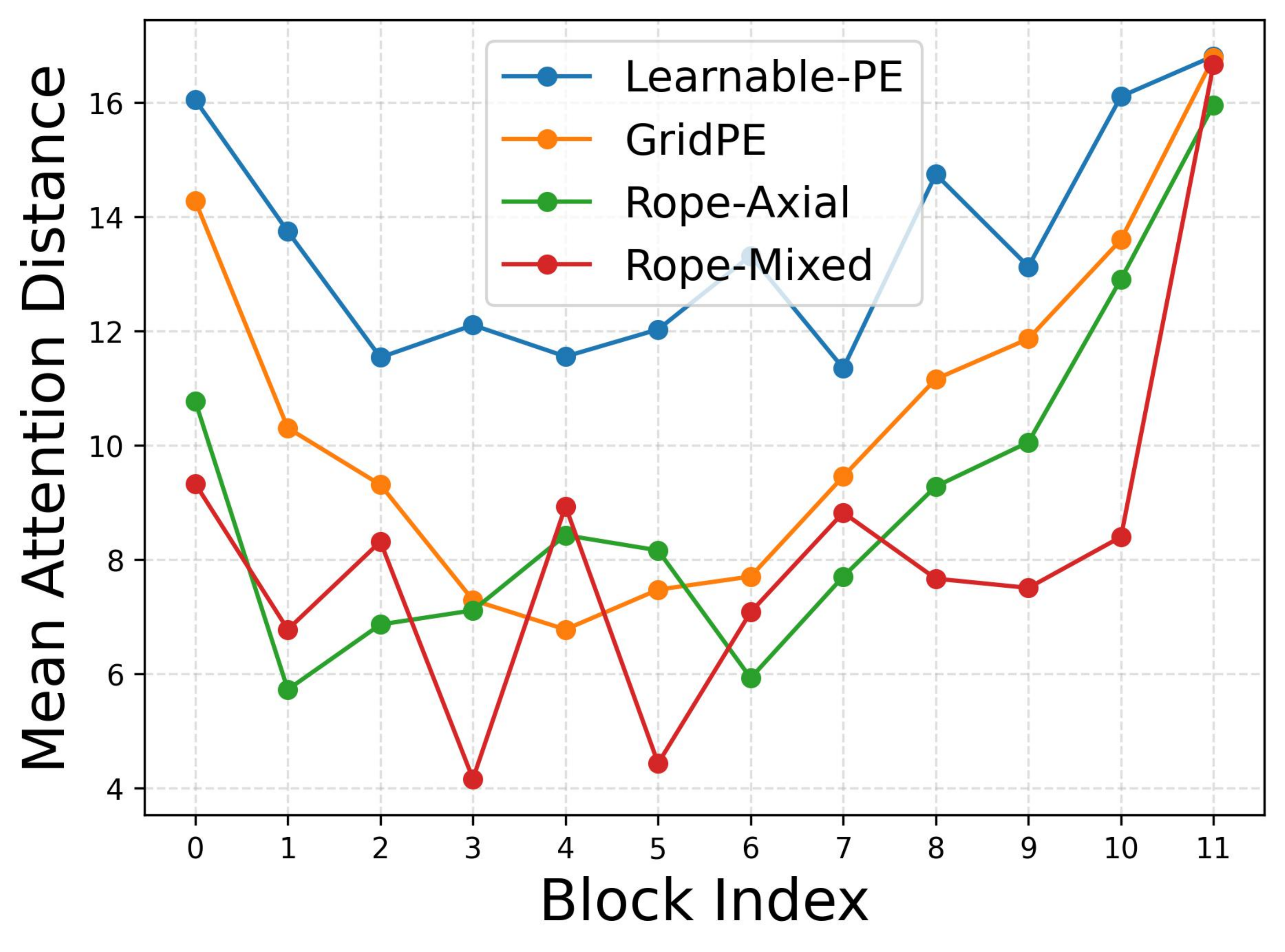}
      \caption{Input size = 512}
      \label{fig:attn_512}
    \end{subfigure}
  
    \caption{Mean attention distance per block on ViT-S for 2D image classification across input resolutions.}
    \label{fig:attn_all}
\end{figure}

\begin{figure}[t]
    \centering
    \begin{subfigure}[b]{0.24\linewidth}
      \includegraphics[width=\linewidth]{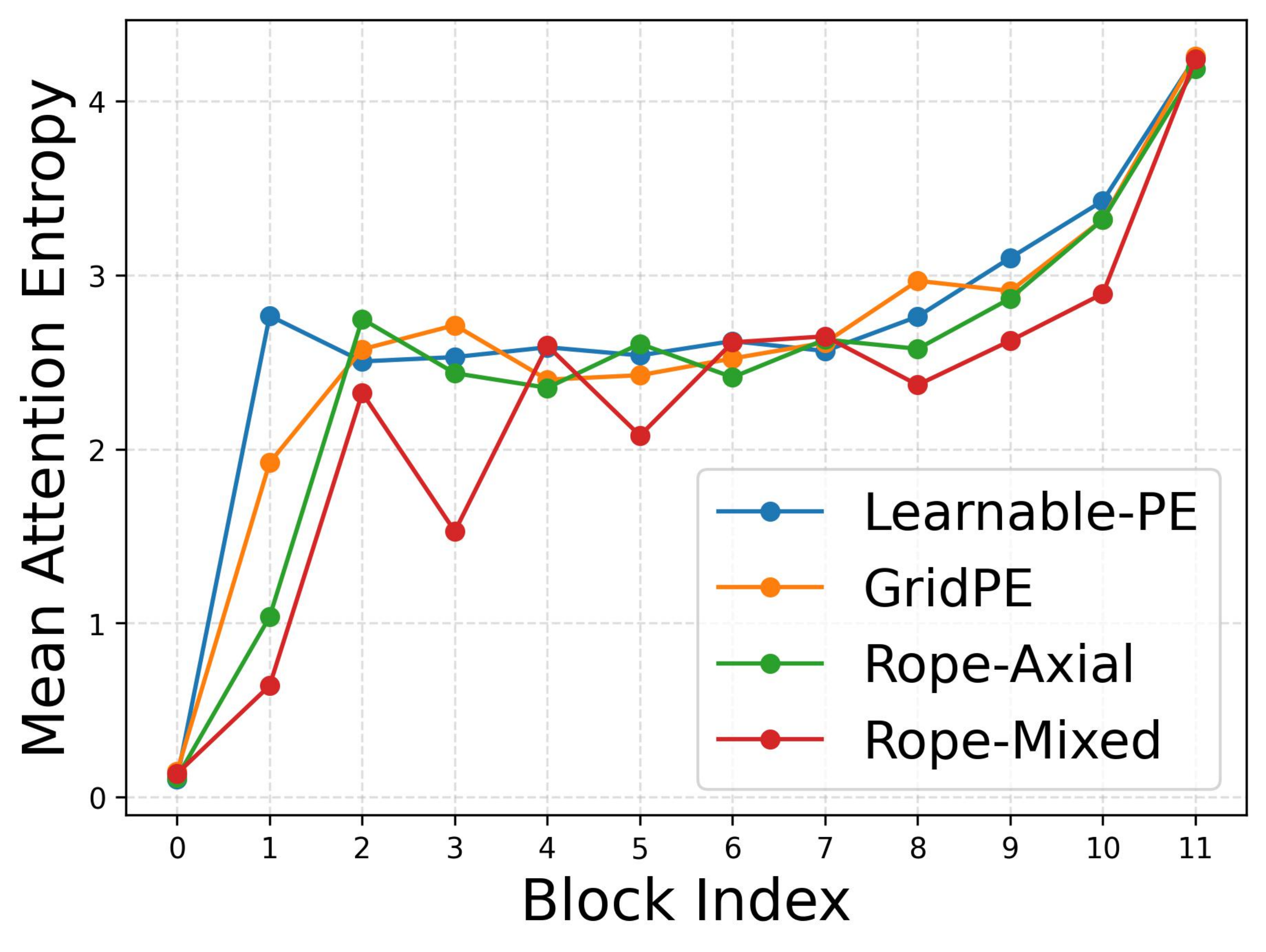}
      \caption{Input size = 160}
      \label{fig:entropy_160}
    \end{subfigure}
    \begin{subfigure}[b]{0.24\linewidth}
      \includegraphics[width=\linewidth]{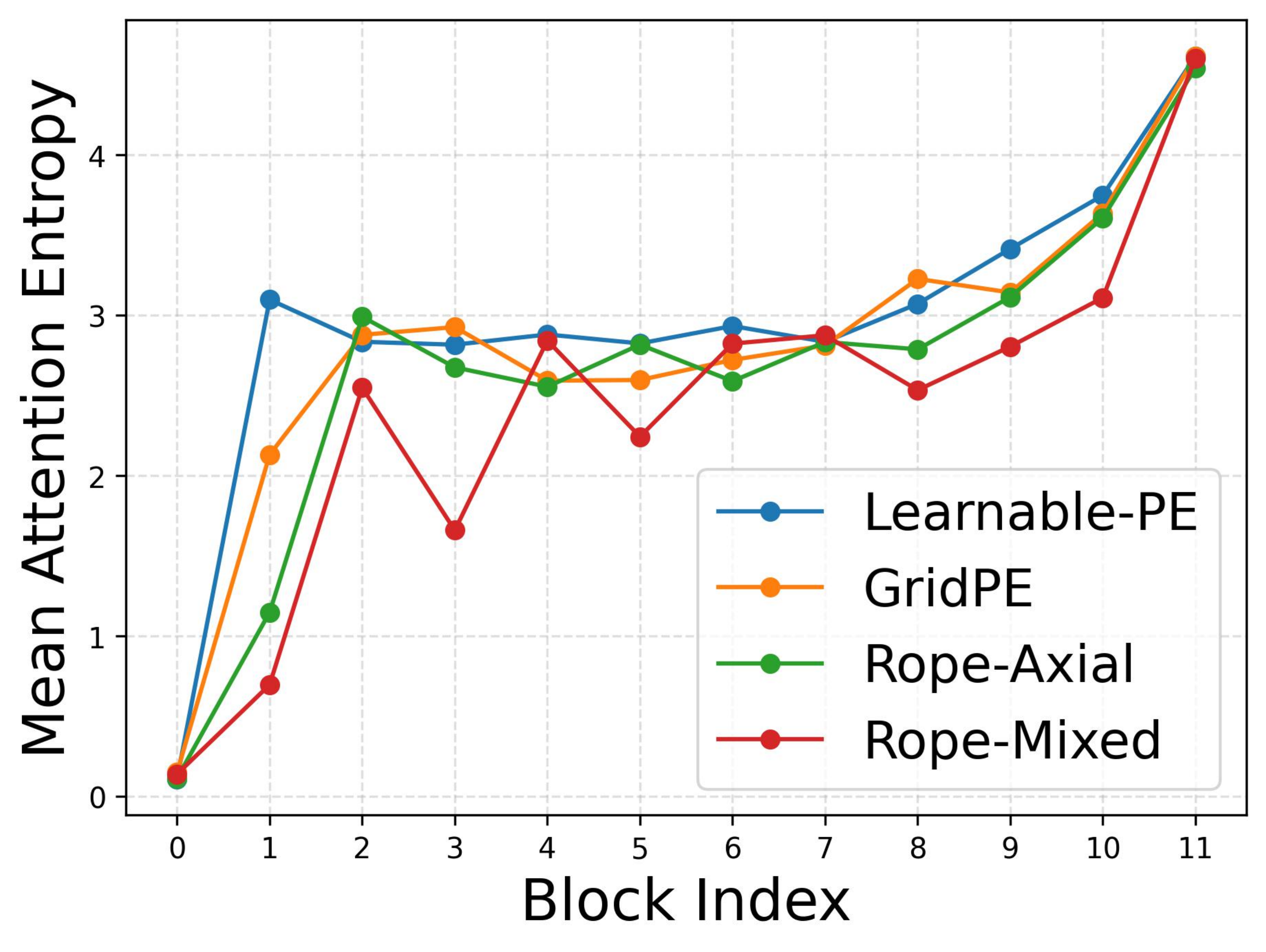}
      \caption{Input size = 192}
      \label{fig:entropy_192}
    \end{subfigure}
    \begin{subfigure}[b]{0.24\linewidth}
      \includegraphics[width=\linewidth]{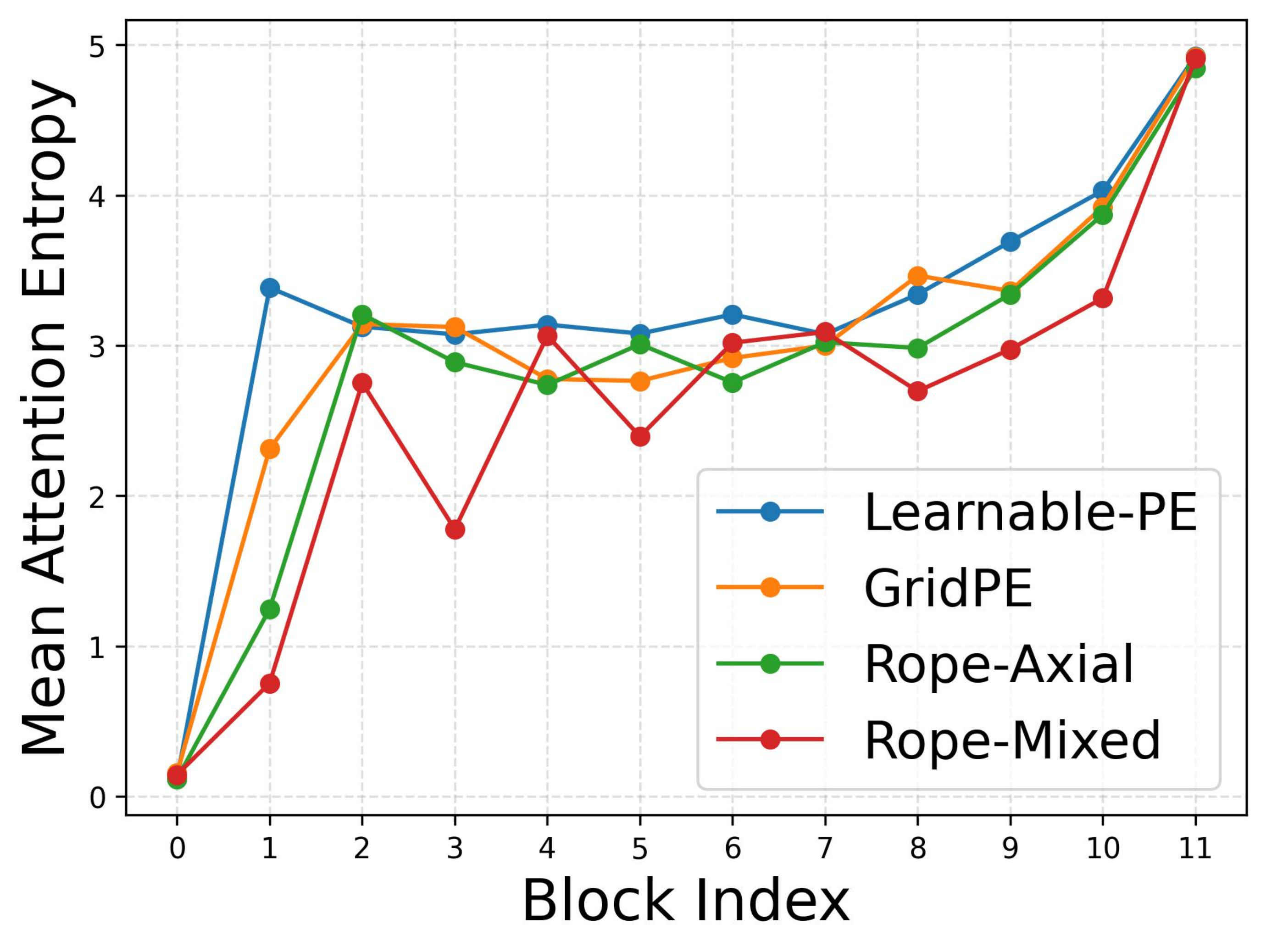}
      \caption{Input size = 224}
      \label{fig:entropy_224}
    \end{subfigure}
    \begin{subfigure}[b]{0.24\linewidth}
      \includegraphics[width=\linewidth]{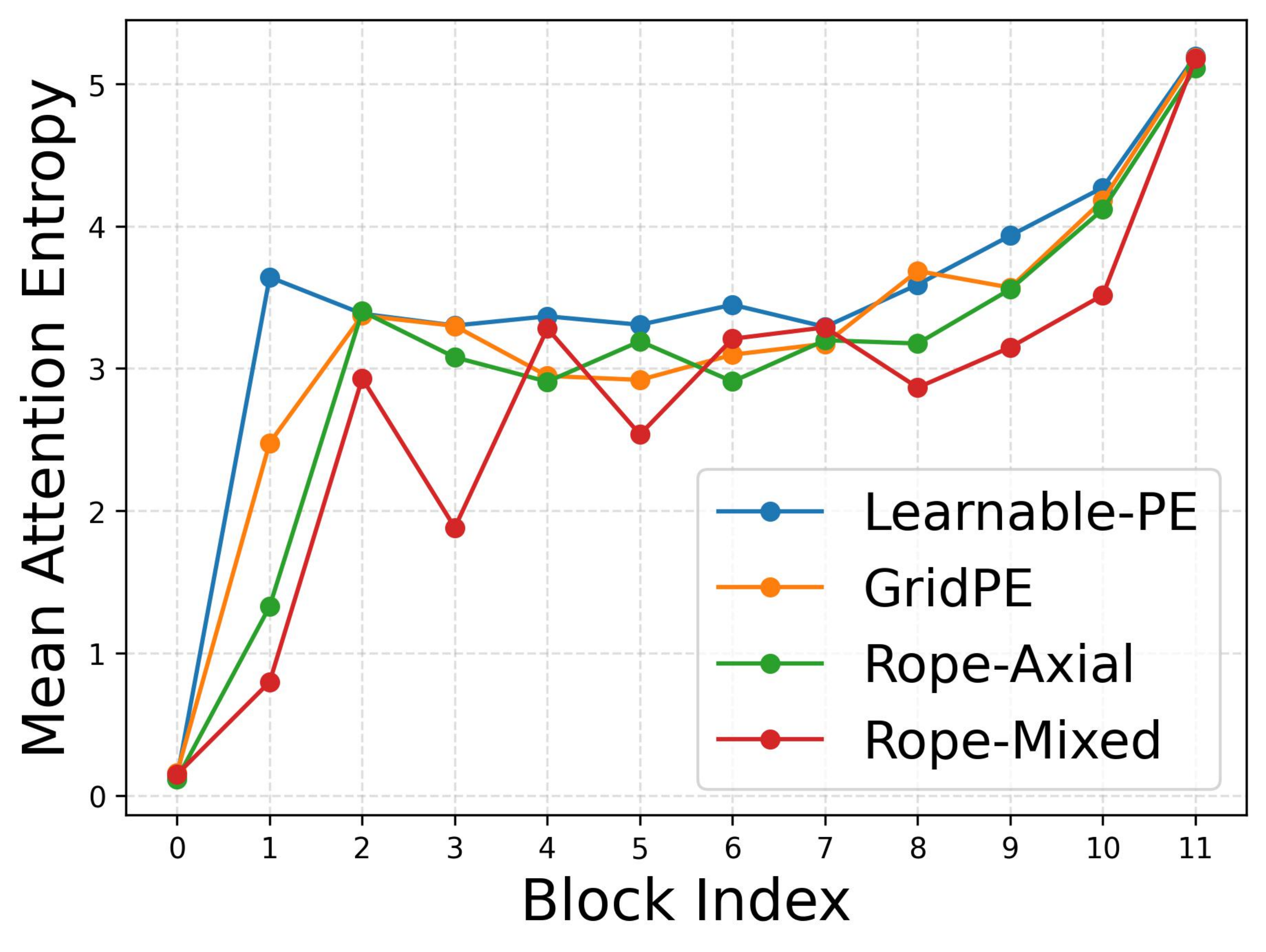}
      \caption{Input size = 256}
      \label{fig:entropy_256}
    \end{subfigure}
  
    \vspace{1ex}
  
    \begin{subfigure}[b]{0.24\linewidth}
      \includegraphics[width=\linewidth]{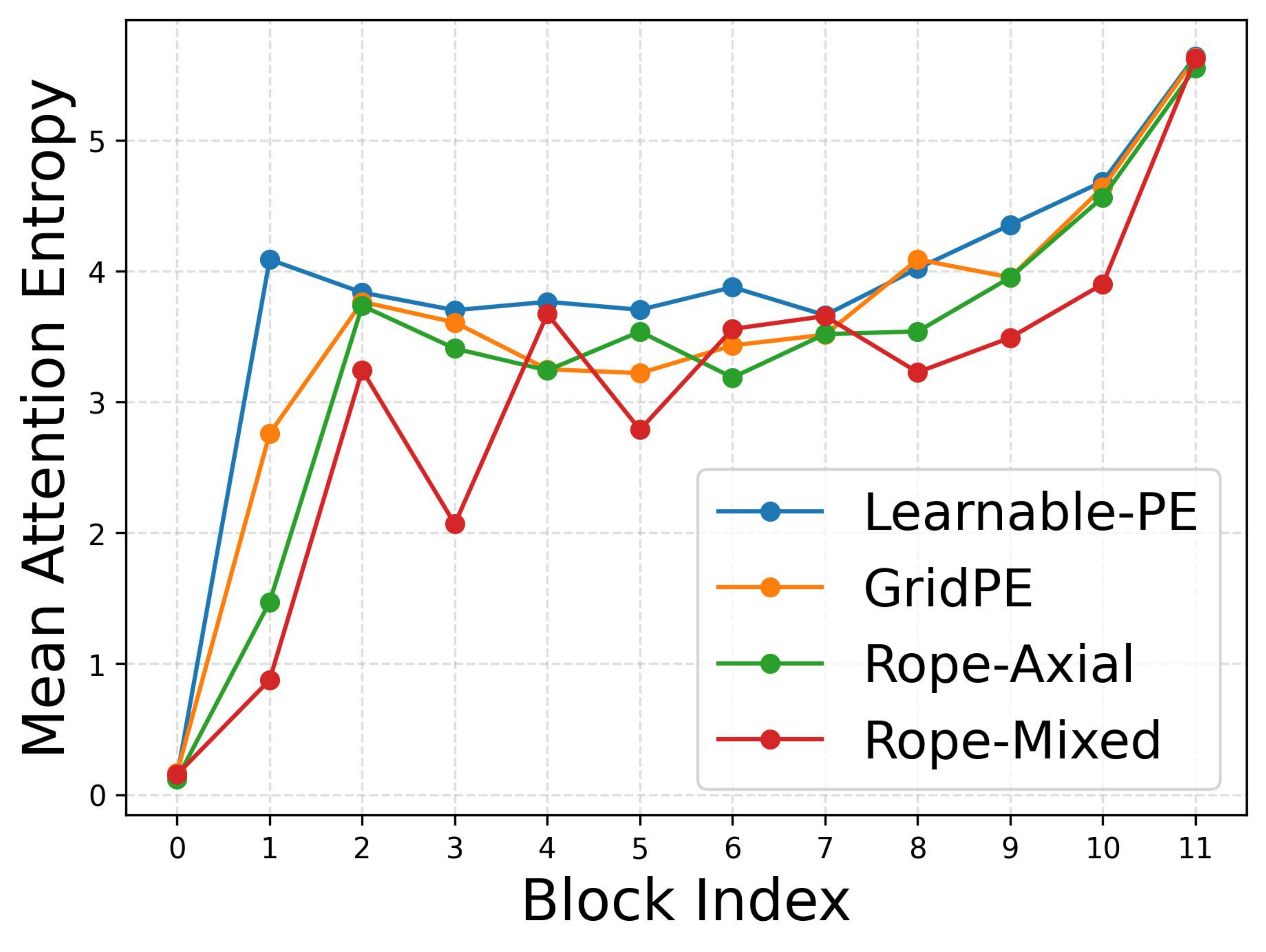}
      \caption{Input size = 320}
      \label{fig:entropy_320}
    \end{subfigure}
    \begin{subfigure}[b]{0.24\linewidth}
      \includegraphics[width=\linewidth]{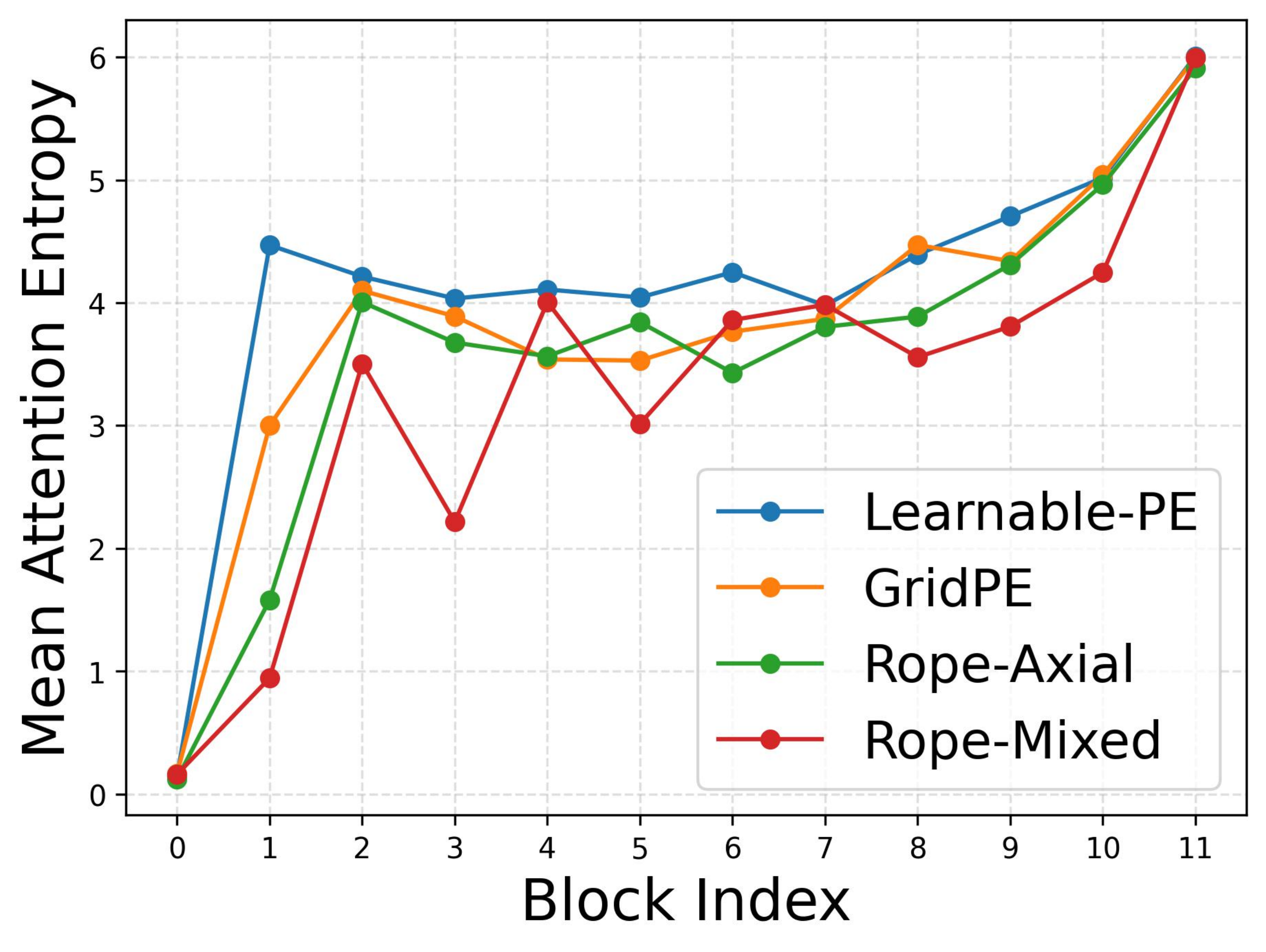}
      \caption{Input size = 384}
      \label{fig:entropy_384}
    \end{subfigure}
    \begin{subfigure}[b]{0.24\linewidth}
      \includegraphics[width=\linewidth]{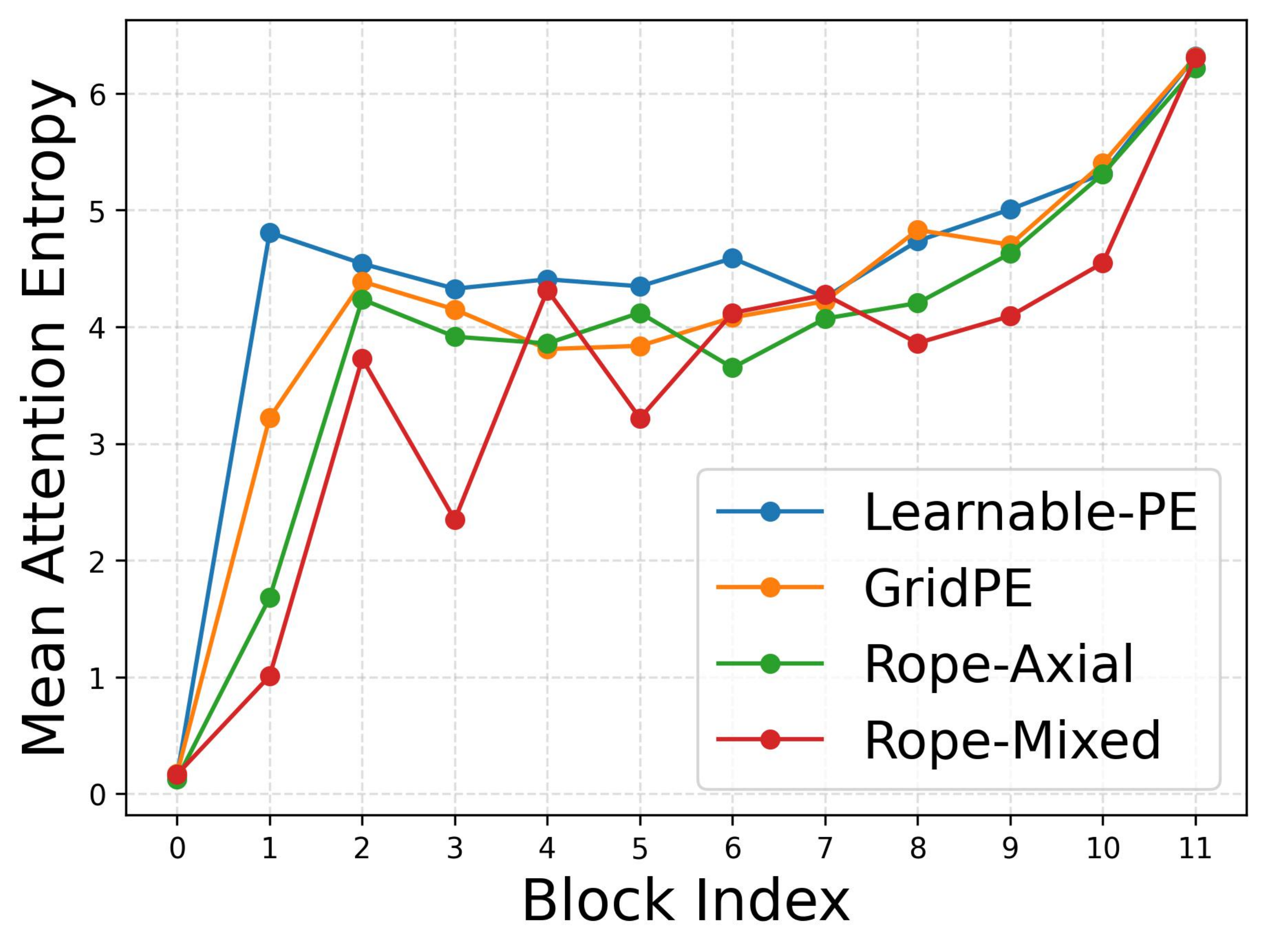}
      \caption{Input size = 448}
      \label{fig:entropy_448}
    \end{subfigure}
    \begin{subfigure}[b]{0.24\linewidth}
      \includegraphics[width=\linewidth]{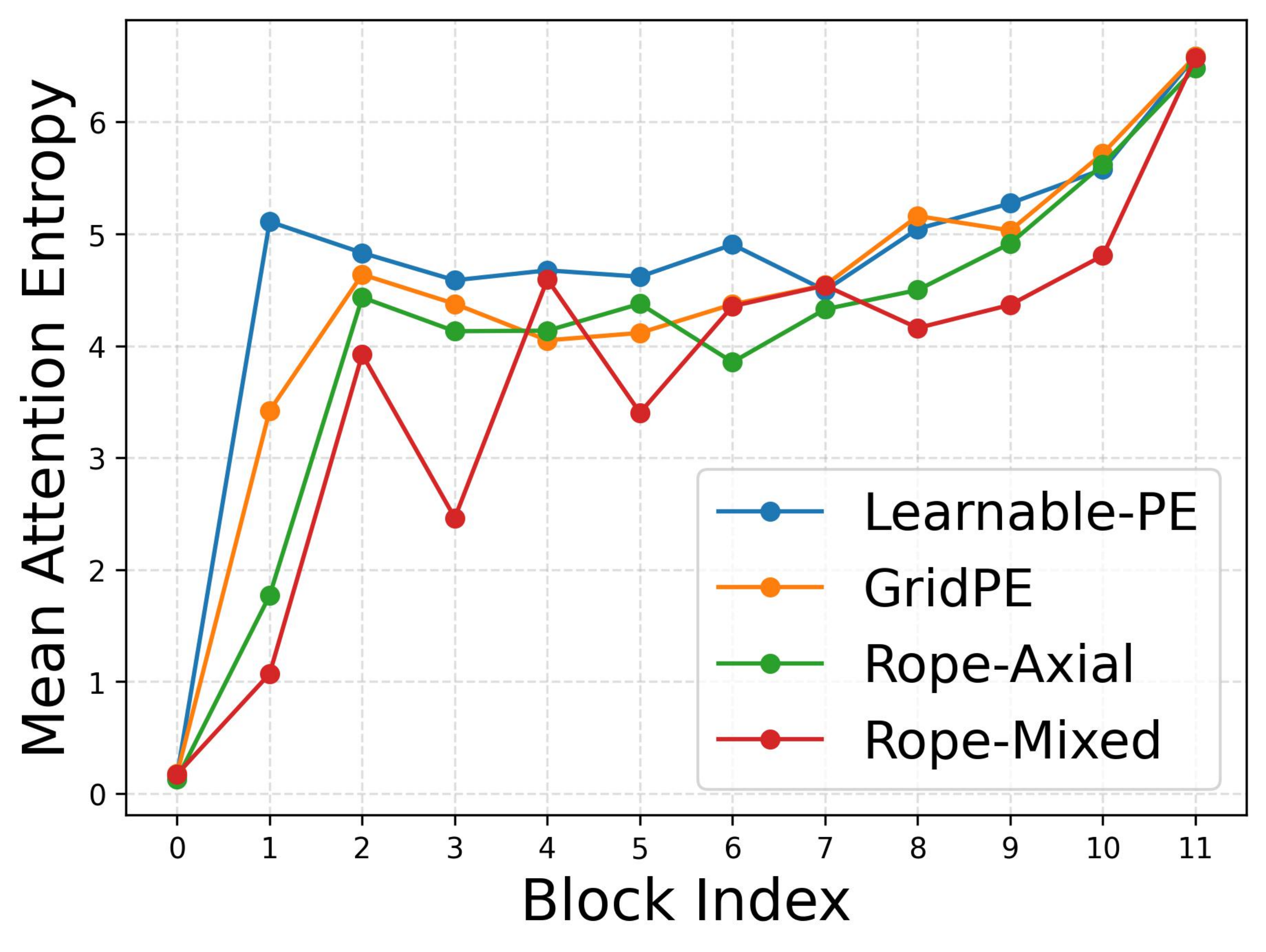}
      \caption{Input size = 512}
      \label{fig:entropy_512}
    \end{subfigure}
  
    \caption{Mean attention entropy per block on ViT-S for 2D image classification across input resolutions.}
    \label{fig:entropy_all}
\end{figure}

\begin{figure}[t]
    \centering
    \begin{subfigure}[b]{0.31\linewidth}
      \includegraphics[width=\linewidth]{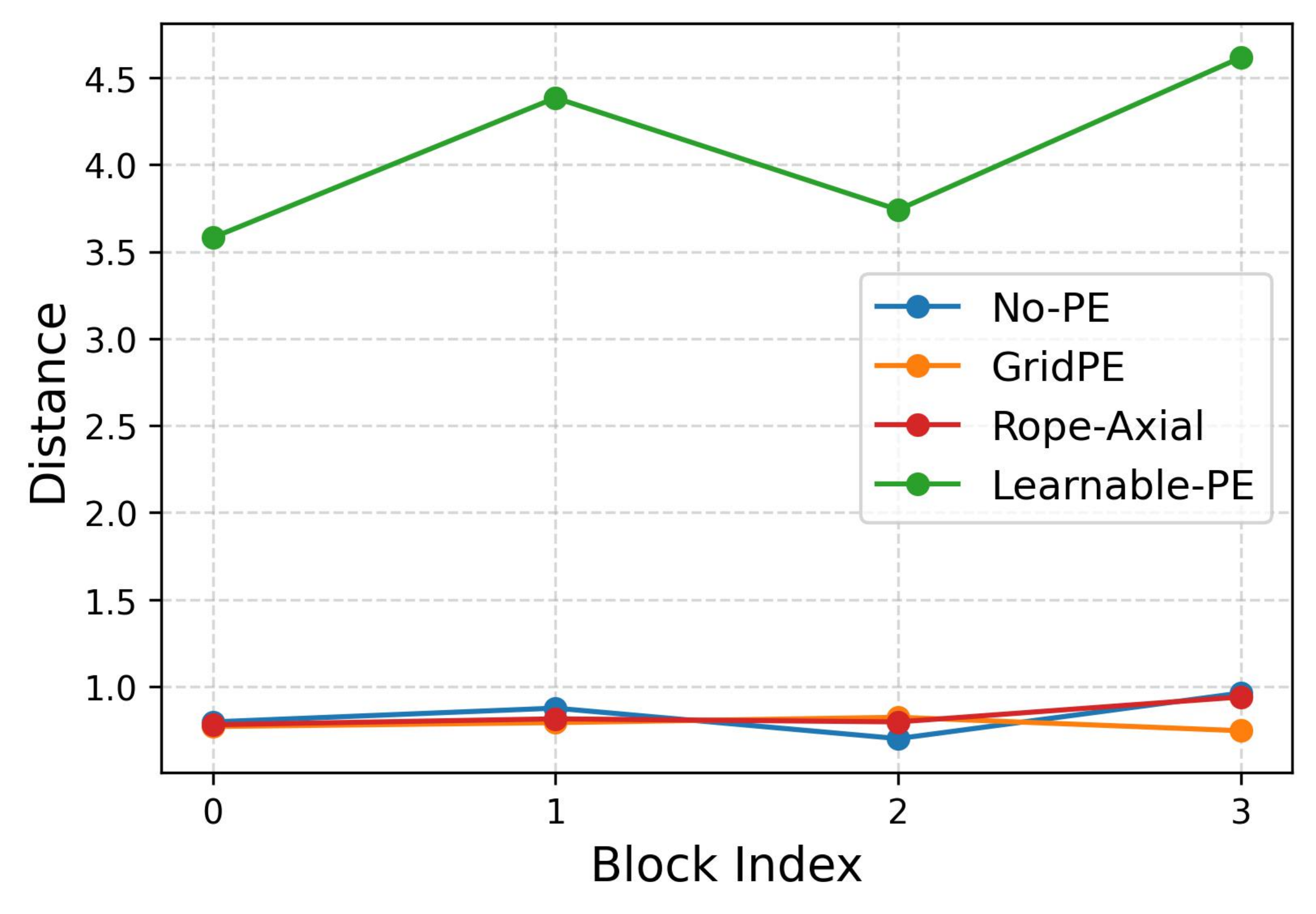}
      \caption{Input point = 256}
    \end{subfigure}\hfill
    \begin{subfigure}[b]{0.31\linewidth}
      \includegraphics[width=\linewidth]{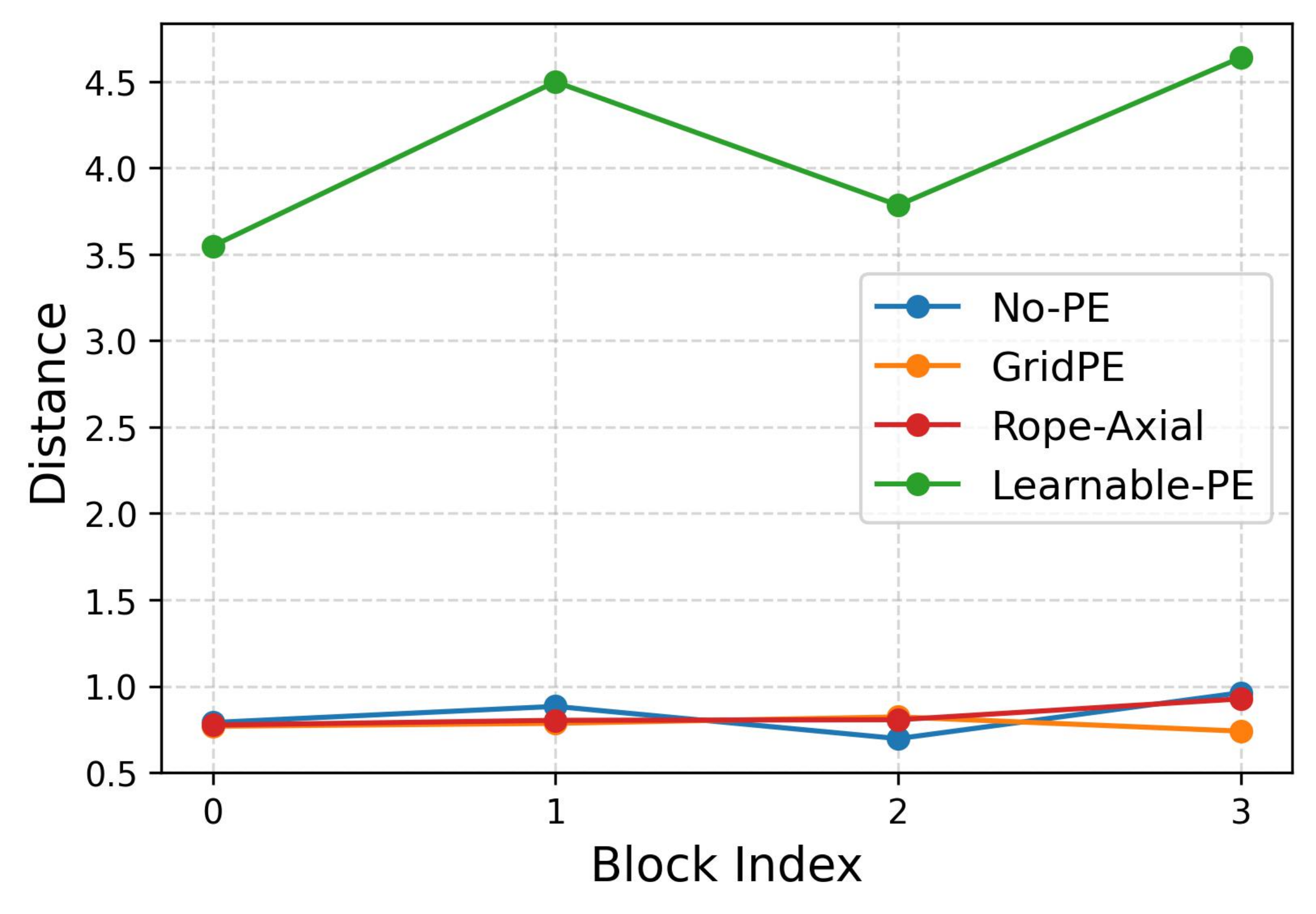}
      \caption{Input point = 384}
    \end{subfigure}\hfill
    \begin{subfigure}[b]{0.31\linewidth}
      \includegraphics[width=\linewidth]{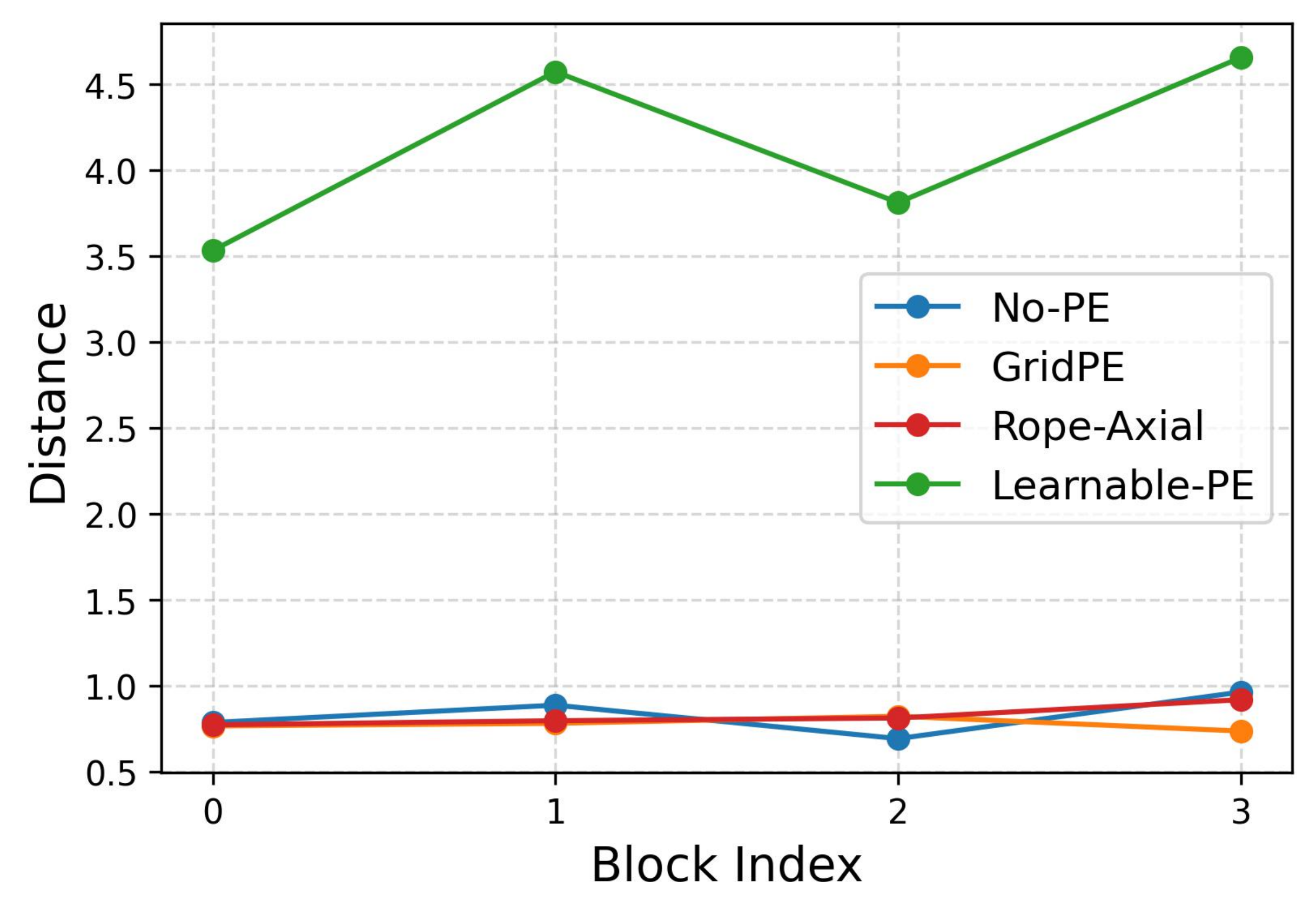}
      \caption{Input point = 512}
    \end{subfigure}

    \vspace{1ex}
    \begin{subfigure}[b]{0.31\linewidth}
      \includegraphics[width=\linewidth]{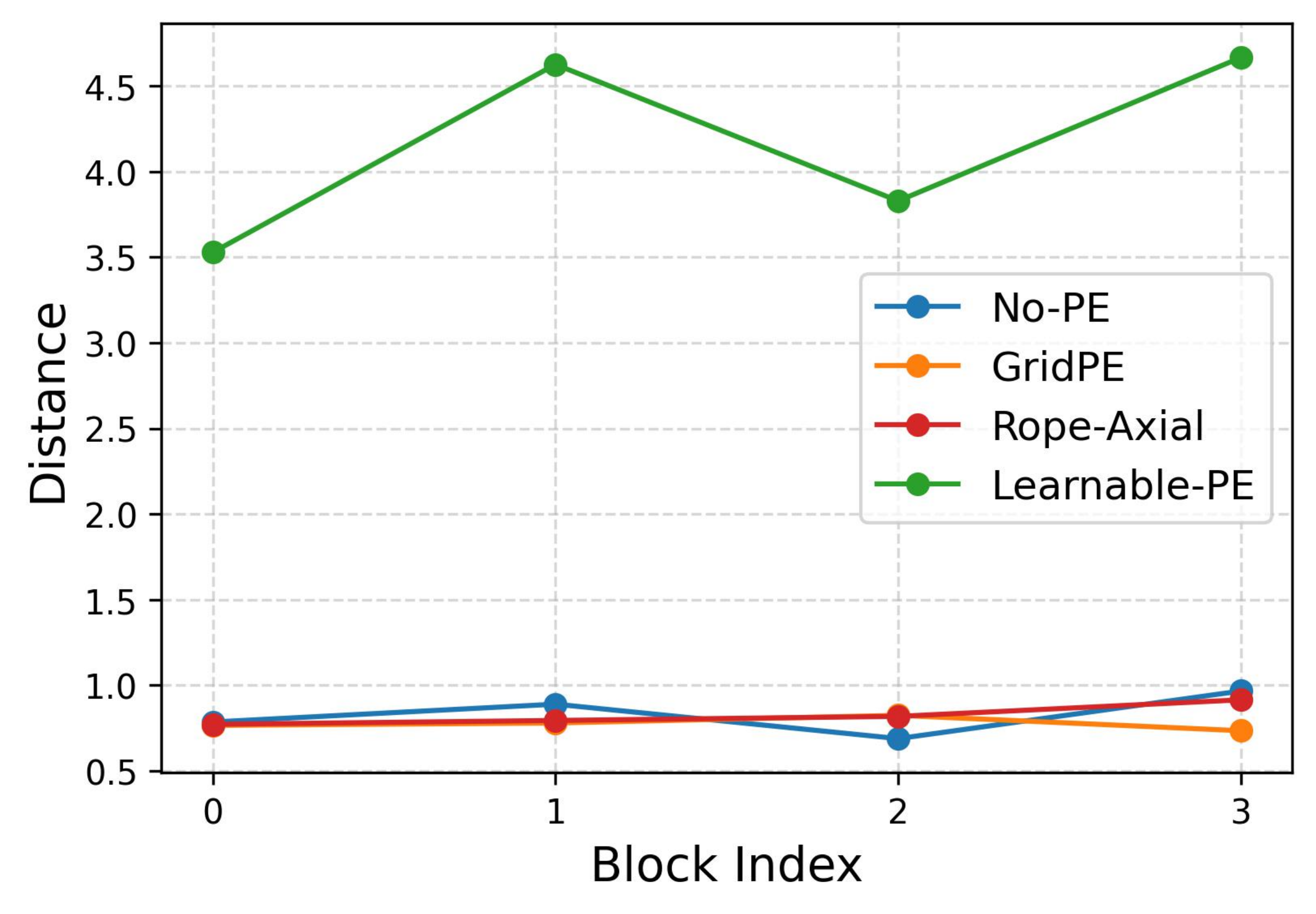}
      \caption{Input point = 640}
    \end{subfigure}\hfill
    \begin{subfigure}[b]{0.31\linewidth}
      \includegraphics[width=\linewidth]{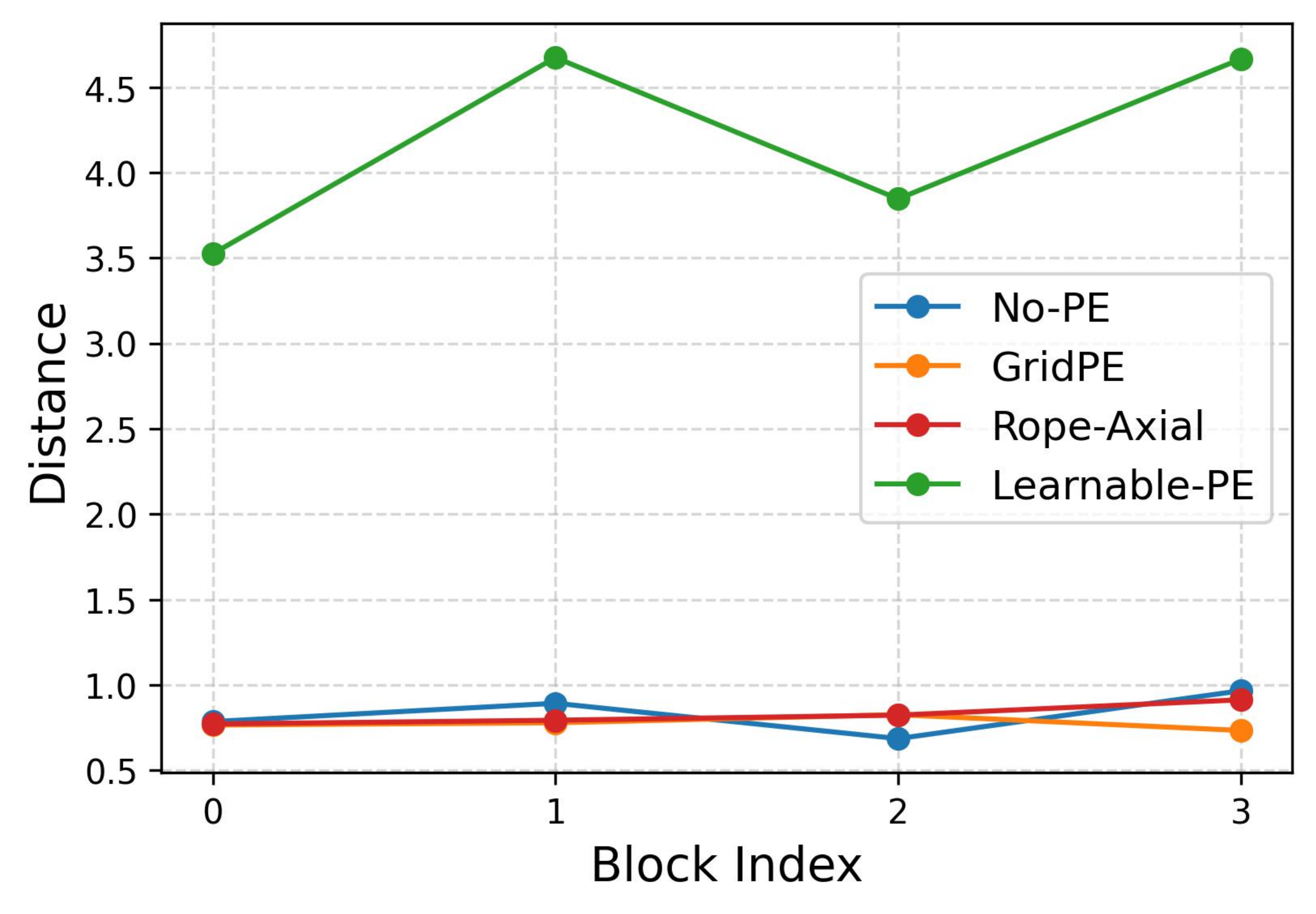}
      \caption{Input point = 768}
    \end{subfigure}\hfill
    \begin{subfigure}[b]{0.31\linewidth}
      \includegraphics[width=\linewidth]{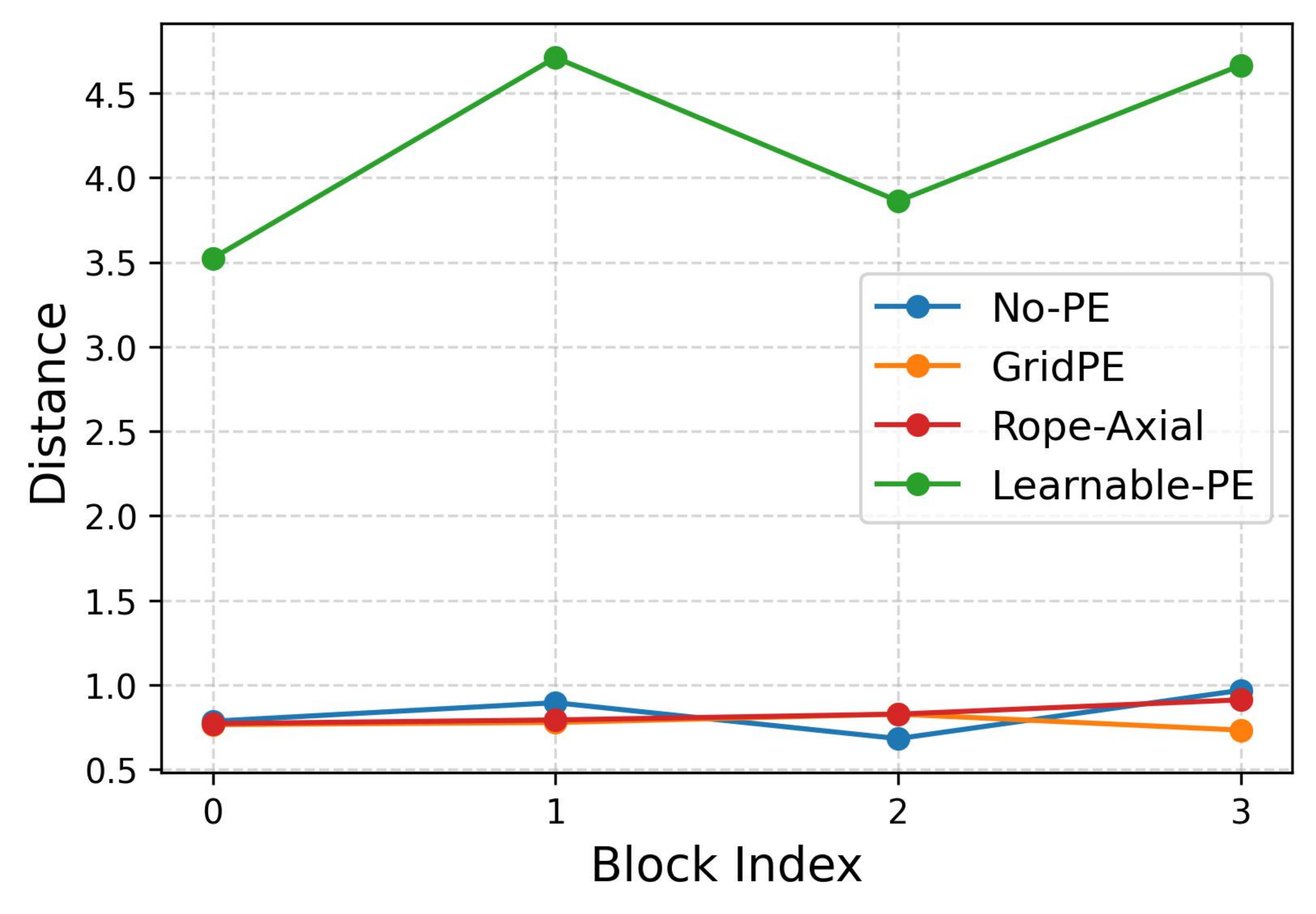}
      \caption{Input point = 896}
    \end{subfigure}

    \vspace{1ex}
    \begin{subfigure}[b]{0.31\linewidth}
      \includegraphics[width=\linewidth]{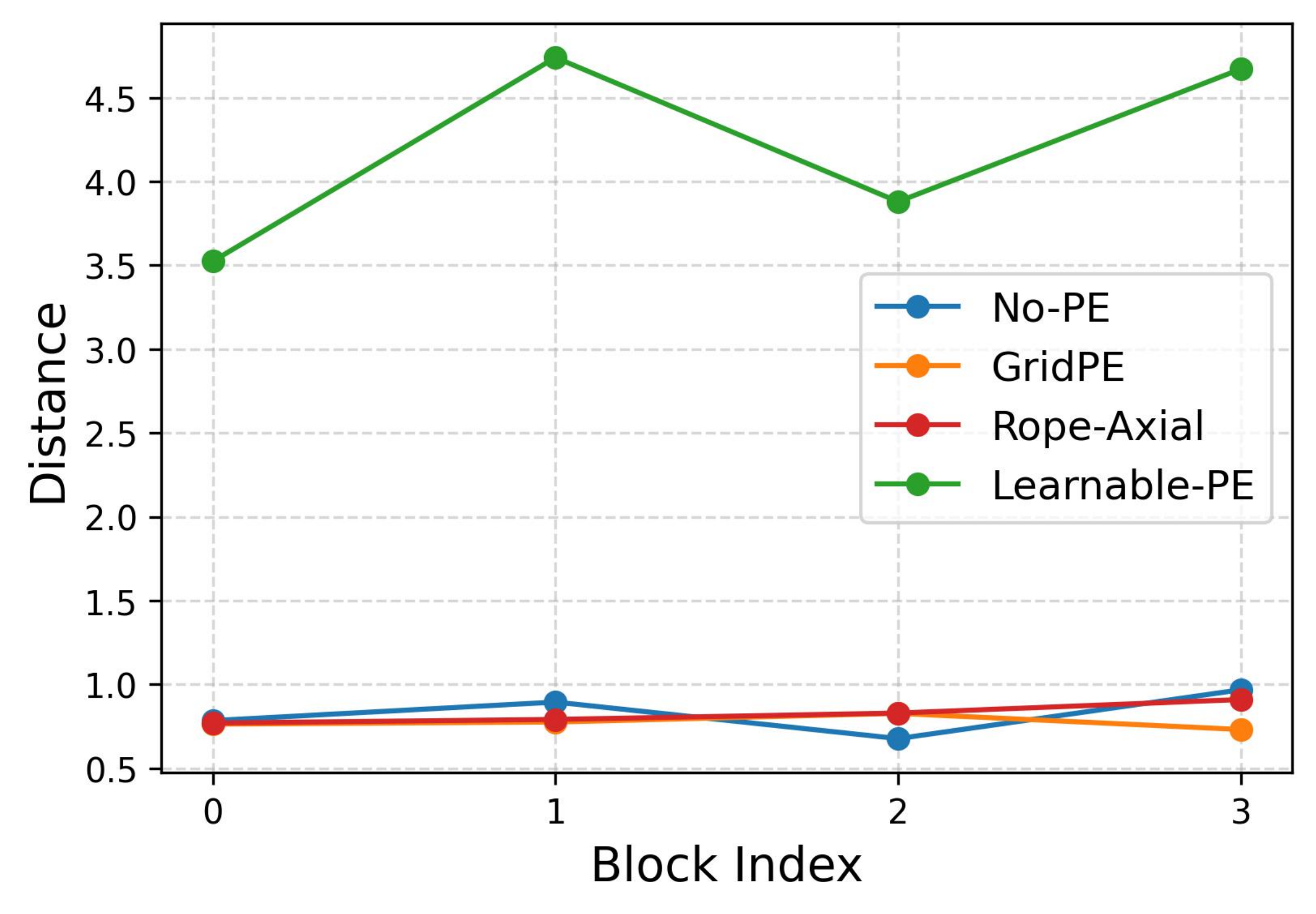}
      \caption{Input point = 1024}
    \end{subfigure}\hfill
    \begin{subfigure}[b]{0.31\linewidth}
      \includegraphics[width=\linewidth]{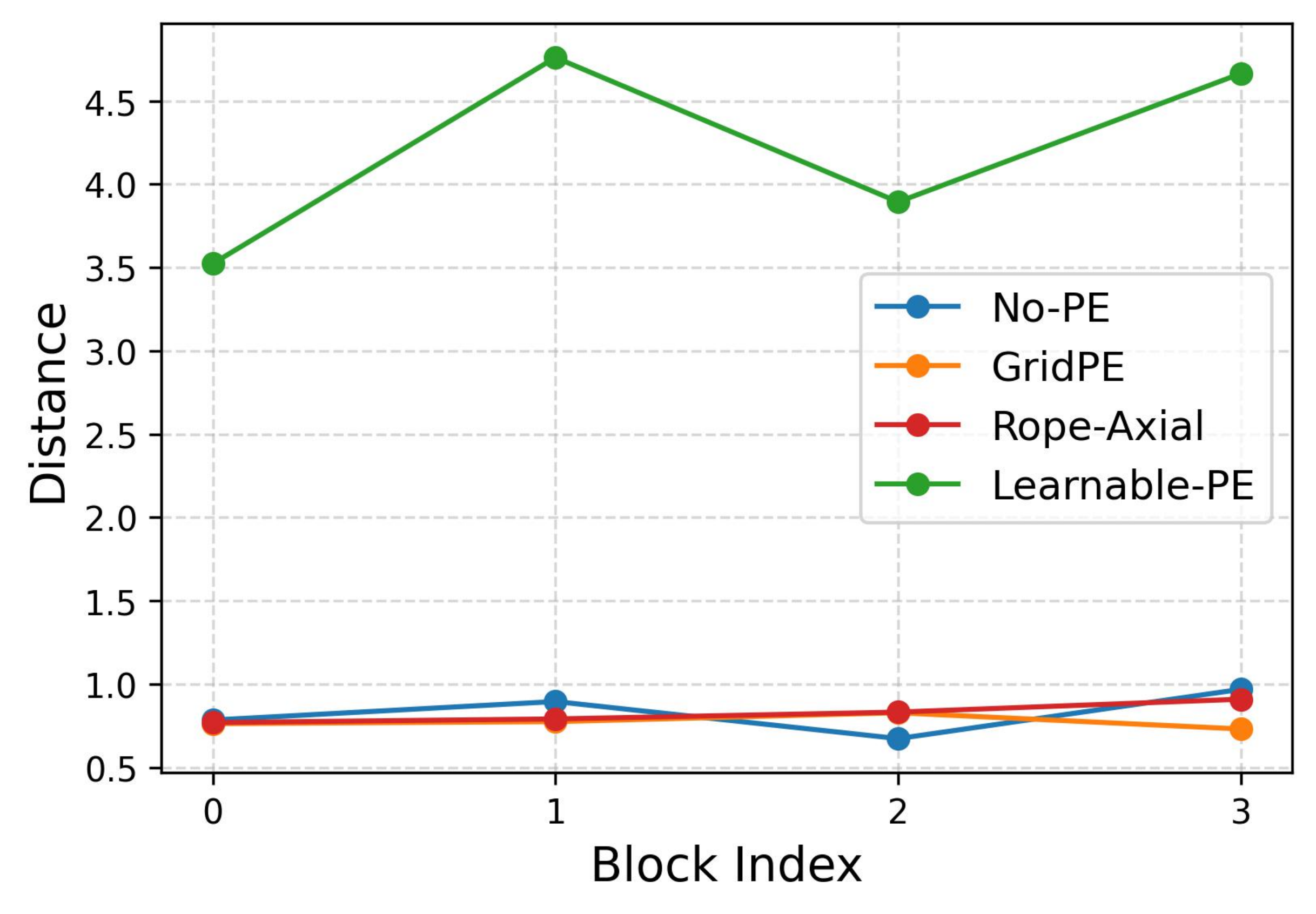}
      \caption{Input point = 1152}
    \end{subfigure}\hfill
    \begin{subfigure}[b]{0.31\linewidth}
      \includegraphics[width=\linewidth]{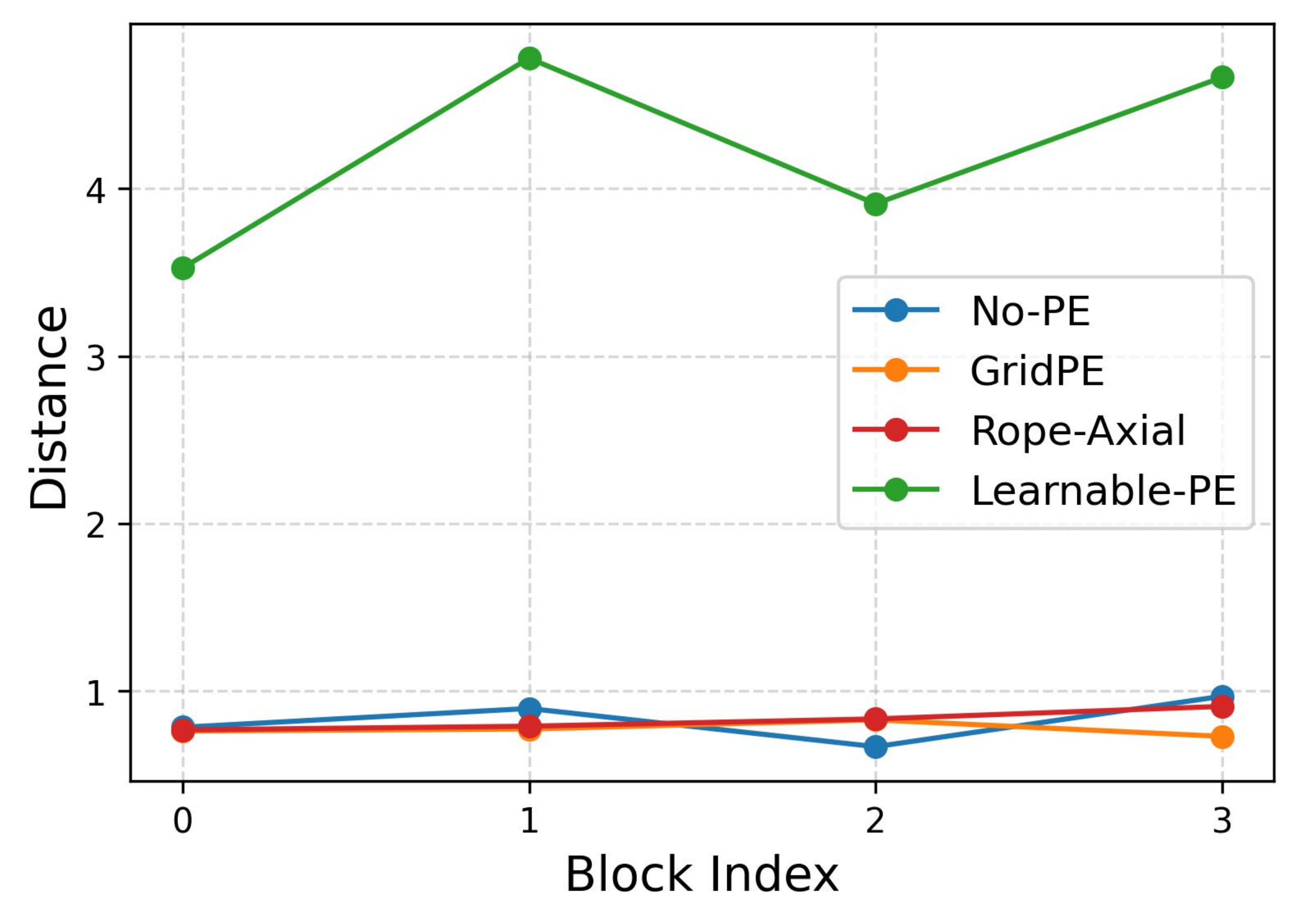}
      \caption{Input point = 1280}
    \end{subfigure}

    \vspace{1ex}
    \begin{subfigure}[b]{0.31\linewidth}
      \includegraphics[width=\linewidth]{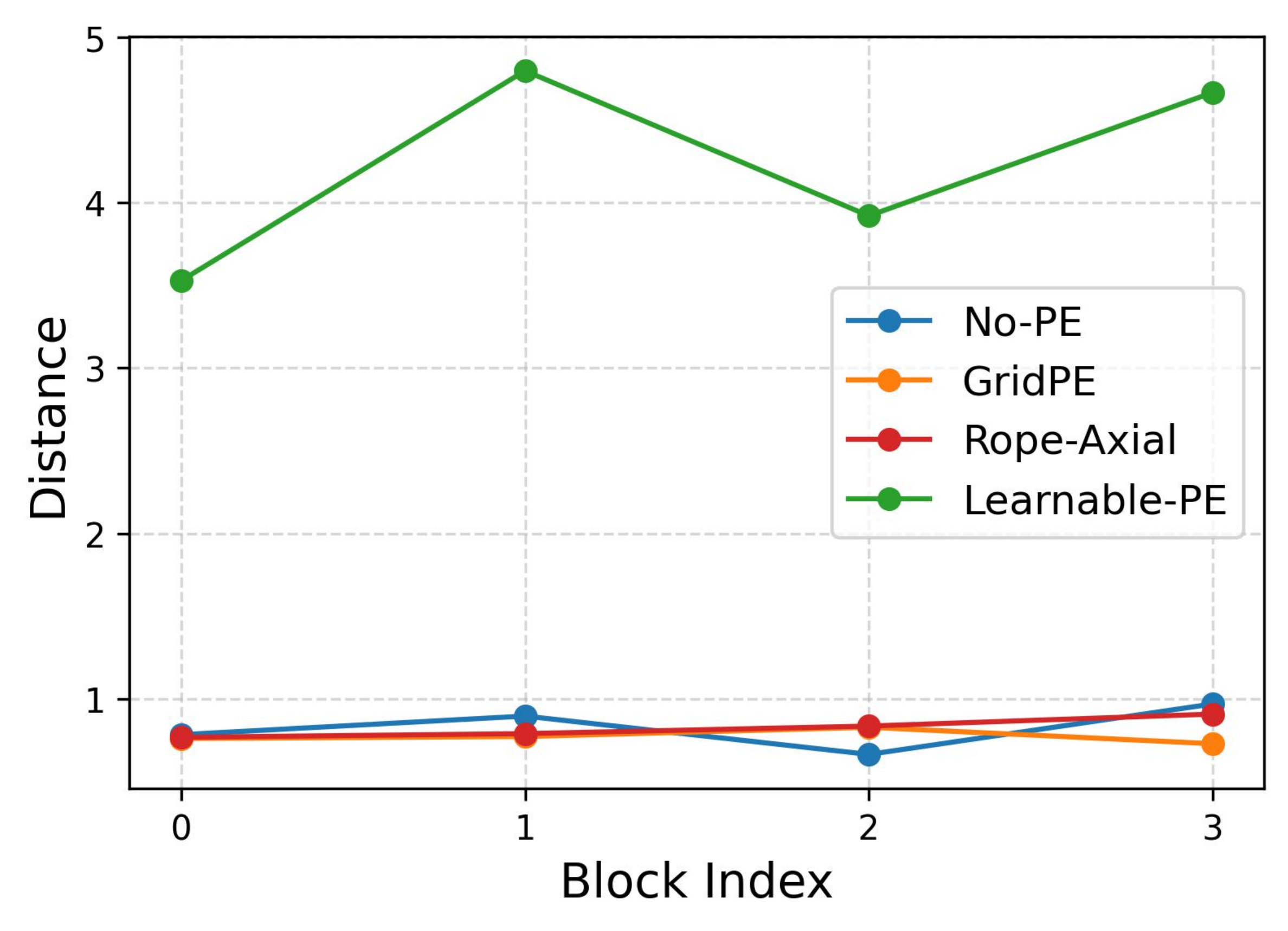}
      \caption{Input point = 1408}
    \end{subfigure}\hfill
    \begin{subfigure}[b]{0.31\linewidth}
      \includegraphics[width=\linewidth]{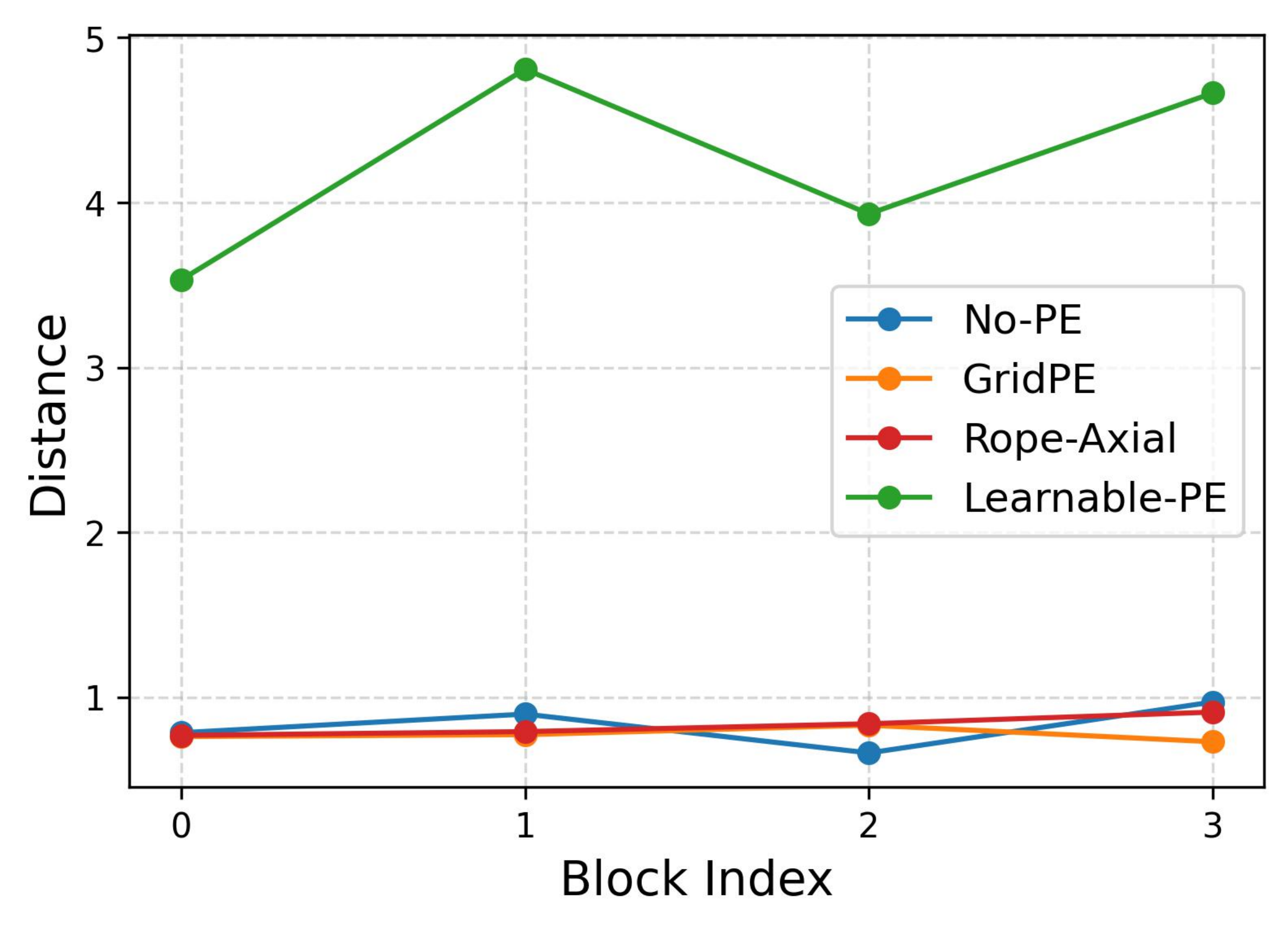}
      \caption{Input point = 1536}
    \end{subfigure}\hfill
    \begin{subfigure}[b]{0.31\linewidth}
      \includegraphics[width=\linewidth]{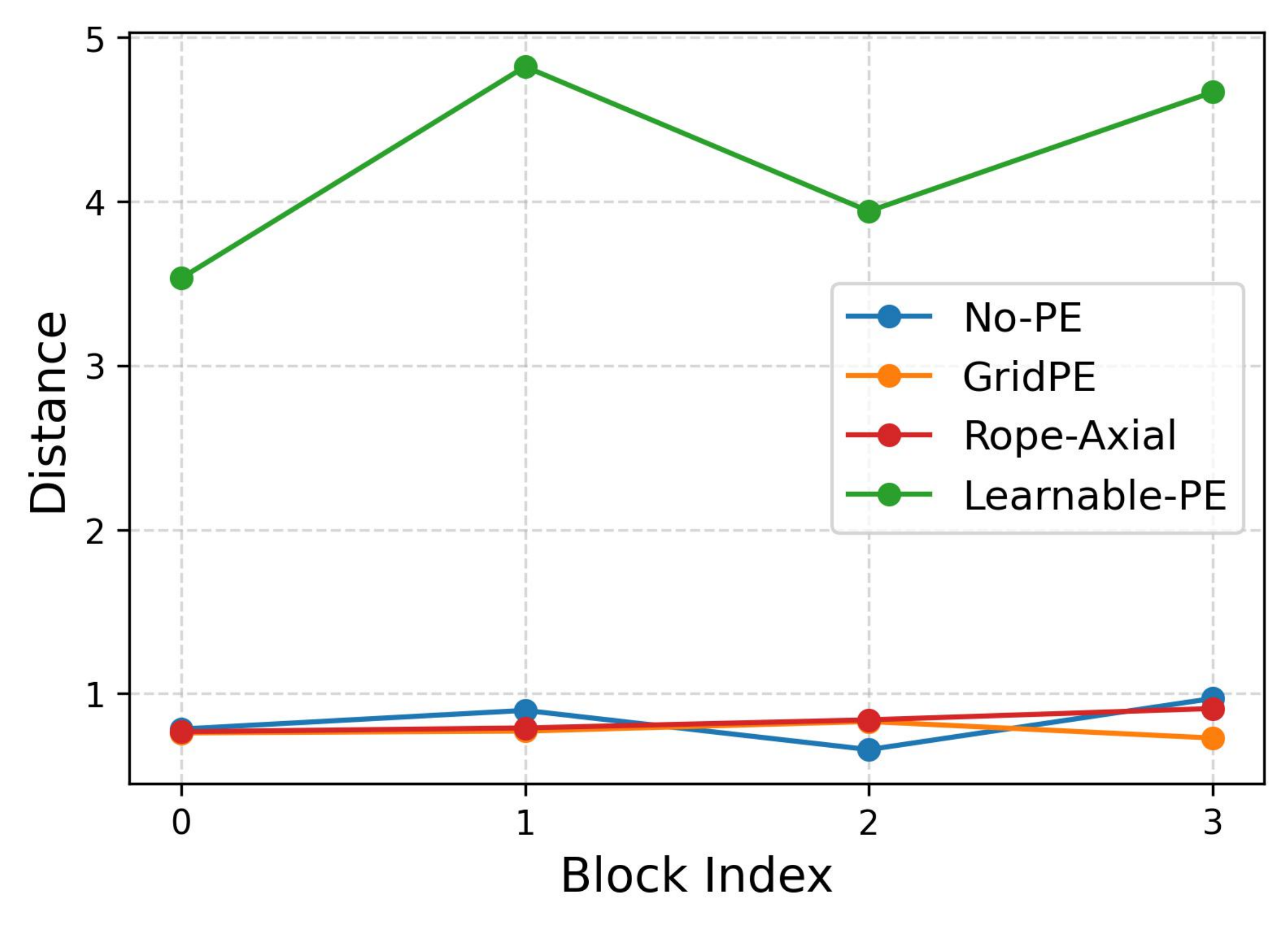}
      \caption{Input point = 1664}
    \end{subfigure}

    \vspace{1ex}
    \begin{subfigure}[b]{0.31\linewidth}
      \includegraphics[width=\linewidth]{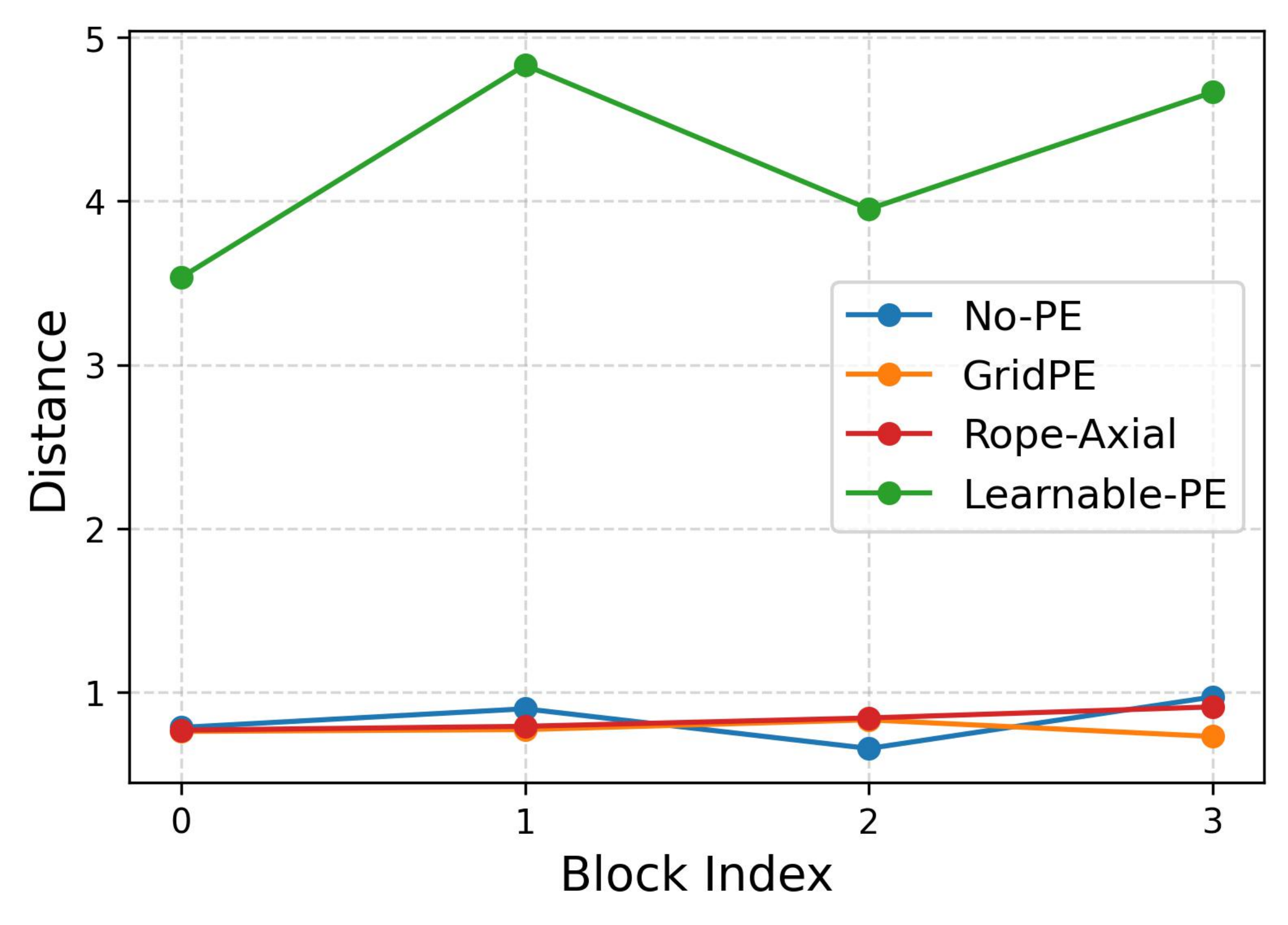}
      \caption{Input point = 1792}
    \end{subfigure}\hfill
    \begin{subfigure}[b]{0.31\linewidth}
      \includegraphics[width=\linewidth]{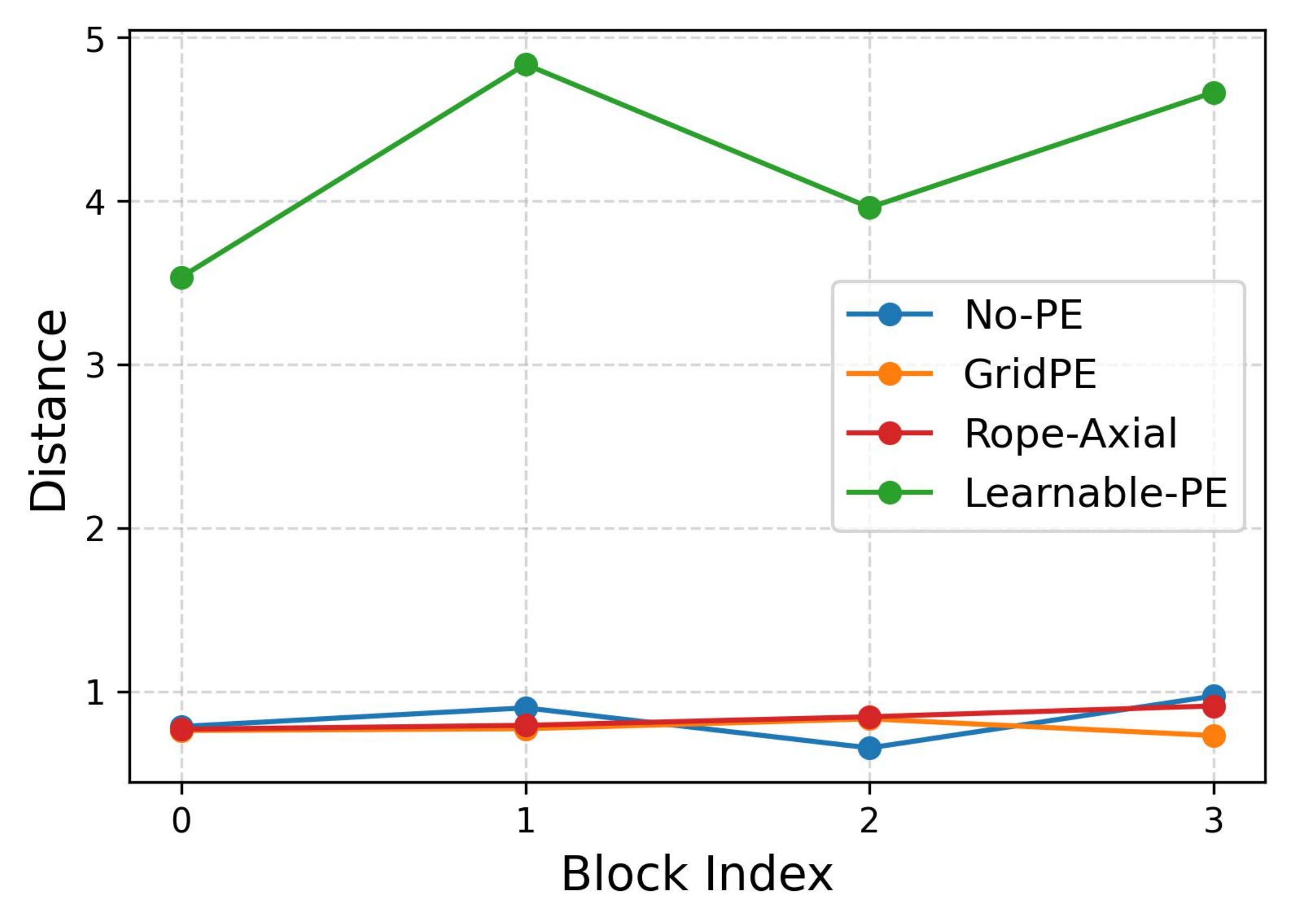}
      \caption{Input point = 1920}
    \end{subfigure}\hfill
    \begin{subfigure}[b]{0.31\linewidth}
      \includegraphics[width=\linewidth]{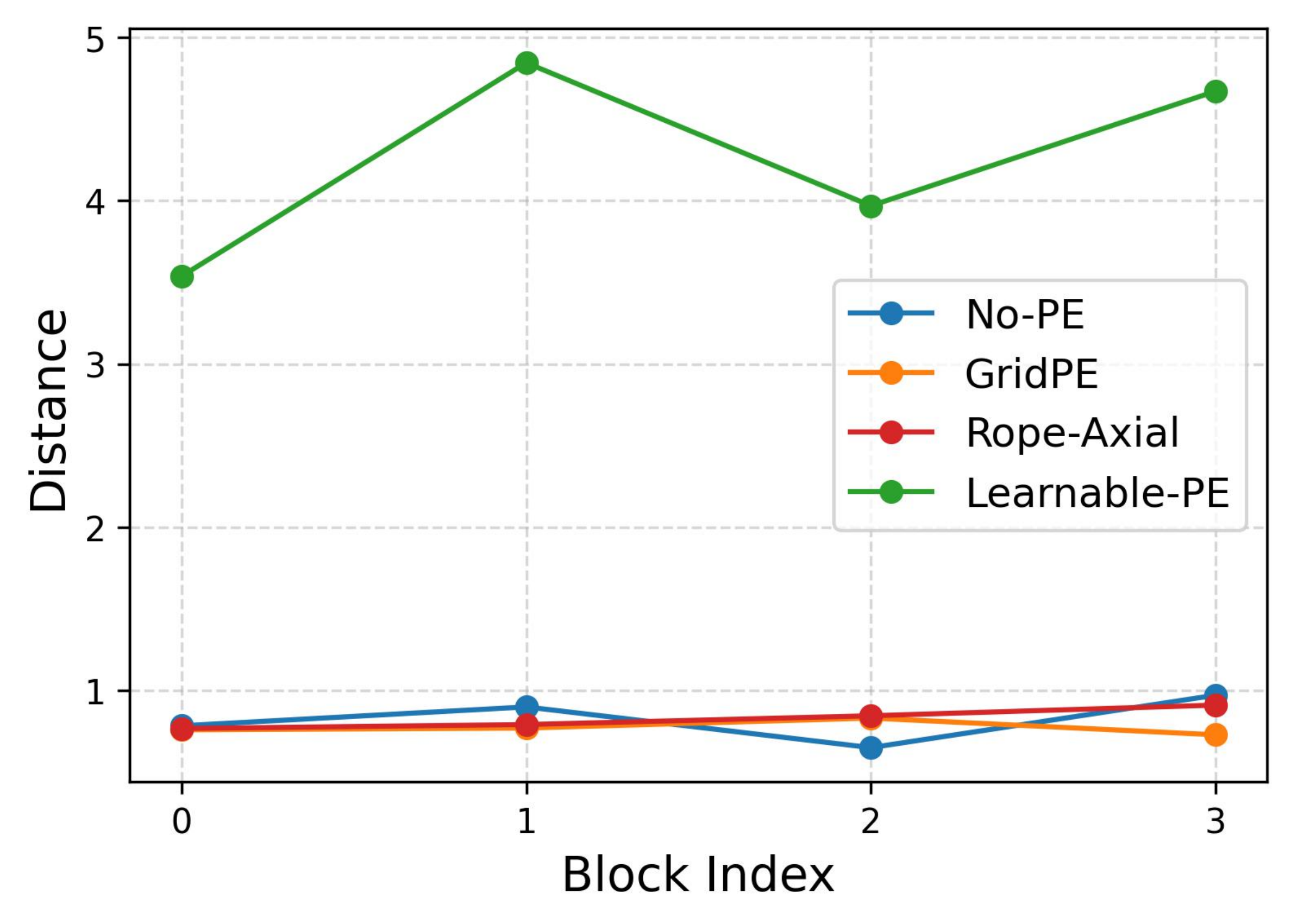}
      \caption{Input point = 2048}
    \end{subfigure}

    \caption{Mean attention distance per block layer across input point counts.}
    \label{fig:distance_all_3d}
\end{figure}

\begin{figure}[t]
    \centering
    \begin{subfigure}[b]{0.31\linewidth}
      \includegraphics[width=\linewidth]{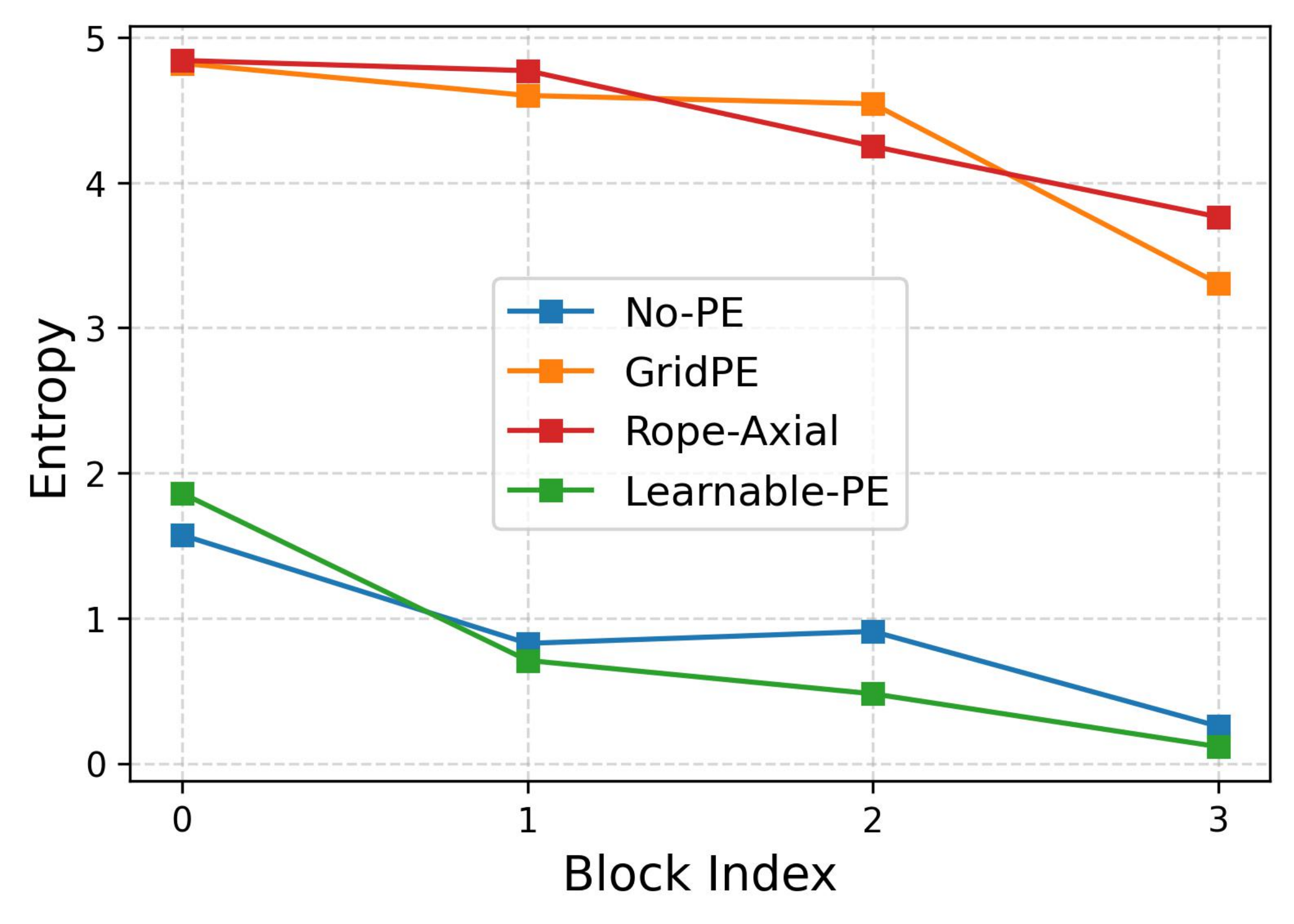}
      \caption{Input point = 256}
    \end{subfigure}\hfill
    \begin{subfigure}[b]{0.31\linewidth}
      \includegraphics[width=\linewidth]{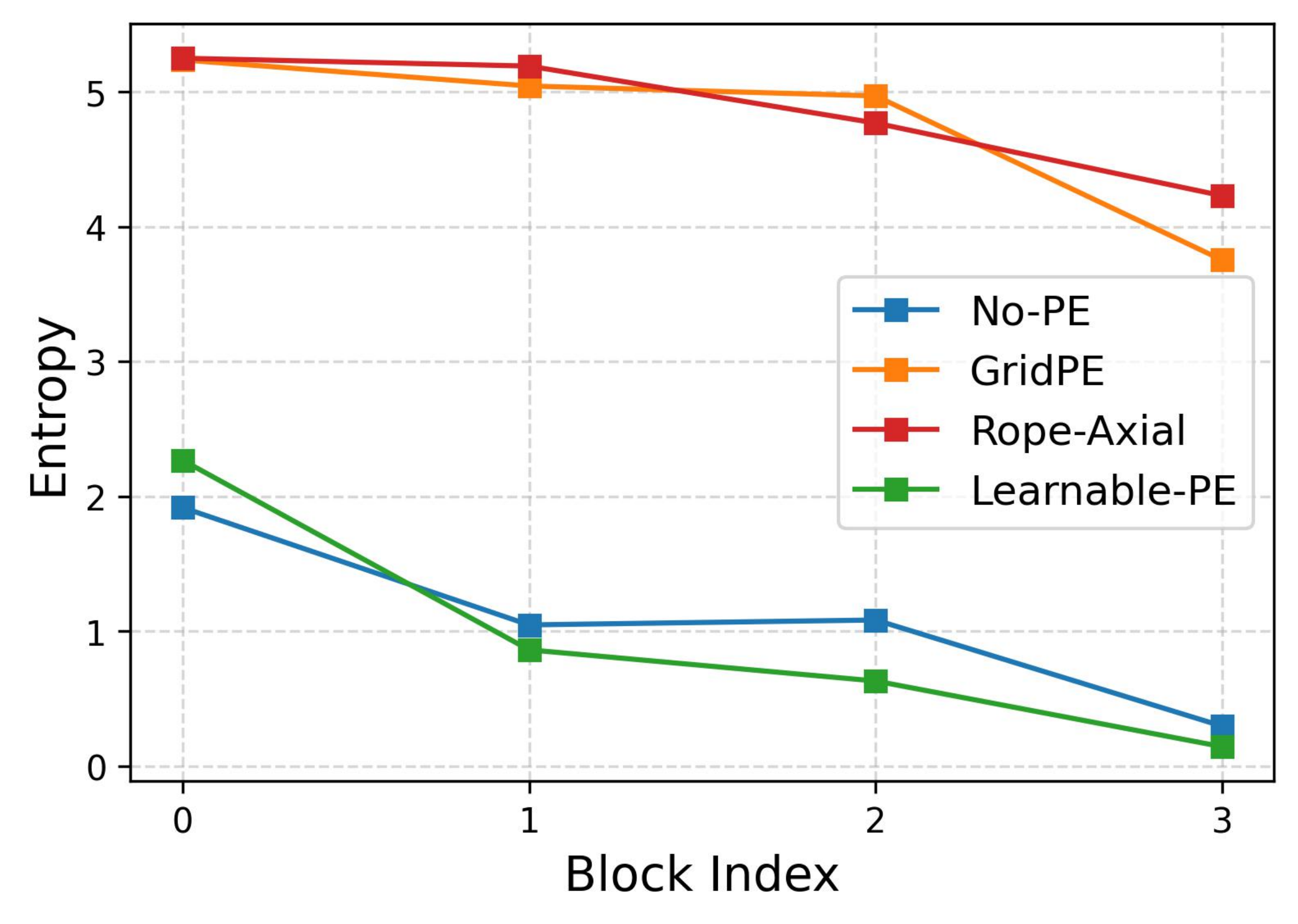}
      \caption{Input point = 384}
    \end{subfigure}\hfill
    \begin{subfigure}[b]{0.31\linewidth}
      \includegraphics[width=\linewidth]{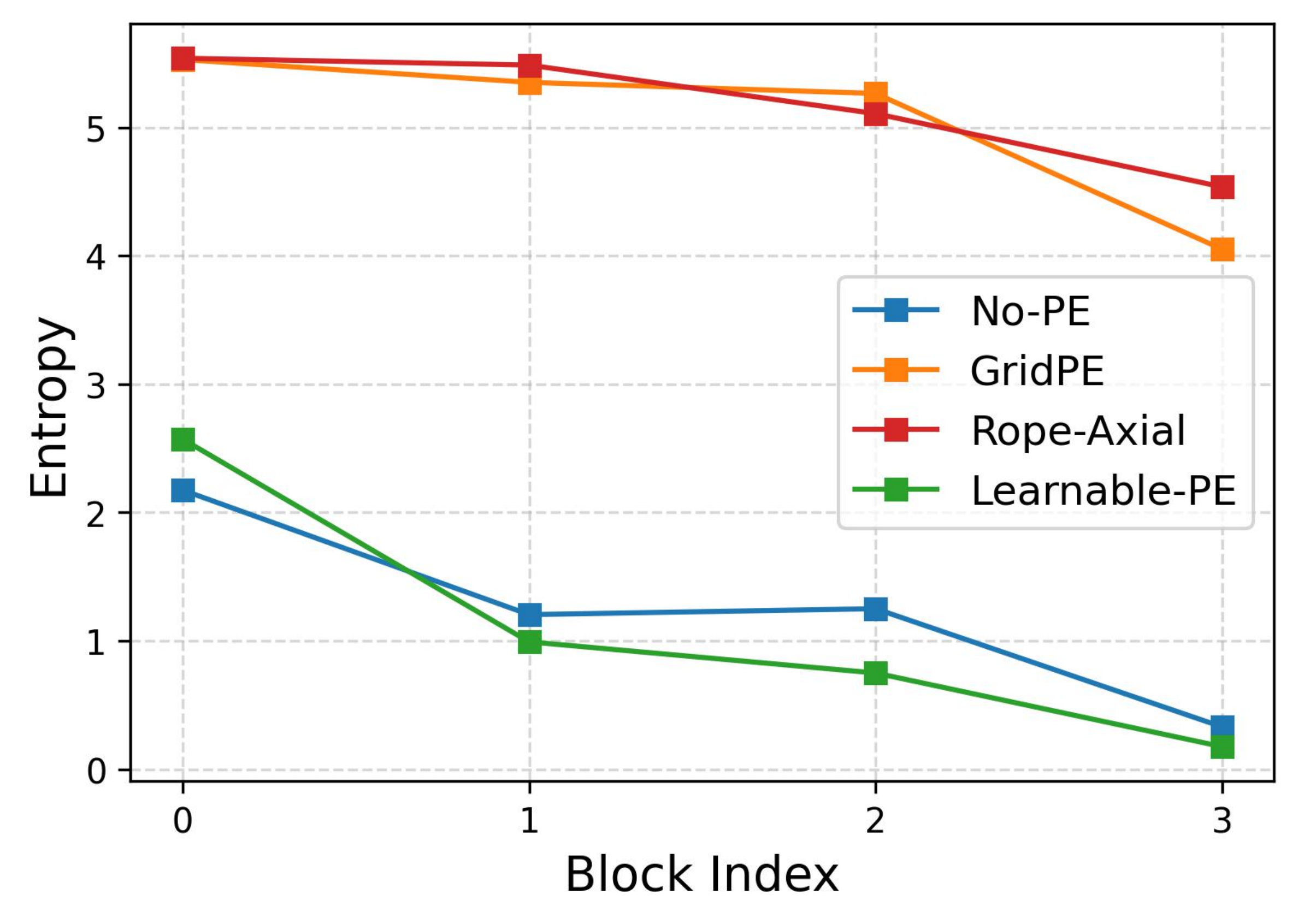}
      \caption{Input point = 512}
    \end{subfigure}

    \vspace{1ex}
    \begin{subfigure}[b]{0.31\linewidth}
      \includegraphics[width=\linewidth]{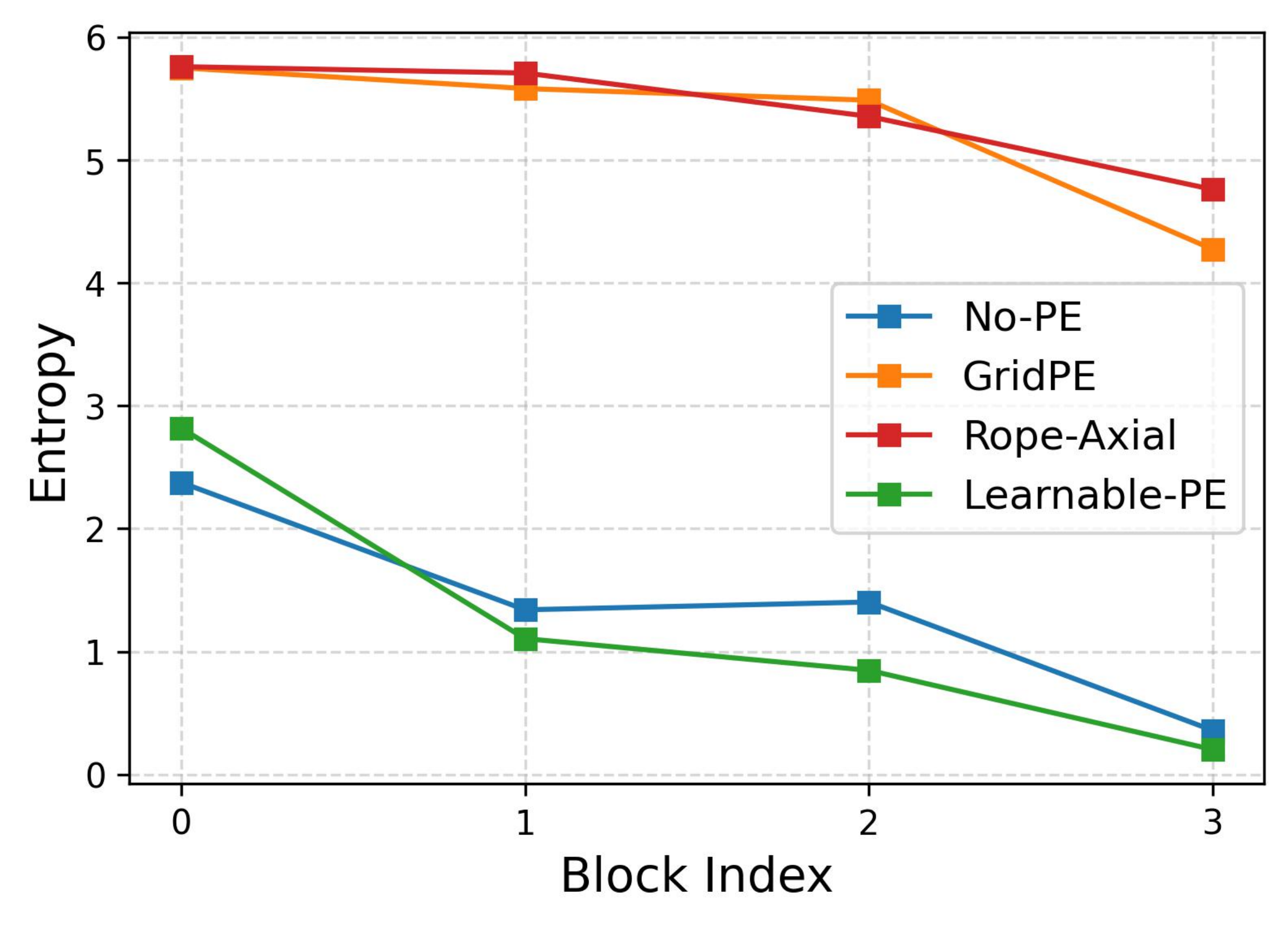}
      \caption{Input point = 640}
    \end{subfigure}\hfill
    \begin{subfigure}[b]{0.31\linewidth}
      \includegraphics[width=\linewidth]{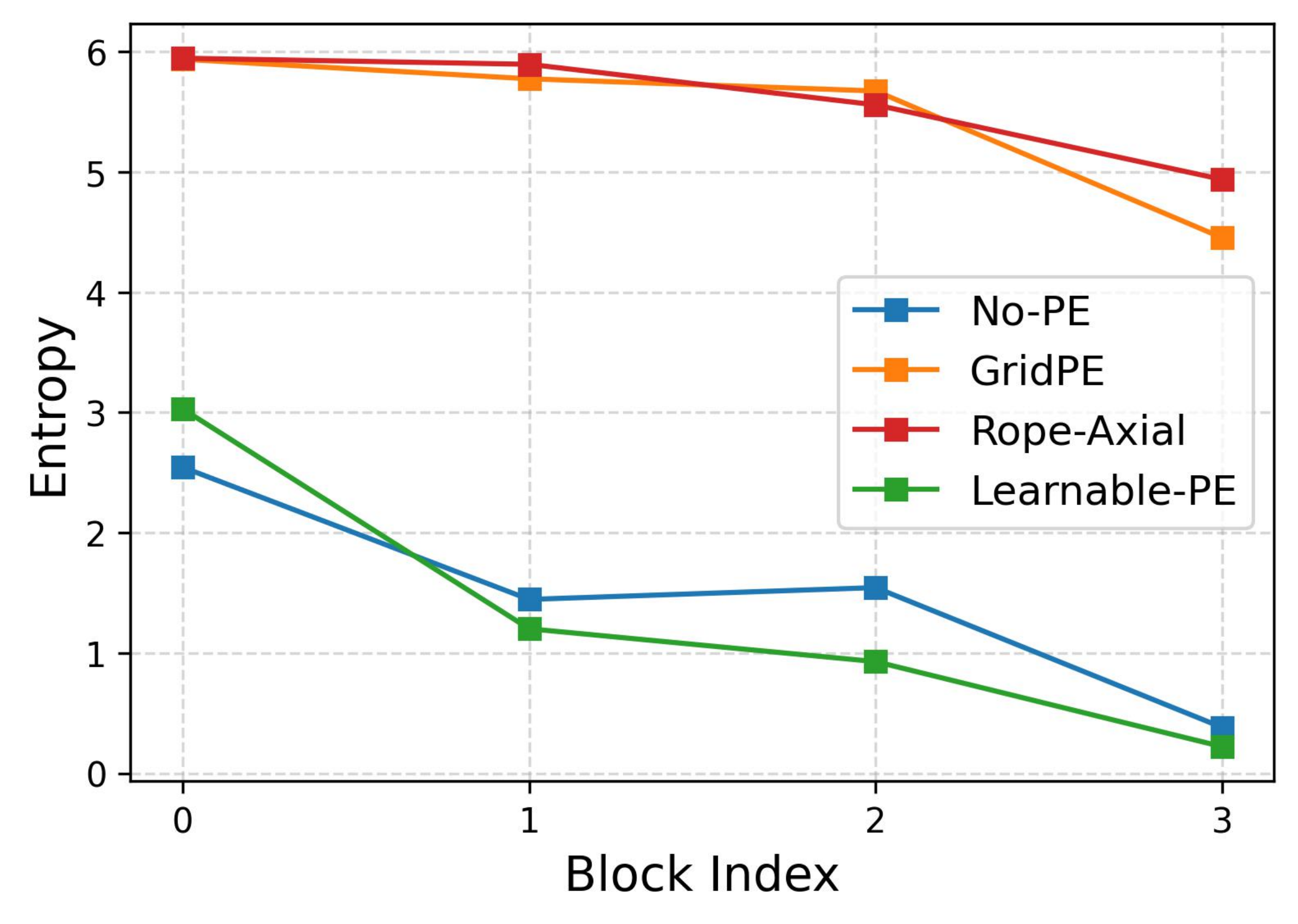}
      \caption{Input point = 768}
    \end{subfigure}\hfill
    \begin{subfigure}[b]{0.31\linewidth}
      \includegraphics[width=\linewidth]{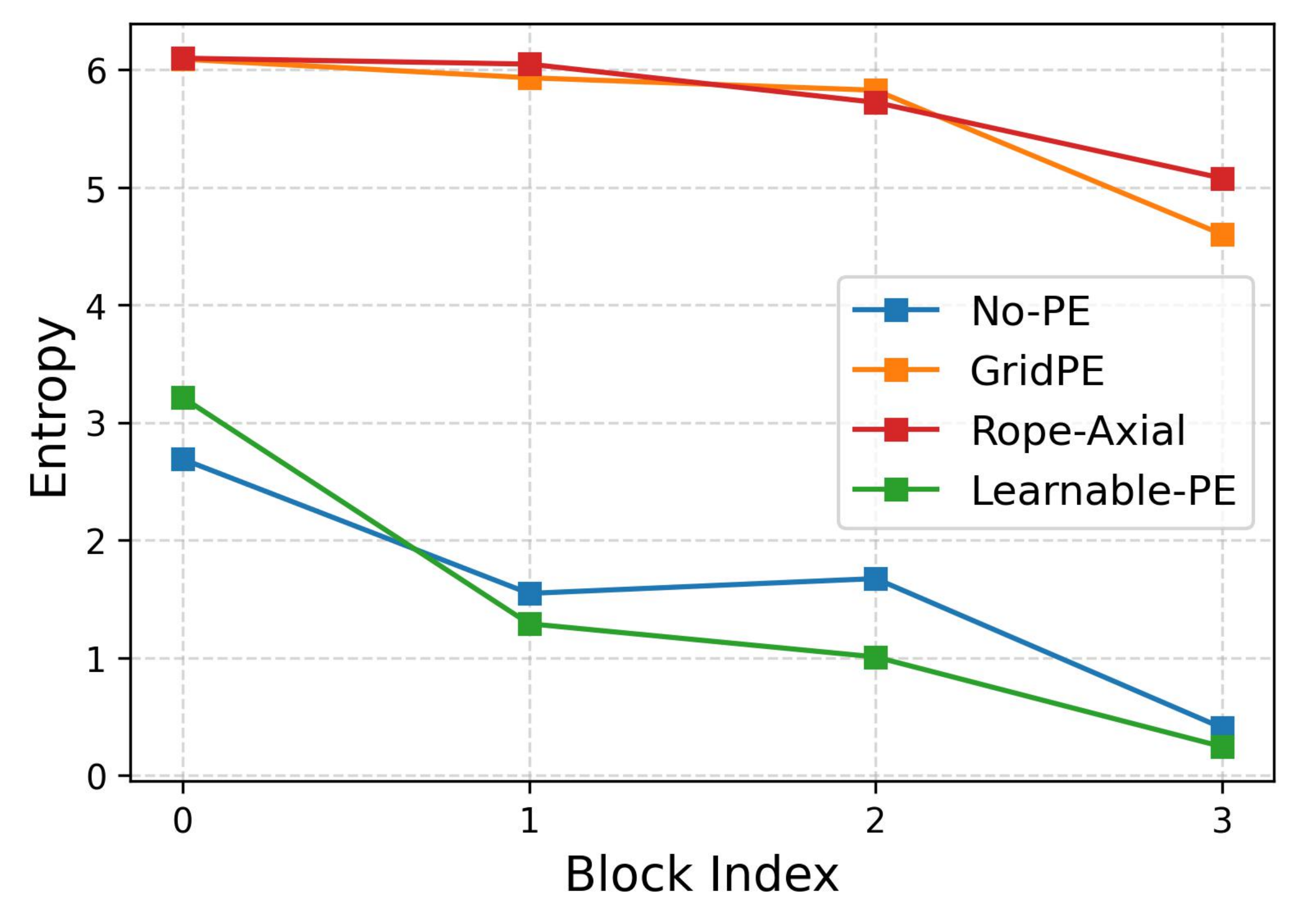}
      \caption{Input point = 896}
    \end{subfigure}

    \vspace{1ex}
    \begin{subfigure}[b]{0.31\linewidth}
      \includegraphics[width=\linewidth]{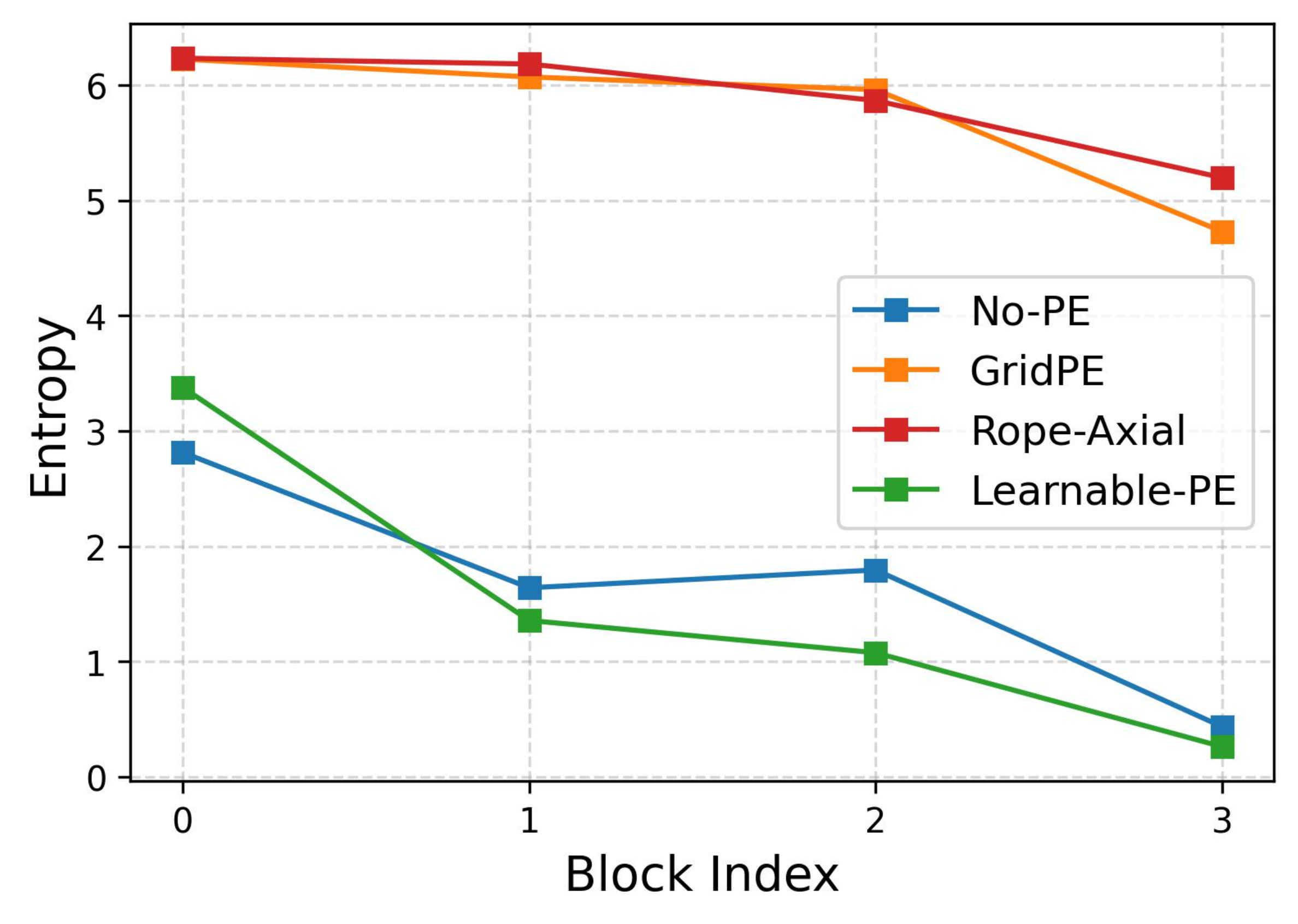}
      \caption{Input point = 1024}
    \end{subfigure}\hfill
    \begin{subfigure}[b]{0.31\linewidth}
      \includegraphics[width=\linewidth]{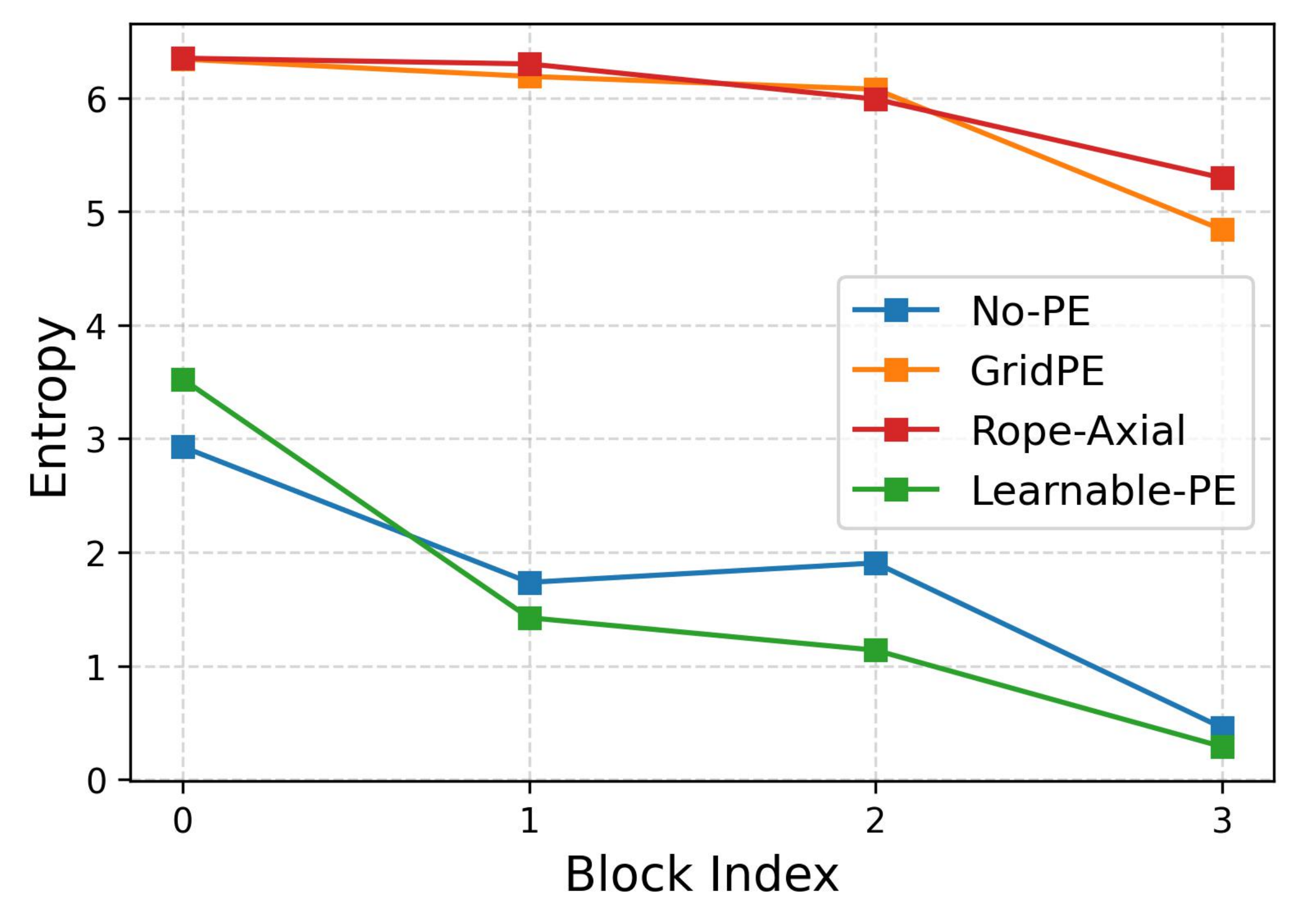}
      \caption{Input point = 1152}
    \end{subfigure}\hfill
    \begin{subfigure}[b]{0.31\linewidth}
      \includegraphics[width=\linewidth]{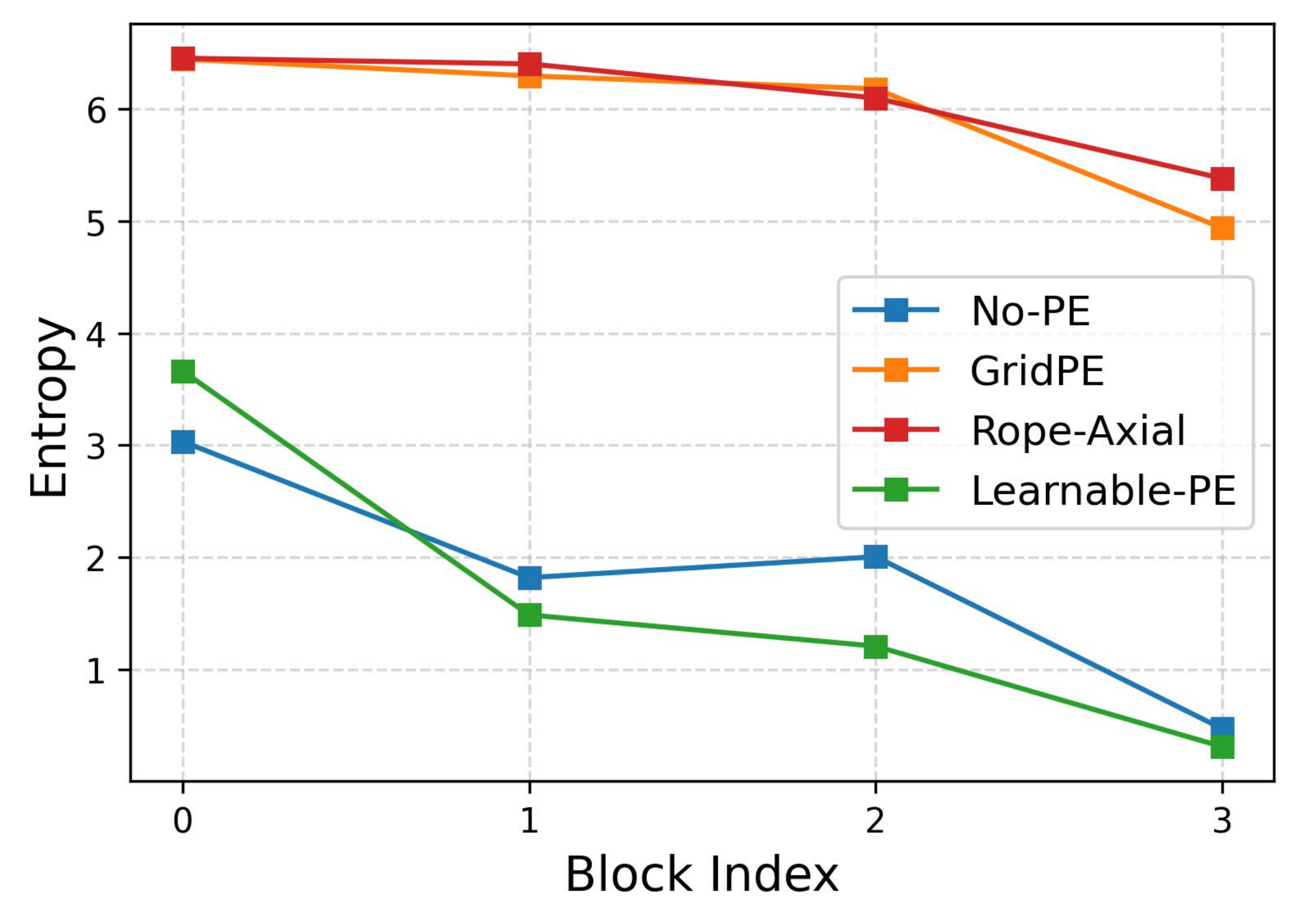}
      \caption{Input point = 1280}
    \end{subfigure}

    \vspace{1ex}
    \begin{subfigure}[b]{0.31\linewidth}
      \includegraphics[width=\linewidth]{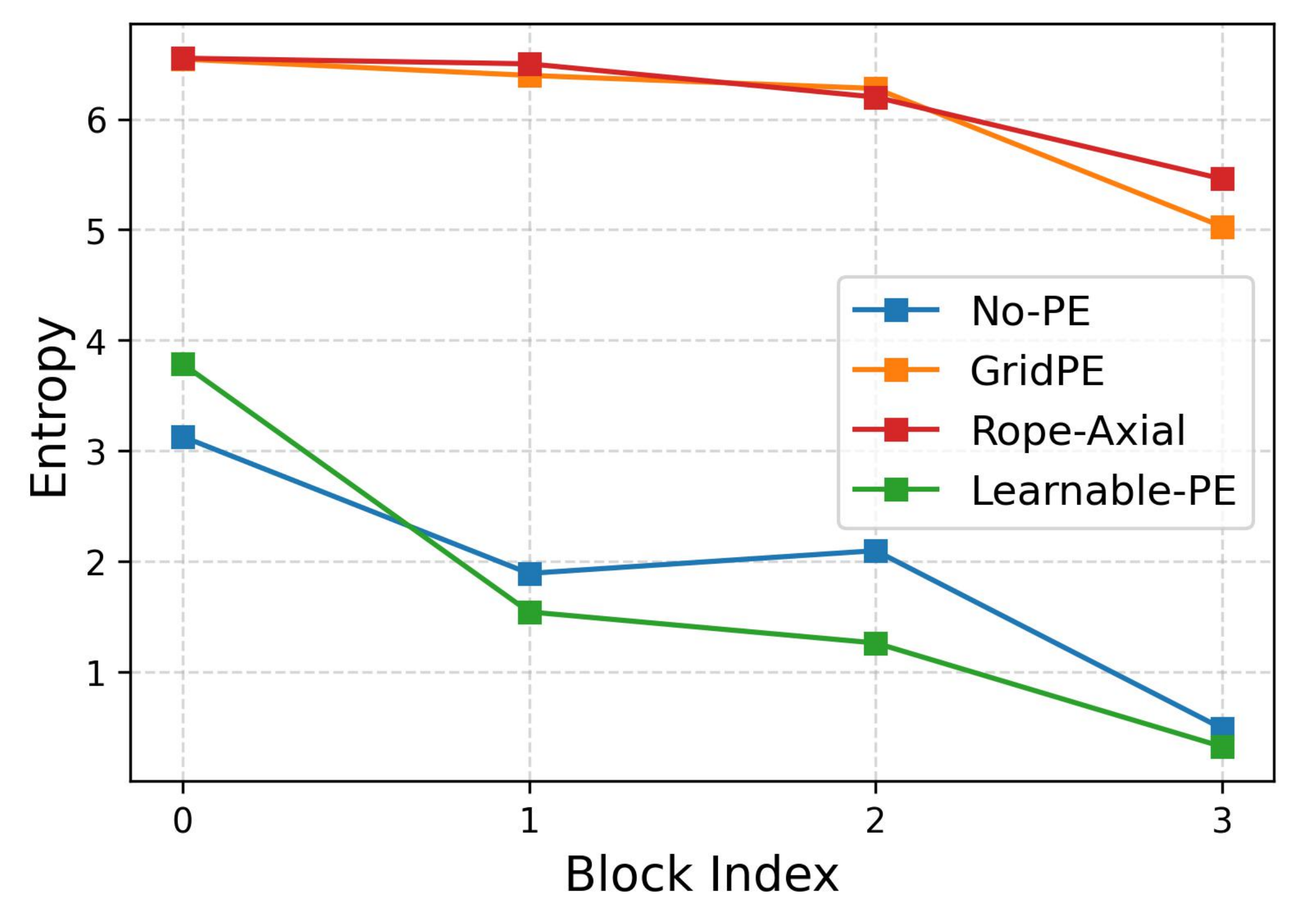}
      \caption{Input point = 1408}
    \end{subfigure}\hfill
    \begin{subfigure}[b]{0.31\linewidth}
      \includegraphics[width=\linewidth]{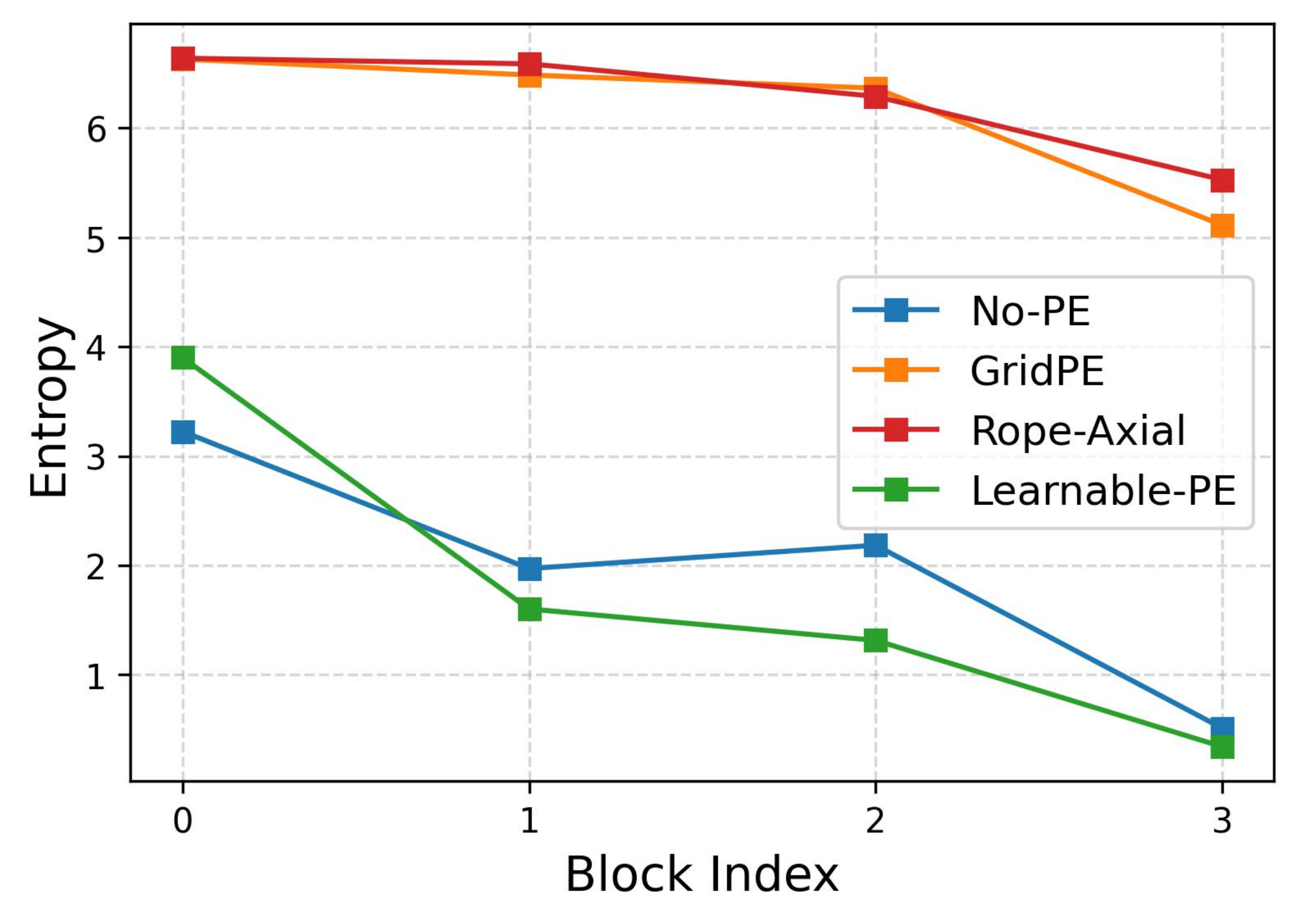}
      \caption{Input point = 1536}
    \end{subfigure}\hfill
    \begin{subfigure}[b]{0.31\linewidth}
      \includegraphics[width=\linewidth]{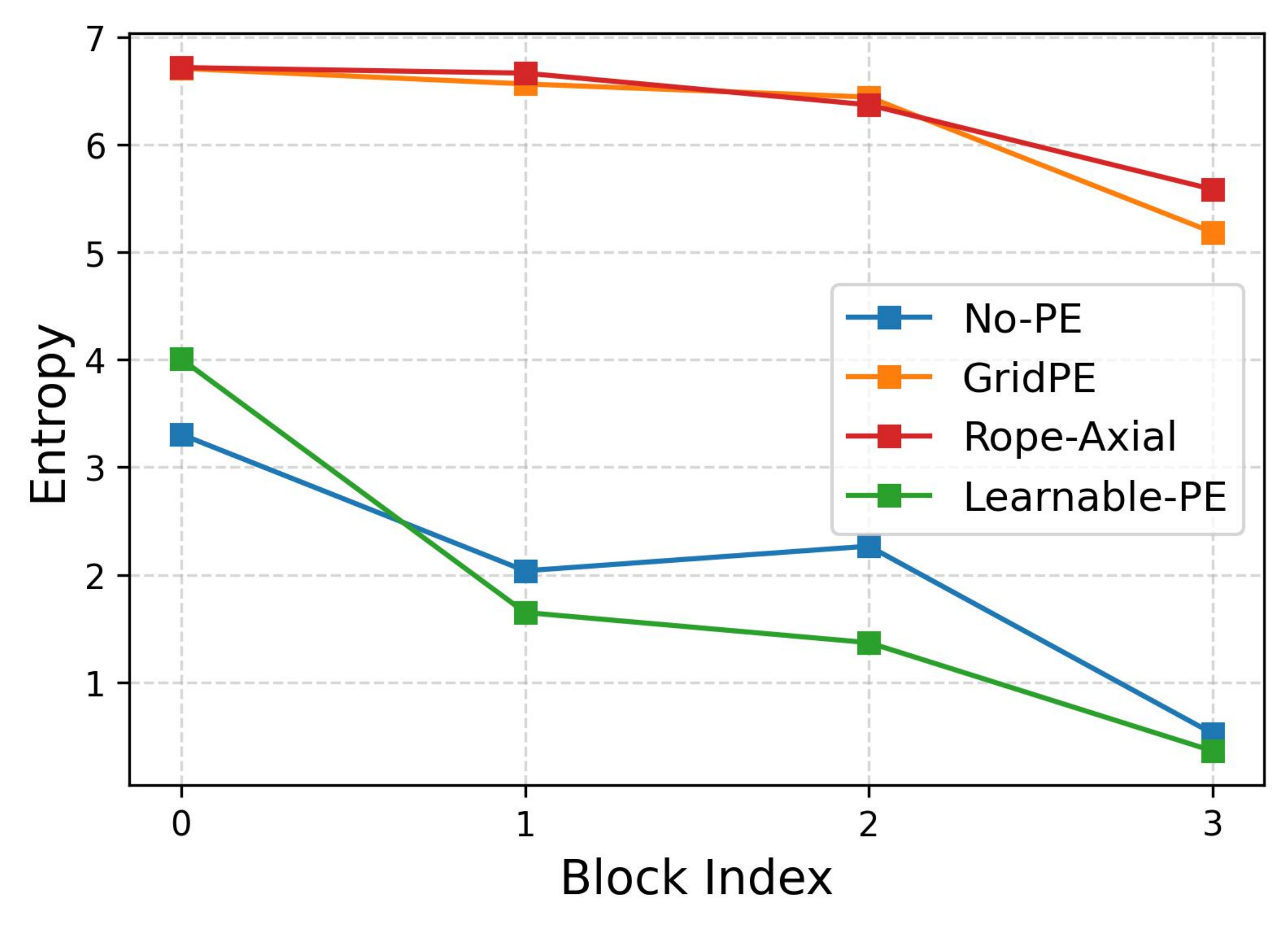}
      \caption{Input point = 1664}
    \end{subfigure}

    \vspace{1ex}
    \begin{subfigure}[b]{0.31\linewidth}
      \includegraphics[width=\linewidth]{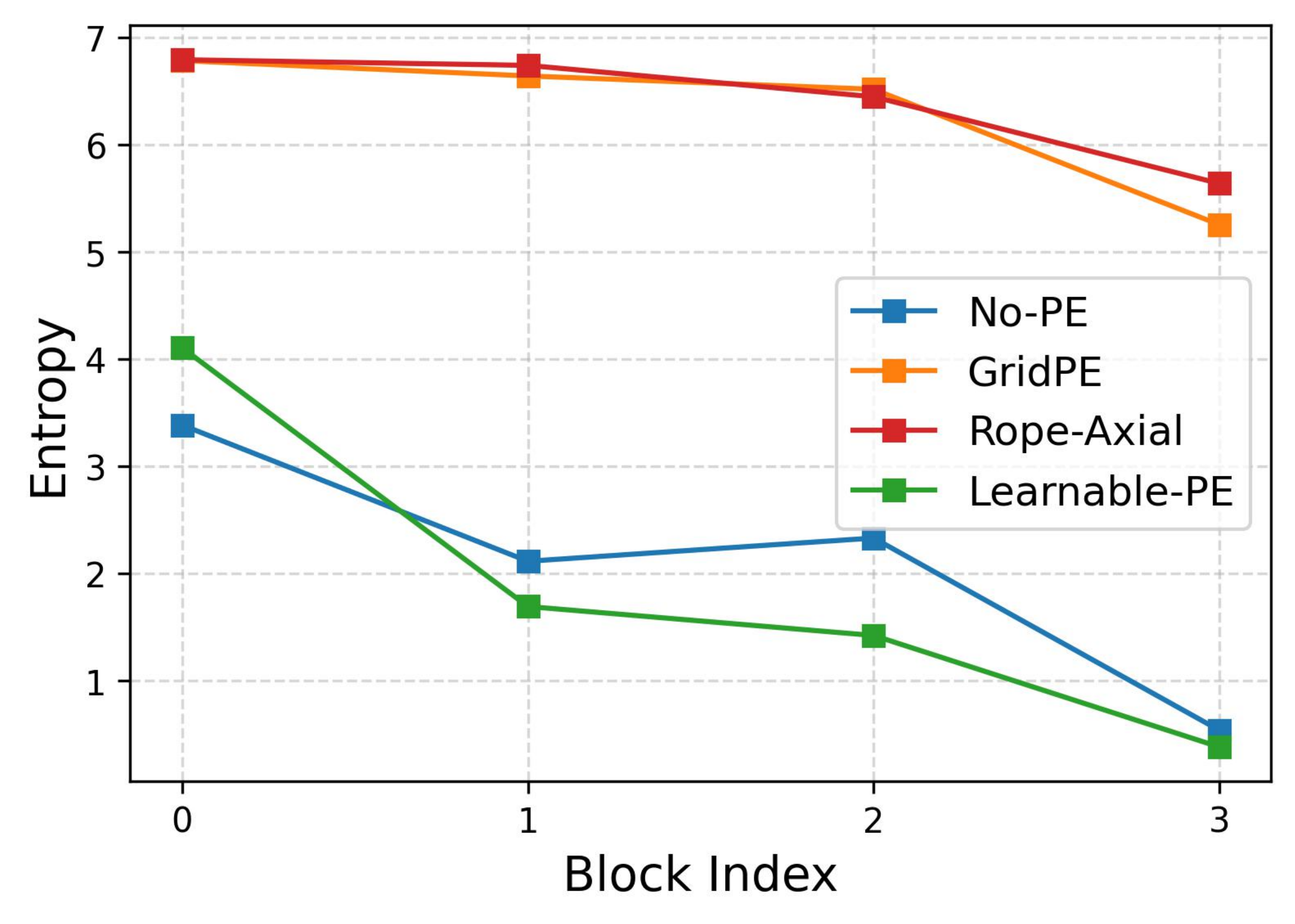}
      \caption{Input point = 1792}
    \end{subfigure}\hfill
    \begin{subfigure}[b]{0.31\linewidth}
      \includegraphics[width=\linewidth]{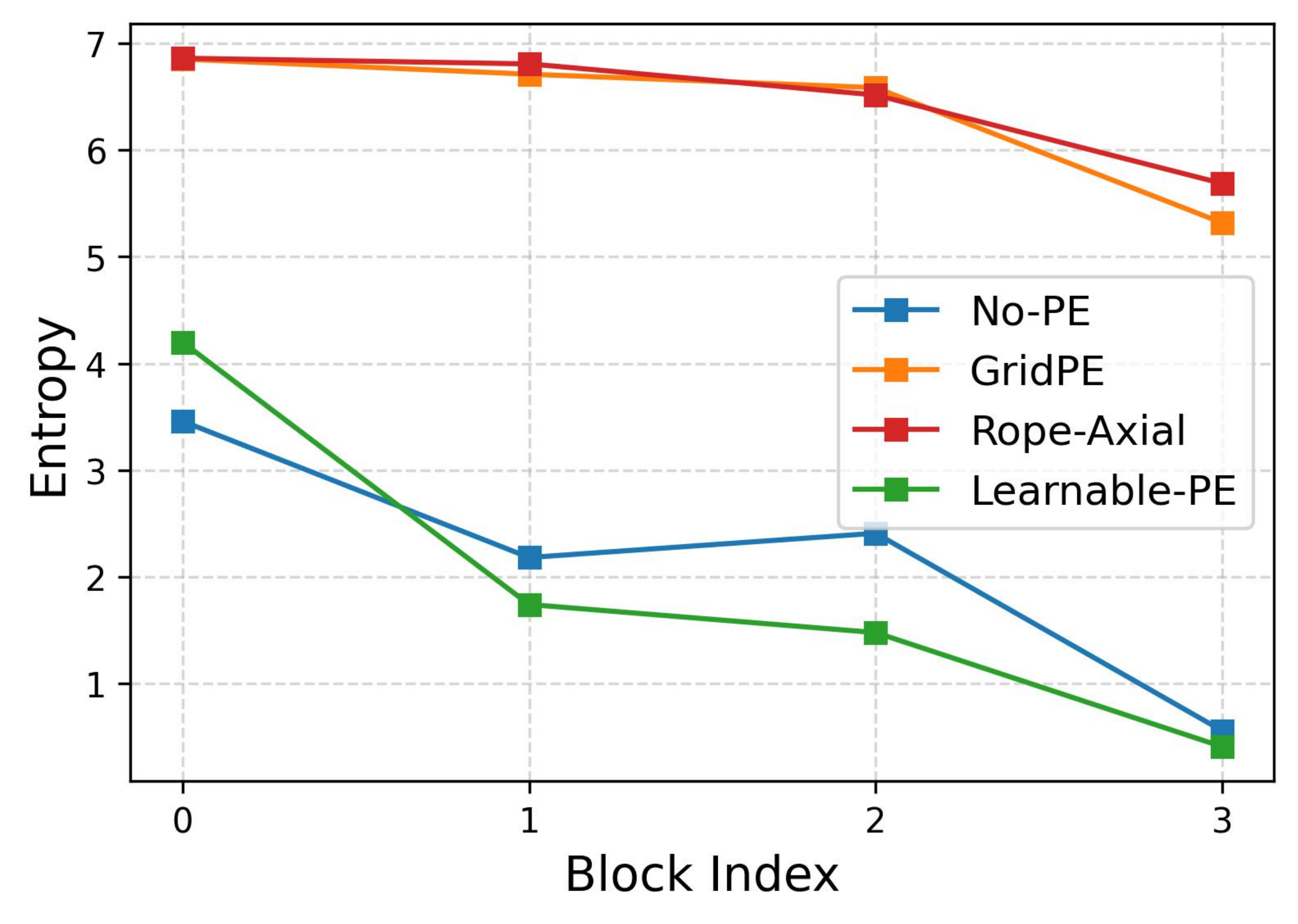}
      \caption{Input point = 1920}
    \end{subfigure}\hfill
    \begin{subfigure}[b]{0.31\linewidth}
      \includegraphics[width=\linewidth]{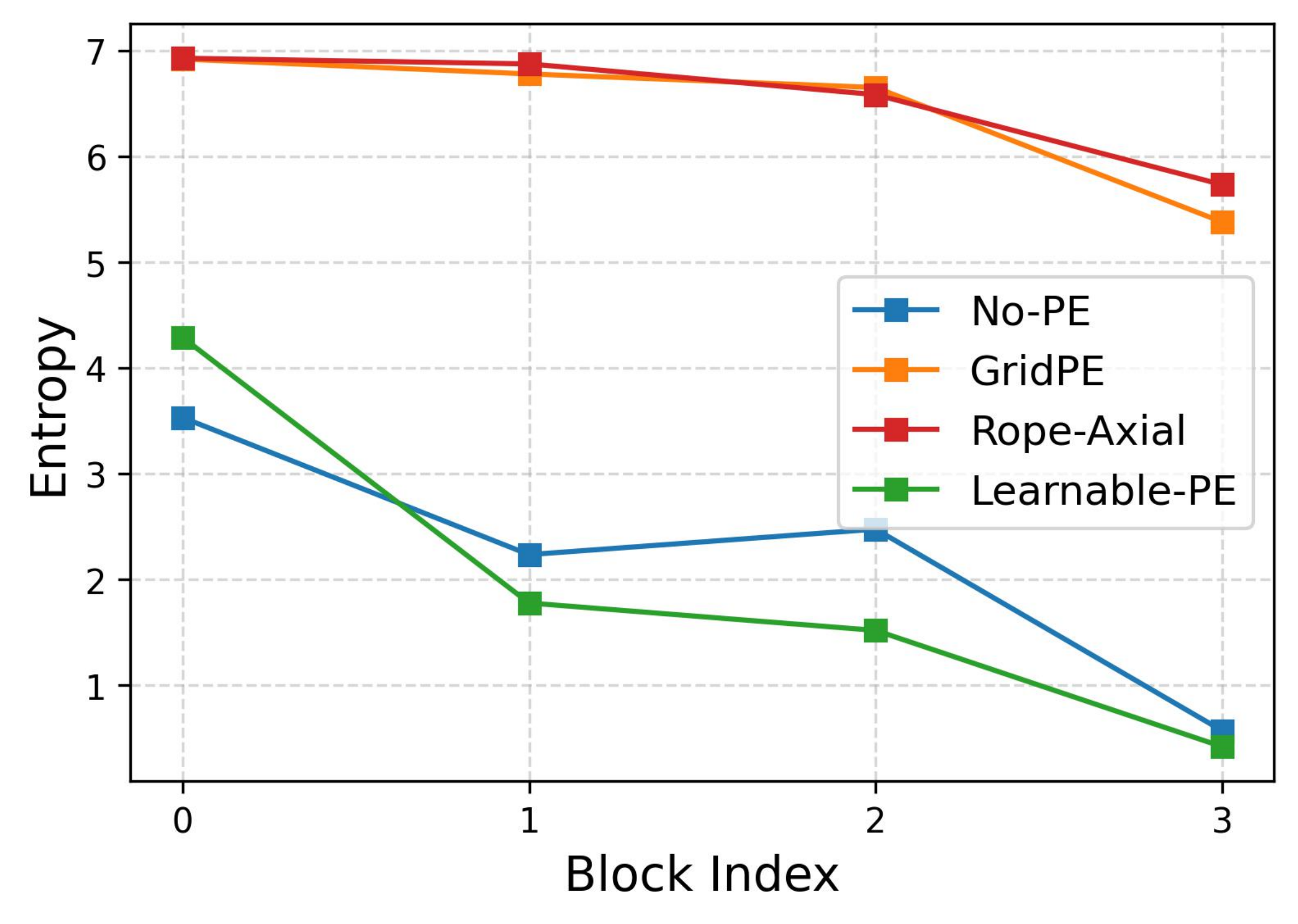}
      \caption{Input point = 2048}
    \end{subfigure}

    \caption{Mean attention entropy per block across input point counts.}
    \label{fig:entropy_all_3d}
\end{figure}

We quantify the effect of different positional embeddings under multi-scale inputs using two metrics: attention distance, which measures how far each query attends on average and reflects the model’s effective receptive field, and attention entropy, which captures how evenly the attention weights are distributed across tokens.
\vspace{-0.3cm}
\paragraph{2‑D Image Classification.} Figures~\ref{fig:attn_all} and~\ref{fig:entropy_all} reveal a common coarse trend: blocks closest to the input and the classifier token attend farther away, whereas intermediate blocks stay local. The key difference is how smoothly each embedding traverses this local–global axis. GridPE and RoPE‑axial exhibit monotonic, layer‑by‑layer changes in both attention distance and entropy, signalling a stable multi‑scale integration of context. By contrast, the ViT with learnable position embeddings exhibits a uniformly large attention distance, signalling limited adaptability to varying resolutions, whereas RoPE‑mixed shows sharp block‑to‑block fluctuations, indicating unstable attention allocation. In comparison, GridPE decomposes position into hierarchical frequency bands, allowing the model to reuse geometric priors at unseen resolutions and thereby achieve superior accuracy and scale generalisation on ViT‑S.
\vspace{-0.3cm}
\paragraph{3‑D Point‑Cloud Classification.} The 3‑D experiments reveal an even more pronounced contrast between models, as illustrated in Figures.~\ref{fig:distance_all_3d} and~\ref{fig:entropy_all_3d}. Learnable-PE exhibits very large attention distances combined with vanishing entropy---its attention either scatters over distant points or collapses onto a few tokens, failing to capture fine-grained geometry or coherent global structure. On the other hand, Pct without position embedding shows extremely low distance and entropy, overly fixating on immediate neighbors and thereby sacrificing neighborhood diversity. In contrast, GridPE and Rope-Axial strike a more favorable balance. They maintain low attention distances---anchoring queries to spatially adjacent points---while also exhibiting a higher entropy profile. This combination of locality and structured distribution allows the model to capture both fine-grained geometry and broader context. Importantly, this pattern remains stable across varying point densities, resulting in substantially better extrapolation and higher classification accuracy, highlighting the advantage of using advanced positional embedding method for spatial understanding in point-cloud data.

\subsection{Impact of Scale Numbers and Directional Selection in GridPE}
\label{appendix:gridpe_aliation}
To complement the findings presented in the main text, we provide additional ablation studies here. These include detailed comparisons on the number of positional embedding scales and the impact of using random versus fixed directional vectors across scales. All other experimental settings remain consistent with the main experiments. The observed trends and conclusions remain stable, reinforcing the robustness and flexibility of GridPE under varying parameter configurations. The following Table \ref{tab:ablation_accuracy} reports accuracy scores across different setups for further reference.

\begin{table}[htbp]
    \centering
    \caption{Effect of Scale Count and Direction Selection on Top-1 and Top-5 Accuracy in a 2D Image Classification Task}
    \begin{tabular}{|l|c|c|c|c|c|c|c|c|}
    \hline
    \textbf{Scale \& Num Head} & \textbf{160} & \textbf{192} & \textbf{224} & \textbf{256} & \textbf{320} & \textbf{384} & \textbf{448} & \textbf{512} \\
    \hline
    \multicolumn{9}{|c|}{\textbf{Top-1 Accuracy}} \\
    \hline
    64 \& 1 (Random) & 50.28 & 55.94 & 59.60 & 60.96 & 60.06 & 57.20 & 54.94 & 52.68 \\
    64 \& 1 (Fixed)  & 49.06 & 55.58 & 58.92 & 59.92 & 58.70 & 56.30 & 52.90 & 50.58 \\
    32 \& 2 (Random) & 54.60 & 61.18 & 65.38 & 65.98 & 64.10 & 61.58 & 57.38 & 54.22 \\
    32 \& 2 (Fixed)  & 51.48 & 59.64 & 63.52 & 64.88 & 63.12 & 60.82 & 58.72 & 56.36 \\
    16 \& 4 (Random) & 57.26 & 63.42 & 67.34 & 68.58 & 67.20 & 64.82 & 61.46 & 58.72 \\
    16 \& 4 (Fixed)  & 56.42 & 62.78 & 66.62 & 68.16 & 67.24 & 64.60 & 61.94 & \textbf{58.88} \\
    8 \& 8 (Random)  & \textbf{57.38} & \textbf{64.36} & \textbf{68.06} & \textbf{69.52} & \textbf{68.10} & 63.44 & 58.14 & 53.70 \\
    8 \& 8 (Fixed)   & 55.44 & 62.70 & 65.70 & 67.74 & 66.96 & 62.00 & 57.02 & 51.88 \\
    4 \& 16 (Random) & 55.30 & 62.28 & 65.98 & 66.94 & 67.40 & \textbf{65.94} & \textbf{63.20} & 58.18 \\
    4 \& 16 (Fixed)  & 54.20 & 61.24 & 64.42 & 66.30 & 67.06 & 64.90 & 60.32 & 54.20 \\
    2 \& 32 (Random) & 47.78 & 53.84 & 58.40 & 58.52 & 59.50 & 58.30 & 56.98 & 54.28 \\
    2 \& 32 (Fixed)  & 48.30 & 54.18 & 58.40 & 58.74 & 59.56 & 58.48 & 56.68 & 54.94 \\
    1 \& 64 (Random) & 43.16 & 49.24 & 55.36 & 54.66 & 55.98 & 54.20 & 52.58 & 50.50 \\
    1 \& 64 (Fixed)  & 44.52 & 50.24 & 54.76 & 55.06 & 56.74 & 54.80 & 53.68 & 50.82 \\
    \hline
    \multicolumn{9}{|c|}{\textbf{Top-5 Accuracy}} \\
    \hline
    64 \& 1 (Random) & 77.12 & 82.18 & 84.92 & 85.78 & 85.54 & 84.34 & 82.14 & 80.60 \\
    64 \& 1 (Fixed)  & 77.14 & 81.32 & 83.66 & 84.46 & 83.78 & 82.96 & 81.18 & 78.92 \\
    32 \& 2 (Random) & 80.42 & 85.86 & 87.76 & 88.50 & 87.96 & 85.32 & 83.74 & 81.34 \\
    32 \& 2 (Fixed)  & 79.08 & 84.22 & 87.04 & 87.92 & 87.34 & 85.50 & 84.26 & 82.88 \\
    16 \& 4 (Random) & 81.68 & \textbf{87.08} & \textbf{89.32} & 89.96 & \textbf{90.12} & \textbf{88.42} & 86.52 & \textbf{85.04} \\
    16 \& 4 (Fixed)  & 82.06 & 86.38 & 88.92 & 89.48 & 89.56 & 88.12 & \textbf{86.56} & 84.10 \\
    8 \& 8 (Random)  & \textbf{82.30} & 86.90 & 89.54 & \textbf{90.12} & 89.86 & 87.10 & 83.32 & 80.06 \\
    8 \& 8 (Fixed)   & 81.34 & 86.18 & 89.04 & 89.68 & 89.12 & 86.18 & 83.10 & 79.82 \\
    4 \& 16 (Random) & 80.34 & 85.94 & 88.12 & 88.98 & 89.48 & 88.22 & 86.28 & 83.48 \\
    4 \& 16 (Fixed)  & 79.54 & 84.56 & 87.12 & 88.08 & 88.82 & 87.26 & 84.74 & 80.06 \\
    2 \& 32 (Random) & 74.88 & 80.06 & 83.52 & 84.06 & 85.00 & 84.36 & 82.78 & 81.10 \\
    2 \& 32 (Fixed)  & 74.82 & 79.52 & 82.78 & 83.72 & 84.46 & 83.40 & 82.46 & 81.42 \\
    1 \& 64 (Random) & 71.44 & 77.42 & 81.02 & 81.80 & 83.20 & 81.60 & 80.06 & 78.50 \\
    1 \& 64 (Fixed)  & 72.52 & 77.66 & 81.12 & 81.92 & 82.78 & 82.24 & 81.10 & 79.66 \\
    \hline
    \end{tabular}
    \label{tab:ablation_accuracy}
\end{table}


\end{document}